\documentclass[numbers,sort&compress]{article}
\usepackage{graphicx}
\usepackage{lipsum}
\usepackage{rotating}
\usepackage[normalem]{ulem}
\usepackage{arxiv}
\usepackage{algpseudocode,algorithm}
\usepackage{mhchem}
\usepackage{amssymb,amsmath,amsthm}
\usepackage[numbers,sort&compress]{natbib}
\usepackage{xcolor}
\usepackage{hyperref}
\usepackage{tabularx}
\usepackage{adjustbox}
\usepackage[toc,page]{appendix}
\graphicspath{{figures/}{imgs/}}
\usepackage{amsfonts}
\usepackage{nicefrac}
\usepackage{microtype}
\usepackage{cleveref}
\usepackage{doi}
\usepackage{tikz}
\usetikzlibrary{positioning,arrows.meta,fit,backgrounds,calc,shapes.geometric}
\definecolor{RuhiTeal}{HTML}{004D40}
\definecolor{RuhiCoral}{HTML}{FF655D}
\definecolor{RuhiYellow}{HTML}{F1DB4B}
\definecolor{RuhiBlue}{HTML}{2196F3}
\definecolor{RuhiPink}{HTML}{E91E63}
\lstdefinelanguage{Rust}{morekeywords={fn,pub,struct,enum,impl,use,let,mut,for,if,else,match,return,self,Self,where,in,loop,while,break,continue,move,ref,as,true,false,mod,crate,super,type,const,static,trait,unsafe,async,await,dyn},sensitive=true,morekeywords=[2]{Vec,f64,usize,bool,Option,Result,Some,None,Ok,Err,Box,String},morecomment=[l]{//},morecomment=[s]{/*}{*/},morestring=[b]",morestring=[b]'}
\author{ \href{https://orcid.org/0000-0002-2393-8056}{\includegraphics[scale=0.06]{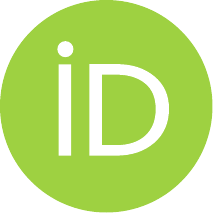}\hspace{1mm}Rohit Goswami}\thanks{Corresponding Author} \\
  Institute IMX and Lab-COSMO \\
  École polytechnique fédérale de Lausanne (EPFL) \\
  Station 12, CH-1015 Lausanne, Switzerland \\[1ex]
  \texttt{rgoswami@ieee.org}
}

\date{\today}
\title{A Tutorial Review of Bayesian Optimization with Gaussian Processes to Accelerate Stationary Point Searches}
\hypersetup{
 pdfauthor={},
 pdftitle={A Tutorial Review of Bayesian Optimization with Gaussian Processes to Accelerate Stationary Point Searches},
 pdfkeywords={},
 pdfsubject={},
 pdfcreator={Emacs 30.2 (Org mode 9.7.11)}, 
 pdflang={English}}
\begin{document}

\maketitle
\begin{abstract} %
Building local surrogates to accelerate stationary point searches on
potential energy surfaces spans decades of effort.
Done correctly, surrogates can reduce the number of expensive electronic
structure evaluations by roughly an order of magnitude while preserving
the accuracy of the underlying theory, with the gain depending on oracle
cost, search distance, and the availability of analytical forces.
We present a unified Bayesian optimization view of minimization, single-point
saddle searches, and double-ended path searches: all three share one
six-step surrogate loop and differ only in the inner optimization target
and the acquisition criterion.
The framework uses Gaussian process regression with derivative observations,
inverse-distance kernels, and active learning, and we develop optional
extensions for production use, including farthest-point sampling with the
Earth Mover's Distance, MAP regularization, an adaptive trust radius, and
random Fourier features for scaling.
Accompanying pedagogical Rust code demonstrates that all three applications
use the same Bayesian optimization loop, bridging the gap between
theoretical formulation and practical execution.
\end{abstract}
\vspace{0.7em}
\noindent
\section{Introduction}
\label{sec:introduction}
Chemical reactions, atomic diffusion in crystals, and conformational changes in proteins all represent trajectories through a high-dimensional configurational space.
For a system of \(N\) atoms, assuming a ground electronic state under the Born-Oppenheimer approximation decouples electronic and nuclear motion.
Fundamentally, this physical decoupling simply maps a 3N-dimensional spatial coordinate \(\mathbf{x}\) to a relative scalar energy \(y\), generating the potential energy surface (PES).
Due to methodological differences in calculations of absolute energies, the exact numerical value of \(y\) holds no intrinsic physical weight for locating geometries.
According to Boltzmann statistics, stable species occupy hyper-volumes around local energy minima \cite{hanggiReactionrateTheoryFifty1990}.
Transitioning between stable states requires the system to cross a dividing surface, optimally through a first-order saddle point where the gradient vanishes and the Hessian matrix possesses exactly one negative eigenvalue, with the eigenvector corresponding to this negative eigenvalue taken to be the reaction coordinate.

Harmonic transition state theory (HTST) \cite{eyringActivatedComplexChemical1935,vineyardFrequencyFactorsIsotope1957,petersReactionRateTheory2017} relates the saddle point to the rate constant:

\begin{equation}
k_{\text{HTST}} = \frac{\prod_{i=1}^{3N} \nu_i^{\text{min}}}{\prod_{i=1}^{3N-1} \nu_i^{\text{SP}}} \exp\left(-\frac{E_{\text{SP}} - E_{\text{min}}}{k_B T}\right)
\label{eq:htst}
\end{equation}

The rate depends exponentially on the barrier \(E_{\text{SP}} - E_{\text{min}}\), with the vibrational frequency ratio (\(\nu_i^{\text{min}}\) at the minimum, \(\nu_i^{\text{SP}}\) at the saddle) acting as a prefactor that captures the width of the two basins.
The denominator runs over \(3N-1\) modes because the saddle has one imaginary frequency along the reaction coordinate under four assumptions covered elsewhere \cite{refiorentinMethodologicalFrameworksComputational2026}.
In practice, predicting the rate reduces to finding the minima and saddle points \footnote{along with the Hessian test for the first order saddle point estimate, often skipped in high throughput workflows}.
To this end we consider minimization on an energy surface, along with two common modalities to find saddle points.
We will use the term "point searches" to cover the union of the search methods.

Only energy differences matter for locating stationary points, so the additive zero of energy is arbitrary and the map from geometry to energy and atomistic forces (the pointwise derivative w.r.t positions) is defined only up to a constant offset.
One can construct several alternative surrogate energy surfaces that perfectly preserve the minima and saddles of the true PES, even if those surrogates distort the broader configurational space.
This inherently local understanding of the process provides the leeway required to handle the extreme dimensionality of the \(3N\) space despite there being "no-free-lunch" \cite{jamesIntroductionStatisticalLearning2013} and has connections to vibrational analysis \cite{madsenApproximateFactorizationMolecular1997}.

Point searches typically need hundreds of such evaluations to converge, and this cost becomes prohibitive in large-scale studies where workflow-driven screening with tools such as AiiDA \cite{pizziAiiDAAutomatedInteractive2016} or Snakemake \cite{molderSustainableDataAnalysis2021} may require characterizing thousands of distinct transitions.
The problem is compounded in applications that embed saddle searches as an inner loop.
Adaptive kinetic Monte Carlo (AKMC) \cite{henkelmanLongTimeScale2001,goswamiEfficientExplorationChemical2025} discovers escape routes on the fly from each visited minimum, reaction network exploration catalogs competing pathways in complex catalytic cycles, and high-throughput screening for materials design may require characterizing saddle points across hundreds of candidate systems.
In each case, the per-search cost of electronic structure evaluations is the rate-limiting step, and reducing it from hundreds of calls to tens opens qualitatively different scales of investigation.

When electronic structure methods calculate \(y\) and its gradients, single queries consume minutes to hours.
Machine learning has opened two distinct approaches to this bottleneck.
The first strategy constructs \emph{global} machine-learned interatomic potentials (MLIPs), models trained on a large database of electronic structure calculations that approximates the PES over a broad region of configuration space by mapping onto a descriptor space \cite{musilPhysicsInspiredStructuralRepresentations2021,grisafiAtomicScaleRepresentationStatistical2019}.
Gaussian Approximation Potentials (GAP) \cite{bartokGaussianApproximationPotential2010} using SOAP descriptors, moment tensor potentials (MTP) \cite{shapeevMomentTensorPotentials2016}, and neural network potentials (NNP) \cite{behlerGeneralizedNeuralNetworkRepresentation2007} exemplify this approach; the review by Deringer, Bartok, and Csanyi \cite{deringerGaussianProcessRegression2021} covers this area in detail.
More recently, universal foundation models such as PET-MAD \cite{mazitovPETMADLightweightUniversal2025,bigiPushingLimitsUnconstrained2026} and MACE-MP-0 \cite{batatiaFoundationModelAtomistic2025} have demonstrated transferability across the periodic table.
These global models enable fast energy and force evaluation and allow molecular dynamics, geometry optimizations, and NEB calculations to run at near-electronic-structure accuracy.

Saddle points in particular however, occupy a vanishingly small, rarely sampled fraction of the total volume.
Applying a global MLIP to saddle point searches encounters a fundamental sampling problem.
Saddle points are rare events, and equilibrium sampling almost never visits them.
Random structure searches explore configuration space without bias toward the transition region.
Most training sets assembled from these approaches will have a blind spot precisely where it matters for kinetics \cite{schreinerTransition1xDatasetBuilding2022}.
The MLIP may give accurate energies near minima (where training data is plentiful) but unreliable predictions near the transition state (where it is sparse).
Retraining a global potential or even fine-tuning for every novel reaction pathway could defeat the purpose of high-throughput screening.

The second strategy, and the subject of this review, constructs a \emph{local, ephemeral} surrogate of the PES on-the-fly during each individual search, using only the data generated in the course of that specific calculation.
This is an active learning approach \cite{hennigProbabilisticMachineLearning2023,suttonReinforcementLearningIntroduction2018} in which the surrogate model decides where to sample next, balancing exploitation of the current prediction against exploration of uncertain regions \cite{snoekPracticalBayesianOptimization2012,suttonReinforcementLearningIntroduction2018}.
The surrogate need not represent the PES globally; it only needs to be accurate near the path being optimized.
Convergence to a saddle point typically requires on the order of 30 electronic structure evaluations \cite{goswamiEfficientImplementationGaussian2025,goswamiAdaptivePruningIncreased2025}, compared to the thousands needed for even a modest MLIP.
The surrogate is discarded after each search completes.

Several other groups have built per-search GP surrogates along similar lines: FLARE and committee-based GAP for on-the-fly molecular dynamics \cite{vandermauseOntheflyActiveLearning2020,bartokGaussianApproximationPotential2010b}, restricted-variance and gradient-enhanced kriging for geometry optimization \cite{raggiRestrictedVarianceMolecularGeometry2020,fdez.galvanRestrictedVarianceConstrainedReaction2021}, and earlier GPR-accelerated NEB and dimer codes \cite{koistinenNudgedElasticBand2017,koistinenNudgedElasticBand2019,denzelGaussianProcessRegression2019,vishartAcceleratingCatalysisSimulations,garridotorresLowScalingAlgorithmNudged2019}.
Another branch of the same broader surrogate-optimization literature changes the role of the GP prior mean itself: adaptive prior-mean universal kriging in curvilinear coordinates for molecular geometry optimization \cite{tengSpurMolecularGeometry2023}, physical-prior-mean CI-NEB \cite{tengPhysicalPriorMean2024}, and meta-GPs that recycle previously learned local PESs as prior functions for conformer exploration \cite{tengExploringTorsionalConformer2023}.
These methods solve closely related problems, but with different emphases in coordinates, prior construction, and cross-task reuse.
The present tutorial stays centered on the zero-mean plus constant-offset local-GP formulation shared by the Rust implementation and the companion production studies.
The unifying point is simple: minimization, dimer, and NEB share one Bayesian optimization loop, and the differences reduce to the inner optimizer and the acquisition rule.

The local GP accelerates exactly the PES that the user intends to study.
There is no dependence on a training database assembled at some other level of theory or for some other class of systems.
The GP learns from the electronic structure method of choice (DFT, coupled cluster, or any other) as the search proceeds, and the accuracy of the surrogate improves precisely where it matters, near the transition path under investigation.
In contrast, deploying a global MLIP for saddle point searches requires either (a) retraining or fine-tuning the potential for each new system and electronic structure method, or (b) accepting that the pre-trained potential may be unreliable in the transition state region where training data is sparse.
The local GP sidesteps this problem entirely, being system-specific and method-specific by construction.
This distinction between global and local modeling forms the conceptual backbone of the review: Section \ref{sec:pes} establishes the physical setting and classical methods, Section \ref{sec:gpr} develops the GP framework, and Sections \ref{sec:gprdimer}, \ref{sec:gprneb}, and \ref{sec:gprmin} show how the local GP approach applies to each classical method.

Gaussian process regression (GPR) \cite{gelmanBayesianDataAnalysis,gramacySurrogatesGaussianProcess2020} is well matched to this local surrogate role.
The GP posterior provides two quantities used by the outer loop: the predicted energy surface (the posterior mean, which plays the role of a cheap surrogate PES on which standard optimizers can run) and a posterior variance (which tracks distance from the training data in the geometry induced by the kernel).
The variance is not a direct measure of accuracy against the true PES; it is a self-consistent statement of the surrogate's own disagreement with itself under its fitted hyperparameters, and it is driven down by sampling more (Section\textasciitilde{}\ref{sec:gpr}).
We use it where this is what we need, namely to detect that a region is under-sampled (the LCB stopping rule in the inner loop) and, in NEB, to prioritize image selection when several candidates are available; we do not use it as a per-proposal gate on the oracle, and we pair it with a geometric trust radius that encodes physical displacement bounds directly (Section\textasciitilde{}\ref{sec:gprmin}).

The two approaches differ in what the GP is asked to do.
In the MLIP setting, the GP is a \emph{regression} tool that, given a large, pre-computed training set, interpolates to approximate the energy at new configurations.
In the active learning setting developed here, the GP is a \emph{local surrogate} that is optimized on directly, with a trust region bounding each step and an oracle call concluding every outer iteration; the training set is assembled incrementally as a byproduct of the search itself.
The kernel operates on inverse interatomic distances rather than high-dimensional structural descriptors like the smoothed overlap of atomic positions, or SOAP \cite{caroOptimizingManybodyAtomic2019}, because the model only needs to resolve the PES in a local neighborhood where pairwise distance features suffice.
The inverse-distance kernel has been validated across 500+ reaction benchmarks with both SE and Matern kernels \cite{goswamiEfficientImplementationGaussian2025,goswamiAdaptivePruningIncreased2025,koistinenNudgedElasticBand2019}.
The computational overhead of the GP (dominated by the \(\mathcal{O}(M^3)\) Cholesky decomposition with \(M \sim 30\)) is negligible compared to the electronic structure cost it replaces, whereas an MLIP-scale GP with \(M \sim 10^4\) would require sparse approximations.

This tutorial review provides a self-contained treatment of GPR-accelerated stationary point searches.
We develop the mathematical foundations of GPR with derivative observations in sufficient detail for a practitioner to implement the method from scratch, and apply the framework to the dimer method, the nudged elastic band (NEB), and local minimization.
Accompanying Rust code (\texttt{chemgp-core}, source at \url{https://github.com/lode-org/ChemGP}) is the pedagogical reference implementation of every algorithm in the review; each equation maps to a specific function, and the crate runs the illustrative examples reported here.
Documentation is available at \url{https://lode-org.github.io/ChemGP/}.
Production-scale saddle-point studies on hundreds of molecular reactions use the C++ \texttt{gpr\_optim} code, reported in detail elsewhere \cite{goswamiEfficientImplementationGaussian2025,goswamiAdaptivePruningIncreased2025}, and the two implementations share the same algorithmic core.
We give practical guidance on hyperparameter selection, coordinate systems, trust regions, and data management, while keeping the main text focused on what a reader needs to understand the common loop before worrying about production refinements.
A unifying observation runs through the three applications: every GP method shares the same Bayesian optimization outer loop of training a surrogate, selecting a query point by an acquisition criterion, and evaluating the oracle.
Four shared components (FPS subset selection, EMD trust, RFF approximation, and LCB-style uncertainty handling) are formalized as the Bayesian surrogate loop in Section \ref{sec:bo-framework} and then specialized in the method sections.
Accordingly, the tutorial is organized in three passes: first the classical search problems (Section \ref{sec:pes}), then the GP machinery that all three methods share (Sections \ref{sec:gpr} and \ref{sec:bo-framework}), and finally the method-specific and practical refinements (Sections \ref{sec:gprdimer}--\ref{sec:otgp}).
Algorithm \ref{alg:bo_loop_overview} gives the high-level skeleton shared by minimization, the dimer, and NEB; the detailed unified framework with its acquisition, training, and trust-region steps appears as Algorithm \ref{alg:bo_loop} in Section \ref{sec:bo-framework}.

The methods differ only in the optimization and acquisition phases.
Minimization uses L-BFGS descent and implicit acquisition; the dimer uses CG rotation plus L-BFGS translation with trust-clipped acquisition; the NEB uses path relaxation with UCB acquisition from unevaluated images.

\begin{algorithm}[H]
\caption{Unified Bayesian surrogate loop for stationary point searches (overview)}
\label{alg:bo_loop_overview}
\begin{algorithmic}[1]
\Require Initial configuration(s) $\mathbf{x}_0$, oracle tolerance $\epsilon$, trust radius $\Delta$
\State Initialize dataset $\mathcal{D} \leftarrow \{(\mathbf{x}_0, V(\mathbf{x}_0), \nabla V(\mathbf{x}_0))\}$
\State $\mathbf{x}^* \leftarrow \mathbf{x}_0$
\While{$|\nabla V(\mathbf{x}^*)| > \epsilon$}
    \State \textsc{Select} training subset $\mathcal{D}_{\text{train}} \subset \mathcal{D}$ via farthest point sampling
    \State \textsc{Train} hyperparameters $\boldsymbol{\theta}$ via MAP estimation 
    \State \textsc{Build} surrogate model $V_{\text{GP}}$ (Exact GP or Random Fourier Features)
    \State \textsc{Optimize} $\mathbf{x}_{\text{prop}}$ on $V_{\text{GP}}$ via method-specific inner loop
    \State \textsc{Clip} $\mathbf{x}_{\text{prop}}$ to trust region boundary $\Delta$ (EMD or Euclidean)
    \State \textsc{Acquire} new oracle evaluation at $\mathbf{x}_{\text{prop}}$ based on acquisition criterion
    \State Update dataset $\mathcal{D} \leftarrow \mathcal{D} \cup \{(\mathbf{x}_{\text{prop}}, V(\mathbf{x}_{\text{prop}}), \nabla V(\mathbf{x}_{\text{prop}}))\}$
    \State Update trust radius $\Delta$ based on surrogate accuracy
    \State $\mathbf{x}^* \leftarrow \mathbf{x}_{\text{prop}}$
\EndWhile
\State \Return Converged stationary point(s) $\mathbf{x}^*$
\end{algorithmic}
\end{algorithm}

The review is organized as follows.
Section \ref{sec:pes} establishes the physical setting and the classical search algorithms.
Section \ref{sec:gpr} develops the GPR framework, including molecular kernels and gradient observations.
Section \ref{sec:bo-framework} formalizes the Bayesian surrogate loop that unifies the three applications.
Section \ref{sec:gprdimer} presents GPR-accelerated minimum mode following (the GP-dimer), Section \ref{sec:gprneb} covers GPR-accelerated NEB, and Section \ref{sec:gprmin} treats GPR-accelerated minimization.
Section \ref{sec:otgp} introduces the Optimal Transport GP (OT-GP) extensions that address scaling and stability.
Section \ref{sec:examples} illustrates the methods on the Muller-Brown, LEPS, and PET-MAD systems, with reproducibility details and pointers to the executable code in the supporting information.
\section{The Potential Energy Surface and Stationary Point Searches}
\label{sec:pes}
\subsection{The PES and Its Stationary Points}
\label{sec:pes-def}
We collect the positions of \(N\) atoms into a single vector \(\mathbf{x} \in \mathbb{R}^{3N}\).
The PES \(V(\mathbf{x})\) \cite{lewarsComputationalChemistry2016} gives the energy at each configuration, and the atomic force vector is its negative gradient:

\begin{equation}
\mathbf{F}(\mathbf{x}) = -\nabla V(\mathbf{x})
\label{eq:force}
\end{equation}

Every algorithm in this review queries \(V\) and \(\mathbf{F}\) at chosen configurations to locate stationary points, configurations where \(\nabla V(\mathbf{x}^*) = \mathbf{0}\).
For the search algorithms that follow, only two types of stationary point matter: local minima (all Hessian eigenvalues positive) and first-order saddle points (exactly one negative eigenvalue).
The eigenvector belonging to that negative eigenvalue is the \emph{minimum mode} and points along the reaction coordinate.
Six zero eigenvalues from rigid-body translation and rotation must be projected out.
Figure \ref{fig:mb_neb} shows the Muller-Brown surface \cite{mullerLocationSaddlePoints1979}, a standard 2D test PES, with its three minima, two saddle points, and a converged minimum energy path.

The \emph{minimum energy path} (MEP) \cite{petersReactionRateTheory2017} connects a saddle point to the adjacent minima along the steepest descent:

\begin{equation}
\frac{d\mathbf{x}}{ds} = -\frac{\nabla V(\mathbf{x})}{|\nabla V(\mathbf{x})|}
\label{eq:mep}
\end{equation}

where \(s\) is arc length.
The energy difference between the saddle and the minimum is the barrier that enters the HTST rate (Eq. \ref{eq:htst}) \cite{refiorentinMethodologicalFrameworksComputational2026}.
The two families of algorithms in Sections \ref{sec:dimer} and \ref{sec:chain-of-states} approach the problem from opposite ends.
The dimer method \cite{henkelmanDimerMethodFinding1999} searches for the saddle without knowing the MEP, while the NEB \cite{jonssonNudgedElasticBand1998} approximates the entire MEP and locates the saddle as its highest point through the climbing image extension \cite{henkelmanClimbingImageNudged2000}.
Both classical methods rely on repeated evaluations of the true PES and its gradients, which motivates the GP acceleration strategies developed in Sections \ref{sec:gpr} and \ref{sec:gprdimer}.
\subsection{Local Minimization}
\label{sec:minimization}
Local minimization is the simplest stationary point problem, where the system relaxes along the negative gradient until the forces vanish.
Limited-memory Broyden--Fletcher--Goldfarb--Shanno (L-BFGS) \cite{liuLimitedMemoryBFGS1989,byrdLimitedMemoryAlgorithm1995,nocedalNumericalOptimization2006} approximates the Hessian using gradient evaluations, to save on computing the full Hessian.
The L-BFGS here finds use for both dimer translation and surrogate optimization, and plays a central role throughout the GPR-accelerated algorithms.
L-BFGS maintains a history of \(m\) recent position and gradient differences \(\{(\mathbf{s}_k, \mathbf{y}_k)\}\) and computes a search direction via the two-loop recursion without explicitly forming the Hessian.

\begin{equation}
\mathbf{s}_k = \mathbf{x}_{k+1} - \mathbf{x}_k, \quad \mathbf{y}_k = \nabla V(\mathbf{x}_{k+1}) - \nabla V(\mathbf{x}_k)
\label{eq:lbfgs_pairs}
\end{equation}

\begin{equation}
\rho_k = \frac{1}{\mathbf{y}_k^T \mathbf{s}_k}, \quad H_k^0 = \frac{\mathbf{s}_{k-1}^T \mathbf{y}_{k-1}}{\mathbf{y}_{k-1}^T \mathbf{y}_{k-1}} \mathbf{I}
\label{eq:lbfgs_scaling}
\end{equation}

The L-BFGS direction is then obtained by the standard two-loop recursion applied to the current gradient.
This optimizer is used both within the dimer method (for translation) and within the GPR-accelerated algorithms (for optimization on the surrogate surface).

Section \ref{sec:gpr} develops the GP framework that makes surrogate-accelerated optimization possible, including the inverse-distance kernel for molecular systems and the derivative observations that provide 3N+1 constraints per evaluation.
\subsection{Minimum Mode Following, the Dimer Method}
\label{sec:dimer}
The dimer method \cite{henkelmanDimerMethodFinding1999} estimates PES curvature by comparing forces at two nearby points, much like a finite-difference approximation to the second derivative but applied directionally.
This costs just two force evaluations per direction, rather than the \(\sim 6N\) evaluations needed for the full \(3N \times 3N\) Hessian (or an analytic Hessian, which most electronic structure codes do not expose).
The single lowest curvature direction, the minimum mode, points along the reaction coordinate.
The two-point probe rotates until the most negative curvature is located, and the system walks uphill along it.
The dimer's cost is dominated by the rotation phase, where each translation step requires 5--15 rotation evaluations (each a full electronic structure call) to converge the orientation.
This inner-loop cost is where the GP surrogate provides the largest savings \cite{goswamiEfficientImplementationGaussian2025,goswamiAdaptivePruningIncreased2025,koistinenMinimumModeSaddle2020,denzelGaussianProcessRegression2018a}, as we discuss in Section \ref{sec:gprdimer}.

We take the original formulation of the dimer for simplicity; extensions such as single-force rotation variants and Householder-based translations \cite{olsenComparisonMethodsFinding2004,goswamiEnhancedClimbingImage2026} do not affect the surrogate integration developed below.

Concretely, two replicas of the system are placed symmetrically about a midpoint \cite{goswamiEfficientExplorationChemical2025}:

\begin{equation}
\mathbf{R}_{1,2} = \mathbf{R} \pm \Delta R \, \hat{\mathbf{N}}
\label{eq:dimer_images}
\end{equation}

where \(\hat{\mathbf{N}}\) is a unit vector (the dimer axis) and \(\Delta R\) is the midpoint-to-endpoint separation, matching the \texttt{dimer\_sep} convention used in \texttt{chemgp-core}, typically 0.01 \AA{}.
The algorithm alternates between two operations (Figure S1, left): rotating the dimer to find the minimum curvature direction, and translating the midpoint uphill along that direction while relaxing perpendicular to it.
\subsubsection{Rotation}
\label{sec:dimer-rotation}
The curvature along the dimer axis is estimated from the force difference at the two endpoints:

\begin{equation}
C(\hat{\mathbf{N}}) \approx \frac{(\mathbf{F}_2 - \mathbf{F}_1) \cdot \hat{\mathbf{N}}}{\Delta R}
\label{eq:curvature}
\end{equation}

where \(\mathbf{F}_{1,2} = \mathbf{F}(\mathbf{R}_{1,2})\).
Eq. \ref{eq:curvature} is a directional finite-difference curvature estimate: the force difference across the dimer, projected onto the axis, divided by the midpoint-to-endpoint separation.
When \(C < 0\), the axis points along a direction of negative curvature.
The rotation phase minimizes \(C\) over all orientations of \(\hat{\mathbf{N}}\), which aligns the dimer with the lowest curvature mode.
The perpendicular component of the force difference supplies the gradient for this minimization, and conjugate gradient (CG) \cite{eppersonIntroductionNumericalMethods2012} with the Polak-Ribiere \cite{polakNoteConvergenceMethodes1969} update provides the search direction:

\begin{equation}
\beta_i = \frac{(\mathbf{F}_i^{\perp} - \mathbf{F}_{i-1}^{\perp}) \cdot \mathbf{F}_i^{\perp}}{|\mathbf{F}_i^{\perp}|^2}
\label{eq:polak_ribiere}
\end{equation}

CG is the default for rotation \cite{heydenEfficientMethodsFinding2005}.
L-BFGS \cite{kastnerSuperlinearlyConvergingDimer2008} converges in fewer steps when the minimum mode is well-separated, but it can lock onto a wrong mode when the gap is small \cite{olsenComparisonMethodsFinding2004}.
Bayesian benchmarking \cite{goswamiBayesianHierarchicalModels2025} and theoretical considerations favor CG overall for the rotation phase \cite{lengEfficientSoftestMode2013}, with the advantage concentrated in systems with near-degenerate curvature modes.

When the dimer operates on molecular systems (as opposed to 2D model surfaces), rigid-body translations and rotations must be projected out of the \emph{translation step} \cite{goswamiAdaptivePruningIncreased2025,melanderRemovingExternalDegrees2015} to prevent the molecule from drifting through space.
Projecting these modes out of the \emph{orientation vector} \(\hat{\mathbf{N}}\) itself is catastrophic: it removes the component of the minimum mode that distinguishes a saddle point from a minimum.
In an 8-atom system, projecting translations from the orient vector changed the estimated curvature from \(-8.4\) eV/\AA{}\(^2\) (correct, negative) to \(+103\) eV/\AA{}\(^2\) (wrong sign), causing the dimer to walk away from the saddle point.
The rule is: project rigid-body modes from translation steps only, never from the dimer orientation.
The projection formula and Gram-Schmidt basis construction are detailed in the SI.
\subsubsection{Translation}
\label{sec:dimer-translation}
Once the rotation has identified the minimum mode, translation must move the midpoint \emph{uphill} along that mode while simultaneously relaxing in all other directions.
Geometrically, this is a Householder reflection \cite{nocedalNumericalOptimization2006}: the force vector is reflected about the hyperplane perpendicular to \(\hat{\mathbf{N}}\), which flips the sign of the component along the minimum mode while leaving the \(3N-1\) perpendicular components unchanged.
The system therefore climbs the ridge while sliding down into the valley on each side of it.
The modified force is:

\begin{equation}
\mathbf{F}^{\dagger} = \begin{cases}
\mathbf{F}(\mathbf{R}) - 2[\mathbf{F}(\mathbf{R}) \cdot \hat{\mathbf{N}}]\hat{\mathbf{N}} & \text{if } C(\hat{\mathbf{N}}) < 0 \\
-[\mathbf{F}(\mathbf{R}) \cdot \hat{\mathbf{N}}]\hat{\mathbf{N}} & \text{if } C(\hat{\mathbf{N}}) \geq 0
\end{cases}
\label{eq:modified_force}
\end{equation}

\begin{figure}[htbp]
\centering
\includegraphics[width=0.6\textwidth]{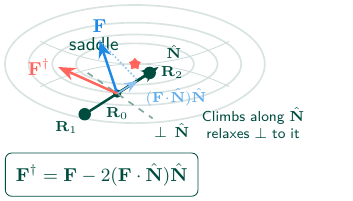}
\caption{\label{fig:dimer_geometry}Geometry of the dimer and Householder reflection. The dimer pair (\(\mathbf{R}_1, \mathbf{R}_2\)) straddles the midpoint \(\mathbf{R}_0\) with axis \(\hat{\mathbf{N}}\). The true force \(\mathbf{F}\) (blue) is reflected about the hyperplane perpendicular to \(\hat{\mathbf{N}}\), producing the modified force \(\mathbf{F}^{\dagger}\) (coral) that climbs along the minimum mode while relaxing perpendicular to it.}
\end{figure}

The first case (\(C < 0\)) applies the Householder reflection just described: the dimer is already in a region of negative curvature, so climbing along \(\hat{\mathbf{N}}\) while relaxing perpendicular to it drives the system toward the saddle.
The second case (\(C \geq 0\)) handles the early phase of the search when the dimer has not yet reached the transition region; here only the component along \(\hat{\mathbf{N}}\) is retained to push the system toward the ridge.
Translation uses L-BFGS, and the search terminates when the true force magnitude \(|\mathbf{F}(\mathbf{R})|\) drops below a threshold \(\epsilon_{\text{force}}\).
Figure \ref{fig:dimer_geometry} shows the dimer geometry and Householder reflection used in the translation step.
Algorithm \ref{alg:dimer} summarizes the full iteration.

\begin{algorithm}[H]
\caption{Classical dimer method}\label{alg:dimer}
\begin{algorithmic}[1]
\Require Midpoint $\mathbf{R}$, initial dimer axis $\hat{\mathbf{N}}$, separation $\Delta R$, force threshold $\epsilon_{\text{force}}$
\Repeat
  \State Place endpoints $\mathbf{R}_{1,2} = \mathbf{R} \pm \Delta R\,\hat{\mathbf{N}}$
  \State Evaluate $\mathbf{F}(\mathbf{R}_1), \mathbf{F}(\mathbf{R}_2)$ on true PES \Comment{2 oracle calls}
  \Repeat \Comment{Rotation}
    \State $C(\hat{\mathbf{N}}) \gets (\mathbf{F}_2 - \mathbf{F}_1) \cdot \hat{\mathbf{N}} / \Delta R$ \Comment{Eq.~\ref{eq:curvature}}
    \State Rotate $\hat{\mathbf{N}}$ to minimize $C$ (CG, Eq.~\ref{eq:polak_ribiere})
    \State Evaluate $\mathbf{F}(\mathbf{R}_1)$ at new endpoint \Comment{1 oracle call per rotation}
  \Until{rotation converged}
  \State \Comment{Translation}
  \If{$C(\hat{\mathbf{N}}) < 0$}
    \State $\mathbf{F}^{\dagger} \gets \mathbf{F}(\mathbf{R}) - 2[\mathbf{F}(\mathbf{R}) \cdot \hat{\mathbf{N}}]\hat{\mathbf{N}}$ \Comment{Eq.~\ref{eq:modified_force}, climb}
  \Else
    \State $\mathbf{F}^{\dagger} \gets -[\mathbf{F}(\mathbf{R}) \cdot \hat{\mathbf{N}}]\hat{\mathbf{N}}$ \Comment{push toward saddle}
  \EndIf
  \State $\mathbf{R} \gets \mathbf{R} + \text{L-BFGS step on } \mathbf{F}^{\dagger}$
\Until{$|\mathbf{F}(\mathbf{R})| < \epsilon_{\text{force}}$}
\State \Return saddle point at $\mathbf{R}$
\end{algorithmic}
\end{algorithm}

The dimer's cost is dominated by the rotation phase, since each rotation step requires a force evaluation at the new \(\mathbf{R}_1\) position (the midpoint force can be reused from the previous translation).
A typical search requires 5 to 15 rotations per translation step, and 20 to 50 translation steps to converge.
This inner-loop cost is where the GP surrogate provides the largest savings, as we discuss in Section \ref{sec:gprdimer}.

The dimer method establishes the conceptual template for GP acceleration: replace expensive inner-loop evaluations with cheap surrogate predictions, and only return to the true PES to validate and extend the training set.
Section \ref{sec:gprdimer} develops this idea in detail, showing how the GP replaces the rotation-phase force evaluations with surrogate queries, reducing the total evaluation count from hundreds to tens while preserving accuracy.
\subsection{Double-Ended Path Methods}
\label{sec:chain-of-states}
When both initial and final states are known, the MEP connecting them is found by optimizing a discrete chain of \(P+1\) images \(\{\mathbf{R}_0, \mathbf{R}_1, \ldots, \mathbf{R}_P\}\) with fixed endpoints.
Two competing requirements must be satisfied: the images must converge toward the MEP (shape optimization) and remain well-distributed along the path (spacing control).
The nudged elastic band (NEB) \cite{jonssonNudgedElasticBand1998} decouples them through force projections: the true force acts only perpendicular to the local tangent, driving images toward the MEP, while fictitious spring forces act only parallel, maintaining spacing.
The string method \cite{eStringMethodStudy2002,eSimplifiedImprovedString2007} replaces springs with interpolation-based reparameterization (\(k \to \infty\) limit).
Figure S2 (left) summarizes the NEB iteration.
\subsubsection{The Nudged Elastic Band}
\label{sec:neb}
A converged MEP satisfies the condition that the perpendicular force vanishes everywhere along the path:

\begin{equation}
(\nabla V)^{\perp}(\mathbf{R}) = \nabla V(\mathbf{R}) - [\nabla V(\mathbf{R}) \cdot \hat{\boldsymbol{\tau}}] \hat{\boldsymbol{\tau}} = \mathbf{0}
\label{eq:mep_condition}
\end{equation}

A naive elastic band, where springs connect the images and the full potential force acts on each one, fails to converge to the MEP for two reasons.
First, the potential force has a component \emph{along} the path that drags images away from the saddle region and bunches them in low-energy basins (the "sliding-down" artifact).
Second, the spring force has a component \emph{perpendicular} to the path that pulls the chain off the MEP into straight-line shortcuts through high-energy regions (the "corner-cutting" artifact) \cite{sheppardOptimizationMethodsFinding2008}.
The NEB eliminates both by projecting each type of force onto the subspace where it belongs.
The true force acts only perpendicular to the path, driving each image toward the MEP, while the spring force acts only parallel to the path, controlling image spacing.
The total NEB force on image \(i\) is:

\begin{equation}
\mathbf{F}_i^{\text{NEB}} = -\nabla V(\mathbf{R}_i)\big|_{\perp} + \mathbf{F}_i^{s}\big|_{\parallel}
\label{eq:neb_force}
\end{equation}

The perpendicular projection of the true force removes the sliding-down component:

\begin{equation}
-\nabla V(\mathbf{R}_i)\big|_{\perp} = -\nabla V(\mathbf{R}_i) + [\nabla V(\mathbf{R}_i) \cdot \hat{\boldsymbol{\tau}}_i] \hat{\boldsymbol{\tau}}_i
\label{eq:perp_force}
\end{equation}

and the parallel projection of the spring force prevents corner-cutting:

\begin{equation}
\mathbf{F}_i^{s}\big|_{\parallel} = k(|\mathbf{R}_{i+1} - \mathbf{R}_i| - |\mathbf{R}_i - \mathbf{R}_{i-1}|) \hat{\boldsymbol{\tau}}_i
\label{eq:spring_force}
\end{equation}

This decoupling of shape optimization (perpendicular) from spacing control (parallel) is the "nudging" that gives the method its name.

The spring constant \(k\) controls image spacing.
Energy-weighted springs \cite{asgeirssonNudgedElasticBand2021} replace the uniform \(k\) with image-dependent values \(k_i\) that increase near the energy maximum, concentrating resolution where it matters most for the barrier height while allowing wider spacing in the flat approach regions.

The tangent direction \(\hat{\boldsymbol{\tau}}_i\) enters every projection in the NEB force, so errors in the tangent propagate into the path shape.
The simplest estimate, bisecting the vectors to the two neighbors, \((\boldsymbol{\tau}_i^+ + \boldsymbol{\tau}_i^-) / 2\) with \(\boldsymbol{\tau}_i^{\pm} = \mathbf{R}_{i\pm 1} - \mathbf{R}_i\), breaks down at energy extrema along the path.
At a local maximum, the two neighbors are both downhill but in different directions, and their average can point perpendicular to the path rather than along it, producing visible kinks.
The \emph{improved tangent} estimate \cite{henkelmanImprovedTangentEstimate2000} fixes this by selecting the tangent from the higher-energy neighbor:

\begin{equation}
\hat{\boldsymbol{\tau}}_i = \begin{cases}
\boldsymbol{\tau}_i^+ & \text{if } V_{i+1} > V_i > V_{i-1} \\
\boldsymbol{\tau}_i^- & \text{if } V_{i+1} < V_i < V_{i-1} \\
\text{energy-weighted bisection} & \text{otherwise}
\end{cases}
\label{eq:improved_tangent}
\end{equation}

When the energy is monotonically increasing or decreasing through image \(i\), the tangent points toward the uphill neighbor.
When image \(i\) sits at a local extremum (the "otherwise" case), an energy-weighted average smoothly interpolates between the two directions.
The stability condition for the NEB requires that the parallel spring force be bounded by the product of the perpendicular curvature and the image spacing; the bisection tangent violates this at large \(P\), while the improved tangent does not.
\subsubsection{The Climbing Image and Its Connection to Minimum Mode Following}
\label{sec:cineb}
The standard NEB converges to a discretized MEP but does not place an image exactly at the saddle point.
The \emph{climbing image} (CI-NEB) modification \cite{henkelmanClimbingImageNudged2000,goswamiEnhancedClimbingImage2026} promotes the highest-energy image to saddle-point-seeking behavior by removing its spring force and inverting the parallel component of the true force:

\begin{equation}
\mathbf{F}_{i_{\max}}^{\text{CI}} = -\nabla V(\mathbf{R}_{i_{\max}}) + 2[\nabla V(\mathbf{R}_{i_{\max}}) \cdot \hat{\boldsymbol{\tau}}_{i_{\max}}] \hat{\boldsymbol{\tau}}_{i_{\max}}
\label{eq:cineb_force}
\end{equation}

This image minimizes energy perpendicular to the path tangent while maximizing along it, which is precisely the condition for a first-order saddle point.
Compare Eq. \ref{eq:cineb_force} to the dimer translational force (Eq. \ref{eq:modified_force} with \(C < 0\)):

\begin{equation}
\mathbf{F}^{\dagger} = \mathbf{F}(\mathbf{R}) - 2[\mathbf{F}(\mathbf{R}) \cdot \hat{\mathbf{N}}]\hat{\mathbf{N}}
\label{eq:dimer_force_repeat}
\end{equation}

The dimer method and the climbing image NEB (CI-NEB) are intricately linked.
Both employ Householder reflections of the form:

\begin{equation}\mathbf{F}^{\dagger} = \mathbf{F} - 2(\mathbf{F} \cdot
\hat{\mathbf{v}})\hat{\mathbf{v}}\label{eq:householder_generic}
\end{equation}

The only difference lies in the source of the distinguished direction \(\hat{\mathbf{v}}\):
\begin{description}
\item[{Dimer}] (Eq. \ref{eq:modified_force}): \(\hat{\mathbf{v}} = \hat{\mathbf{N}}\), the minimum mode derived directly from finite-difference curvature.
\item[{CI-NEB}] (Eq. \ref{eq:cineb_force}): \(\hat{\mathbf{v}} = \hat{\boldsymbol{\tau}}_{i_{\max}}\), the path tangent estimated from neighboring images.
\end{description}

This structural identity admits a precise geometric interpretation.
The dimer (two images symmetrically displaced about a midpoint) acts as a truncated, free-ended chain.
The midpoint plays the role of the climbing image, and the two endpoints provide the curvature information that determines the climbing direction.
Extending this to a full chain of \(P\) images with fixed endpoints recovers the NEB.
Conversely, truncating the NEB to three free-ended images recovers the dimer.
Both methods converge to the exact same saddle point when the minimum mode aligns with the path tangent (\(\hat{\mathbf{N}} \approx \hat{\boldsymbol{\tau}}\)), a commonality exploited in the OCI-NEB \cite{goswamiEnhancedClimbingImage2026,goswamiEfficientExplorationChemical2025}.

Section \ref{sec:bo-framework} develops the Bayesian optimization framework that unifies GP-accelerated versions of these three methods.

Algorithm \ref{alg:neb} summarizes the NEB iteration with optional climbing image activation.
The NEB's cost is dominated by the force evaluations at each image, with typical runs requiring \(P \sim 10\) images and 20 to 50 iterations.
This makes NEB a natural candidate for GP acceleration, as we discuss in Section \ref{sec:gprneb}.

\begin{algorithm}[H]
\caption{Nudged elastic band with climbing image}\label{alg:neb}
\begin{algorithmic}[1]
\Require Initial chain $\{\mathbf{R}_0, \ldots, \mathbf{R}_P\}$, spring constant $k$, CI threshold $\epsilon_{\text{CI}}$, convergence tolerance $\epsilon_{\text{tol}}$
\State CI $\gets$ \textbf{false}
\Repeat
  \State Evaluate $V(\mathbf{R}_i), \mathbf{F}(\mathbf{R}_i)$ at all movable images \Comment{$P-1$ oracle calls}
  \For{each image $i = 1, \ldots, P-1$}
    \State Compute tangent $\hat{\boldsymbol{\tau}}_i$ (Eq.~\ref{eq:improved_tangent})
    \If{CI is \textbf{true} and $i = i_{\max}$}
      \State $\mathbf{F}_i \gets -\nabla V + 2(\nabla V \cdot \hat{\boldsymbol{\tau}}_i)\hat{\boldsymbol{\tau}}_i$ \Comment{Eq.~\ref{eq:cineb_force}}
    \Else
      \State $\mathbf{F}_i^{\perp} \gets -\nabla V + (\nabla V \cdot \hat{\boldsymbol{\tau}}_i)\hat{\boldsymbol{\tau}}_i$ \Comment{Eq.~\ref{eq:perp_force}}
      \State $F_i^{s} \gets k(|\mathbf{R}_{i+1} - \mathbf{R}_i| - |\mathbf{R}_i - \mathbf{R}_{i-1}|)$ \Comment{Eq.~\ref{eq:spring_force}}
      \State $\mathbf{F}_i \gets \mathbf{F}_i^{\perp} + F_i^{s}\,\hat{\boldsymbol{\tau}}_i$
    \EndIf
  \EndFor
  \State Update all images $\mathbf{R}_i \gets \mathbf{R}_i + \Delta t\,\mathbf{F}_i$
  \If{CI is \textbf{false} and $\max_i \|\mathbf{F}_i\| < \epsilon_{\text{CI}}$}
    \State CI $\gets$ \textbf{true}; $i_{\max} \gets \arg\max_i V(\mathbf{R}_i)$
  \EndIf
\Until{$\max_i \|\mathbf{F}_i\| < \epsilon_{\text{tol}}$}
\State \Return converged MEP $\{\mathbf{R}_0, \ldots, \mathbf{R}_P\}$
\end{algorithmic}
\end{algorithm}
\section{Gaussian Process Regression}
\label{sec:gpr}
\subsection{What the Surrogate Must Provide}
\label{sec:gpr-surrogate}
The surrogate model must interpolate a small, incrementally growing set of energy and force data, provide a posterior-variance signal that shrinks where the model has seen data and grows elsewhere (noting that this signal measures sampling density in the kernel geometry, not accuracy against the true PES), and remain cheap enough that fitting and querying costs far less than the electronic structure call it replaces.
Gaussian process regression satisfies all three.
This section develops the GP from the perspective of building and using the surrogate, covering what quantities need to be computed, how they connect to the physics, and where the bottlenecks arise.
We refer readers to Rasmussen and Williams \cite{rasmussenGaussianProcessesMachine2006} and Gramacy \cite{gramacySurrogatesGaussianProcess2020} for the mathematical foundations.
As noted earlier, Deringer, Bartok, and Csanyi \cite{deringerGaussianProcessRegression2021} has a detailed review of GPR in atomistic simulation from a global MLIP view, including structural descriptors (SOAP, ACE), sparse approximations, and validation methodology.
That review treats the GP as a tool for building global machine-learned potentials from large databases; the present treatment focuses on the complementary regime of local surrogates built on the fly from tens of data points.
The key distinction is hyperparameter management: MLIP approaches optimize hyperparameters once on a large training set and fix them, while the local GP re-optimizes at every step as the training set grows, requiring trust regions and active data selection to maintain stability.

The PES is modeled as a Gaussian process, which means that the energy values at any finite collection of configurations follow a multivariate normal (MVN) distribution \cite{gramacySurrogatesGaussianProcess2020}.
The correlations between configurations are encoded in a kernel \(k(\mathbf{x}, \mathbf{x}')\), and the prior mean is set to zero (the constant kernel offset absorbs the baseline energy):

\begin{equation}
f(\mathbf{x}) \sim \mathcal{GP}\bigl(0,\, k(\mathbf{x}, \mathbf{x}')\bigr)
\label{eq:gp_prior}
\end{equation}

\begin{figure}[htbp]
\centering
\includegraphics[width=\textwidth]{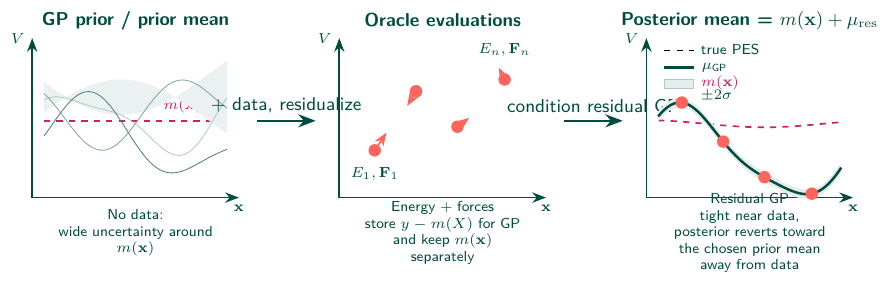}
\caption{\label{fig:gp_prior_posterior}GP conditioning in three panels. (\emph{Left}) Before any data, the prior (Eq. \ref{eq:gp_prior}) admits a wide family of smooth functions around a chosen mean function. (\emph{Center}) Oracle evaluations supply energies and forces at selected configurations. (\emph{Right}) Conditioning on the data collapses the posterior near training points while preserving wide uncertainty elsewhere; the posterior mean serves as the surrogate surface \(V_{\text{GP}}\). The optional nonzero prior-mean branch shown in the schematic is one later extension point in the design space, not the main implementation thread of this tutorial.}
\end{figure}

Before seeing any data, the PES is assumed to be drawn from a distribution over functions whose smoothness and amplitude are governed entirely by \(k\).
For molecular PES this assumption has a theoretical justification beyond convenience.
Near a pronounced global minimum, the vibrational degrees of freedom contribute additively to the potential energy, and by the central limit theorem these many contributions lead the PES on a random coordinate frame to appear approximately Gaussian \cite{madsenApproximateFactorizationMolecular1997,goswamiEfficientExplorationChemical2025}.
Thus the GP prior forms a reasonable model for the local structure of the PES in the neighborhoods that matter for saddle point searches.
Figure \ref{fig:gp_prior_posterior} summarizes the prior, data-acquisition, and posterior-conditioning stages of this regression problem.

After observing \(M\) data points \(\mathbf{y} = [y_1, \ldots, y_M]^T\) at inputs \(\mathbf{X} = [\mathbf{x}_1, \ldots, \mathbf{x}_M]\), the joint distribution of the training observations and a new query point \(\mathbf{x}_*\) is written as a single MVN.
Let \([\mathbf{K}]_{ij} = k(\mathbf{x}_i, \mathbf{x}_j)\) be the kernel matrix over training points, \([\mathbf{k}_*]_i = k(\mathbf{x}_i, \mathbf{x}_*)\) the cross-covariance with the query, and \(k_{**} = k(\mathbf{x}_*, \mathbf{x}_*)\):

\begin{equation}
\begin{bmatrix} \mathbf{y} \\ f(\mathbf{x}_*) \end{bmatrix}
\sim \mathcal{N}\left(\mathbf{0},\,
\begin{bmatrix} \mathbf{K} + \sigma_n^2\mathbf{I} & \mathbf{k}_* \\
\mathbf{k}_*^T & k_{**} \end{bmatrix}\right)
\label{eq:joint_mvn}
\end{equation}

Conditioning this joint distribution on the observed values \(\mathbf{y}\) gives the predictive distribution at \(\mathbf{x}_*\), which is again Gaussian with mean and variance:

\begin{align}
\bar{f}(\mathbf{x}_*) &= \mathbf{k}_*^T (\mathbf{K} + \sigma_n^2 \mathbf{I})^{-1} \mathbf{y} \label{eq:gp_mean} \\
\text{var}[f(\mathbf{x}_*)] &= k_{**} - \mathbf{k}_*^T (\mathbf{K} + \sigma_n^2 \mathbf{I})^{-1} \mathbf{k}_* \label{eq:gp_var}
\end{align}

The posterior mean (Eq. \ref{eq:gp_mean}) coincides with the kernel ridge regression (KRR) estimator with regularization \(\sigma_n^2\) \cite{rasmussenGaussianProcessesMachine2006}, but the GP additionally provides the predictive variance (Eq. \ref{eq:gp_var}), which is the basis for the acquisition criterion in Section \ref{sec:gprneb}.
Before proceeding, it is worth stating clearly what the predictive variance \emph{is} and what it is \emph{not}, because the distinction governs how every acquisition rule in this review should be read.
Inspecting Eq. \ref{eq:gp_var}, \(\sigma^2(\mathbf{x}_*) = k(\mathbf{x}_*,\mathbf{x}_*) - \mathbf{k}_*^{\top}\mathbf{K}^{-1}\mathbf{k}_*\), the formula depends only on the kernel, the training locations, and the hyperparameters; the observed energies \(\mathbf{y}\) enter only through the posterior mean, not through the variance.
By construction, \(\sigma^2\) collapses to the noise floor at every training point and shrinks monotonically in a kernel-dependent neighbourhood of those points as the dataset grows.
A low \(\sigma^2\) therefore reports that \(\mathbf{x}_*\) is close to the training set \emph{in the geometry that the kernel defines}; it does not compare the posterior mean against the true PES and is not a calibrated measure of accuracy.

The spline analogy due to Wahba \cite{wahbaSplineModelsObservational1990} sharpens the point.
A GP with kernel \(k\) and noise \(\sigma_n^2\) has the same MAP predictor as a smoothing spline that minimizes \(\sum_i (y_i - f(\mathbf{x}_i))^2 / \sigma_n^2 + \|f\|_{\mathcal{H}_k}^2\), where \(\|\cdot\|_{\mathcal{H}_k}\) is the reproducing-kernel Hilbert-space norm induced by \(k\).
Viewed through that lens, \(\sigma^2(\mathbf{x}_*)\) is a \emph{kernel-space interpolation radius} around \(\mathbf{x}_*\): the quantity a spline-theorist would write down to measure how close \(\mathbf{x}_*\) is to the scattered data locations \(\{\mathbf{x}_i\}\).
It is exactly as informative as the following heuristic: pass a cubic spline through a cluster of three closely spaced points and a second cluster three Angstroms away; the spline self-report of "error" is small everywhere inside each cluster and between consecutive knots, but the fit to a function that does not actually live in the spline's smoothness class can be arbitrarily bad in the gaps.
The GP inherits the same limitation.
Accuracy against the true PES is the unknown the search is trying to resolve; it is simply not part of the quantities the GP computes until the oracle is called.

For a per-search GP this gap is sharper still, because the kernel length scales and signal variance are fit from the same handful of observations whose variance we then compute.
The resulting signal is self-referential: sampling more drives \(\sigma^2\) down by construction, independently of whether the mean has converged to the truth.
The active learning criteria built on \(\sigma^2\) in Sections \ref{sec:gprdimer} and \ref{sec:gprneb} are therefore best read as sampling-density signals that complement, rather than replace, geometric safeguards like the trust region.

The mean is the surrogate's prediction; the variance measures how much information the training set carries about \(\mathbf{x}_*\).
Near observed data, the variance drops to the noise floor \(\sigma_n^2\).
Far from observed data, it approaches the prior variance \(k_{**}\).
This variance structure is the basis for the acquisition criterion: the point of highest variance is, in the GP's own assessment, the most informative place to sample next.

\begin{figure}[htbp]
\centering
\includegraphics[width=\textwidth]{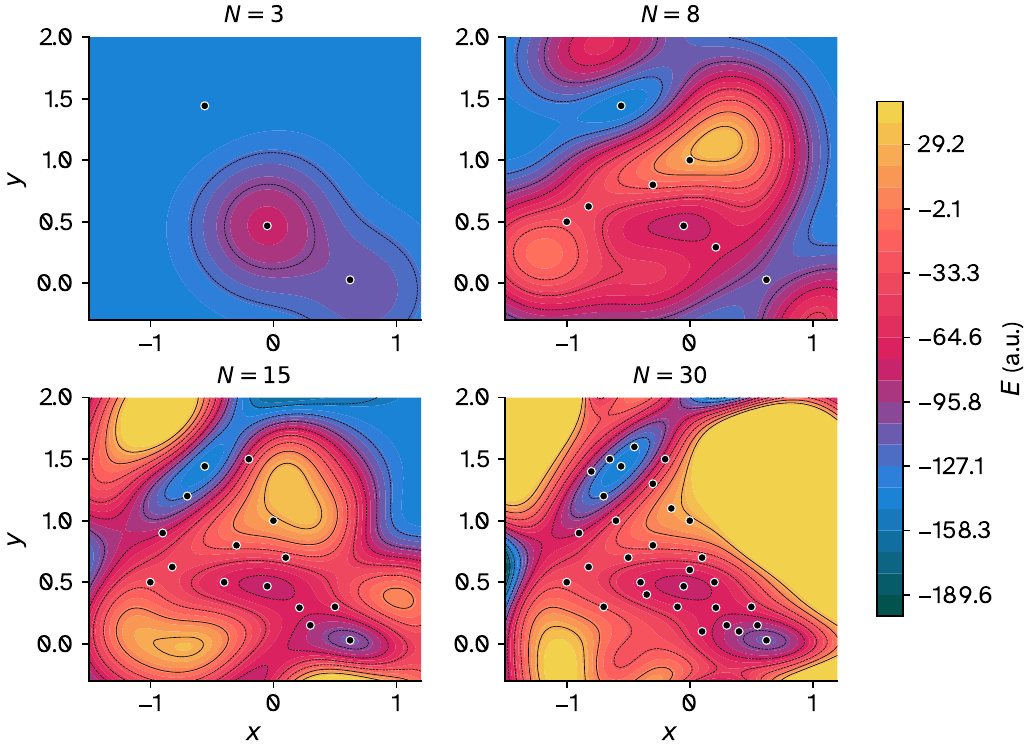}
\caption{\label{fig:mb_gp_progression}GP surrogate fidelity as a function of training set size on the Muller-Brown surface. Each panel shows the GP posterior mean contours after training on \(N = 3, 8, 15, 30\) Latin hypercube-sampled configurations (white markers). With three points the surrogate captures only crude basin structure; by 30 points the contours closely match the true PES (Figure \ref{fig:mb_neb}) in the sampled region.}
\end{figure}

\begin{figure}[htbp]
\centering
\includegraphics[width=\textwidth]{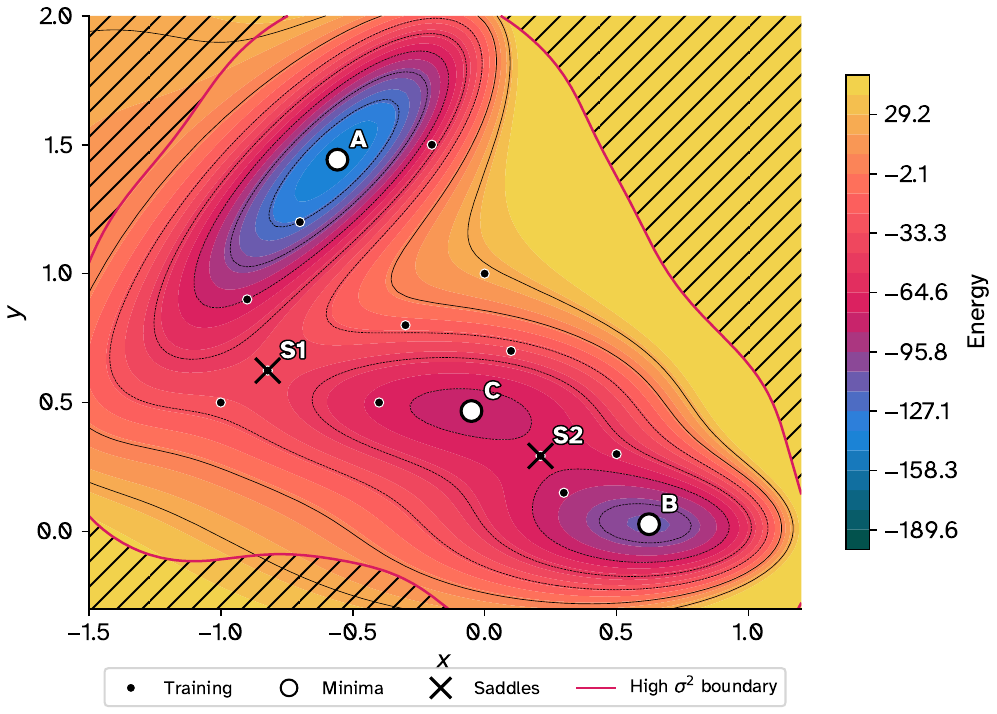}
\caption{\label{fig:mb_variance}GP predictive variance on the Muller-Brown surface after 20 training evaluations clustered near minimum A and saddle S1 (black dots). The variance is near zero close to training data and grows with distance, reaching a maximum (coral diamond) in the unexplored region. This landscape is a map of sampling density in the kernel geometry, not of accuracy against the true surface: it tells the active-learning loop where the surrogate has seen least data, which is the quantity an acquisition rule needs, but a low-variance prediction does not by itself imply a small error.}
\end{figure}

The interaction between these two quantities during a search is illustrated by the following progression.
In the first few iterations, the GP has little data and the variance is large everywhere except at the evaluated configurations (Figure \ref{fig:mb_gp_progression}, top-left panel).
The surrogate prediction is correspondingly uncertain, and the trust region (Section \ref{sec:trust-regions}) constrains the optimizer to small steps.
As data accumulates, the variance shrinks in the neighborhood of the reaction path (Figure \ref{fig:mb_variance}) and the surrogate becomes a faithful replica of the true PES in that local region (Figure \ref{fig:mb_gp_progression}, bottom panels).
The optimizer can now take longer steps on the cheap surrogate, and the active learning criterion directs the next expensive evaluation to the frontier where the variance is still large.
This feedback between uncertainty, data acquisition, and optimization step length can yield factors-of-several, and in favorable saddle-search regimes roughly ten-fold, reductions in electronic structure calls in the production papers \cite{goswamiEfficientImplementationGaussian2025,goswamiAdaptivePruningIncreased2025}.
These larger gains depend on three conditions: an oracle that dominates the per-call cost (so amortized GP overhead remains negligible), an initial configuration far enough from the target that the inner surrogate-driven steps replace many true-PES steps, and access to analytical forces (which provide \(3N+1\) data points per call rather than one).
Minimization near a quadratic basin gains less, because L-BFGS already converges in few steps; saddle searches with steep, anisotropic regions gain the most.
Section \ref{sec:gprmin} revisits this trade-off quantitatively for the LEPS surface.

Both expressions involve the matrix inverse \((\mathbf{K} + \sigma_n^2 \mathbf{I})^{-1}\), which is never formed explicitly.
Instead, the Cholesky factorization \(\mathbf{K} + \sigma_n^2 \mathbf{I} = \mathbf{L}\mathbf{L}^T\) is computed once at \(\mathcal{O}(M^3)\) cost, and the weight vector \(\boldsymbol{\alpha} = (\mathbf{K} + \sigma_n^2 \mathbf{I})^{-1}\mathbf{y}\) is obtained by forward-back substitution against \(\mathbf{L}\):

\begin{equation}
\mathbf{L}\mathbf{z} = \mathbf{y}, \quad \mathbf{L}^T \boldsymbol{\alpha} = \mathbf{z}
\label{eq:cholesky_solve}
\end{equation}

Each new prediction then costs \(\mathcal{O}(M^2)\) for the matrix-vector product \(\mathbf{k}_*^T \boldsymbol{\alpha}\).
For added robustness the implementation in \texttt{chemgp-core} uses a Cholesky factorization wrapped in a guarded routine that applies exponentially increasing jitter when the matrix is nearly singular, starting at \(10^{-8} \max(\text{diag}(\mathbf{K}))\) and increasing by a factor of 10 per attempt.
This adaptive jitter handles the rank deficiency that arises naturally from molecular kernels, where the feature space dimension (number of atom pairs) can be smaller than the coordinate dimension \(3N\).
With \(M \sim 30\), the factorization is instantaneous; the cost only becomes relevant when derivative observations are included (Section \ref{sec:gpr-derivatives}), which inflate the effective training set size.
\subsection{Regression with Derivative Observations}
\label{sec:gpr-derivatives}
Every electronic structure evaluation returns not just the energy but also the atomic forces (the negative gradient of the PES) at negligible extra cost.
For a system of \(N\) atoms, each evaluation therefore provides \(1 + 3N\) scalar constraints on the PES: one energy and \(3N\) force components.
A 10-atom system yields 31 constraints per call, so \(M = 30\) evaluations already give 930 independent observations, enough to pin down a local region of the PES with high fidelity.
Training the GP on energies alone would discard all but \(1/(1+3N)\) of this information \cite{solakDerivativeObservationsGaussian2002}, requiring an impractically large number of evaluations to achieve the same coverage.
The GP accommodates derivative observations naturally because differentiation is a linear operation on the kernel \cite{kochenderferAlgorithmsOptimization}:

\begin{align}
\text{cov}\left[f(\mathbf{x}),\, \frac{\partial f}{\partial x'_j}\right] &= \frac{\partial k(\mathbf{x}, \mathbf{x}')}{\partial x'_j} \label{eq:cov_f_df} \\[6pt]
\text{cov}\left[\frac{\partial f}{\partial x_i},\, \frac{\partial f}{\partial x'_j}\right] &= \frac{\partial^2 k(\mathbf{x}, \mathbf{x}')}{\partial x_i \partial x'_j} \label{eq:cov_df_df}
\end{align}

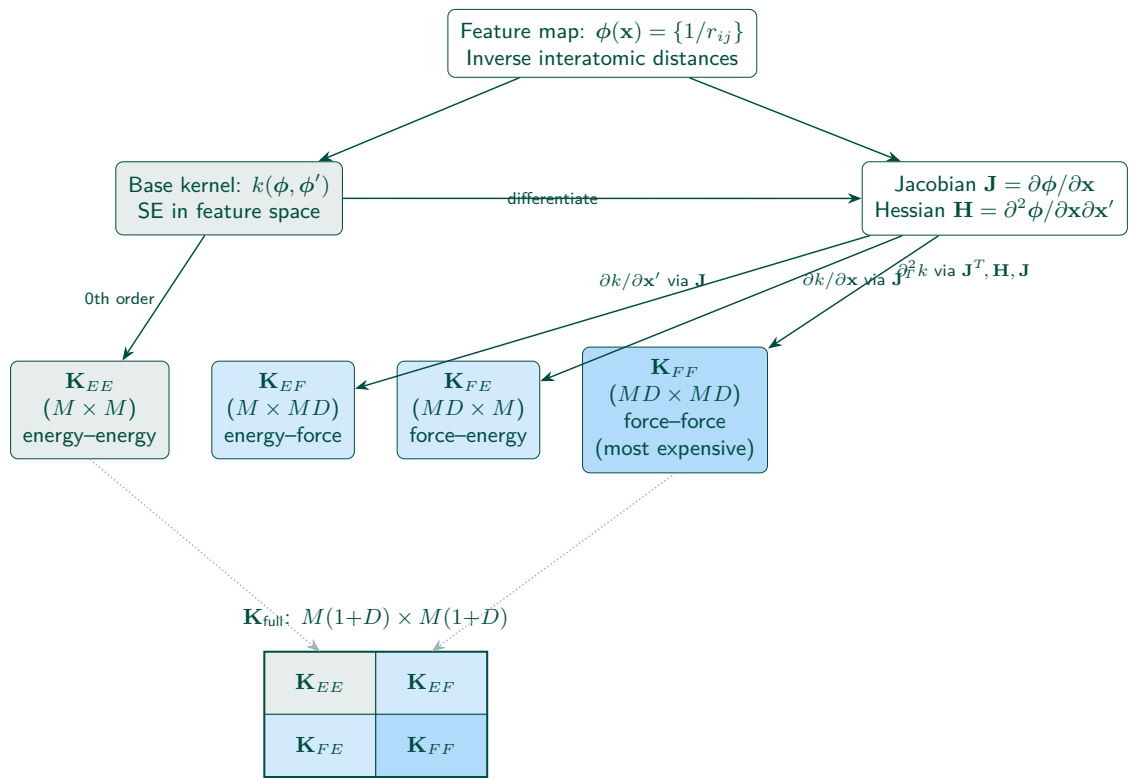
\begin{figure}[H]
\centering
\adjustbox{max width=0.9\textwidth, max totalheight=0.85\textheight}{%
\begin{tikzpicture}[
  >=Stealth,
  node distance=8mm and 10mm,
  box/.style={draw=RuhiTeal, fill=white, rounded corners=3pt,
              text=RuhiTeal, minimum height=2.4em, align=center,
              font=\small\sffamily, inner sep=5pt},
  arr/.style={->, RuhiTeal, semithick},
  lbl/.style={font=\scriptsize\sffamily, text=RuhiTeal},
]
\node[box] (phi) {Feature map: $\boldsymbol{\phi}(\mathbf{x}) = \{1/r_{ij}\}$\\Inverse interatomic distances};

\node[box, fill=RuhiTeal!10, below left=12mm and 15mm of phi] (kbase)
  {Base kernel: $k(\boldsymbol{\phi}, \boldsymbol{\phi}')$\\SE in feature space};
\node[box, below right=12mm and 15mm of phi] (jac)
  {Jacobian $\mathbf{J} = \partial\boldsymbol{\phi}/\partial\mathbf{x}$\\%
   Hessian $\mathbf{H} = \partial^{2}\boldsymbol{\phi}/\partial\mathbf{x}\partial\mathbf{x}'$};

\node[box, fill=RuhiTeal!10, below=18mm of kbase, xshift=-20mm] (kee)
  {$\mathbf{K}_{EE}$\\($M \times M$)\\energy--energy};
\node[box, fill=RuhiBlue!20, right=6mm of kee] (kef)
  {$\mathbf{K}_{EF}$\\($M \times MD$)\\energy--force};
\node[box, fill=RuhiBlue!20, right=6mm of kef] (kfe)
  {$\mathbf{K}_{FE}$\\($MD \times M$)\\force--energy};
\node[box, fill=RuhiBlue!35, right=6mm of kfe] (kff)
  {$\mathbf{K}_{FF}$\\($MD \times MD$)\\force--force\\(most expensive)};

\coordinate (midblock) at ($(kef.south)!0.5!(kfe.south)$);
\node[below=20mm of midblock, anchor=north] (asmtitle)
  {\sffamily\small\color{RuhiTeal}%
   $\mathbf{K}_{\text{full}}$: $M(1{+}D) \times M(1{+}D)$};

\def\bw{1.6cm}   %
\def\bh{0.9cm}   %
\coordinate (gridtl) at ($(asmtitle.south)+(-\bw,-2mm)$);

\fill[RuhiTeal!10] (gridtl) rectangle +(\bw,-\bh);
\draw[RuhiTeal] (gridtl) rectangle +(\bw,-\bh);
\node[font=\small\sffamily, text=RuhiTeal] at ($(gridtl)+(0.5*\bw,-0.5*\bh)$)
  {$\mathbf{K}_{EE}$};

\fill[RuhiBlue!20] ($(gridtl)+(\bw,0)$) rectangle +(\bw,-\bh);
\draw[RuhiTeal] ($(gridtl)+(\bw,0)$) rectangle +(\bw,-\bh);
\node[font=\small\sffamily, text=RuhiTeal] at ($(gridtl)+(1.5*\bw,-0.5*\bh)$)
  {$\mathbf{K}_{EF}$};

\fill[RuhiBlue!20] ($(gridtl)+(0,-\bh)$) rectangle +(\bw,-\bh);
\draw[RuhiTeal] ($(gridtl)+(0,-\bh)$) rectangle +(\bw,-\bh);
\node[font=\small\sffamily, text=RuhiTeal] at ($(gridtl)+(0.5*\bw,-1.5*\bh)$)
  {$\mathbf{K}_{FE}$};

\fill[RuhiBlue!35] ($(gridtl)+(\bw,-\bh)$) rectangle +(\bw,-\bh);
\draw[RuhiTeal] ($(gridtl)+(\bw,-\bh)$) rectangle +(\bw,-\bh);
\node[font=\small\sffamily, text=RuhiTeal] at ($(gridtl)+(1.5*\bw,-1.5*\bh)$)
  {$\mathbf{K}_{FF}$};

\draw[RuhiTeal, thick] (gridtl) rectangle +(2*\bw,-2*\bh);

\draw[arr] (phi) -- node[lbl, left] {} (kbase);
\draw[arr] (phi) -- node[lbl, right] {} (jac);
\draw[arr] (kbase) -- node[lbl, left] {0th order} (kee);
\draw[arr] (kbase) -- node[lbl, right, pos=0.3] {differentiate} (jac);
\draw[arr] (jac) -- node[lbl, left, pos=0.3] {$\partial k/\partial\mathbf{x}'$ via $\mathbf{J}$} (kef);
\draw[arr] (jac) -- node[lbl, right, pos=0.3] {$\partial k/\partial\mathbf{x}$ via $\mathbf{J}^{T}$} (kfe);
\draw[arr] (jac) -- node[lbl, right, pos=0.3]
  {$\partial^{2}k$ via $\mathbf{J}^{T}\!,\mathbf{H},\mathbf{J}$} (kff);

\draw[arr, densely dotted, RuhiTeal!40] (kee.south) -- ($(gridtl)+(0.5*\bw,0)$);
\draw[arr, densely dotted, RuhiTeal!40] (kff.south) -- ($(gridtl)+(1.5*\bw,0)$);
\end{tikzpicture}%
}%
\caption{Block structure of the full covariance matrix $K_{\text{full}}$. The base kernel in feature space generates four Cartesian-space blocks through differentiation via the feature Jacobian J. Darker shading indicates higher computational cost.}
\label{fig:kernel_blocks}
\end{figure}

In the implementation, the kernel matrix acquires a \(2 \times 2\) block structure over the energy and force observations:

\begin{equation}
\mathbf{K}_{\text{full}} = \begin{bmatrix}
\mathbf{K}_{EE} & \mathbf{K}_{EF} \\
\mathbf{K}_{FE} & \mathbf{K}_{FF}
\end{bmatrix}
\label{eq:full_covariance}
\end{equation}

where the blocks are the energy-energy (\(M \times M\)), energy-force (\(M \times MD\)), and force-force (\(MD \times MD\)) covariances with \(D = 3N\). The full matrix is \(M(1+D) \times M(1+D)\), and the Cholesky cost becomes \(\mathcal{O}(M^3 D^3)\). Figure \ref{fig:kernel_blocks} shows this block structure schematically. As a concrete example, a 10-atom molecule with \(M = 30\) accumulated configurations gives a \(930 \times 930\) matrix.
This is still fast, but growth is cubic, which is why the data management strategies in Section \ref{sec:otgp} become necessary for longer searches.

Energies and forces have different magnitudes and units, so separate noise variances \(\sigma_E^2\) and \(\sigma_F^2\) are assigned to each block.
Because the electronic structure data is deterministic (no stochastic noise), these are not physical noise parameters but Tikhonov regularizers; they are set to small values (\(\sim 10^{-8}\)) to keep the matrix well-conditioned.
The ratio \(\sigma_E^2 / \sigma_F^2\) controls the relative weight the GP places on matching energies versus forces, and incorrect specification of this ratio degrades surrogate quality.

Including forces provides substantial payoff.
Each evaluation contributes \(1 + 3N\) scalar constraints, so \(M = 30\) calls for a 10-atom system yield 930 constraints, enough to resolve the PES locally without needing a large training set.
This information density is the core reason the local surrogate strategy works with so few evaluations.
It also imposes a stringent requirement on the kernel implementation, namely that the derivative blocks (Eqs. \ref{eq:cov_f_df}--\ref{eq:cov_df_df}) must be computed analytically, as discussed in detail in Section \ref{sec:idist-kernel}.
The inverse-distance kernel provides the required invariance, but the composition of the inverse, the norm, and the exponential makes it particularly sensitive to numerical noise in the derivative blocks, and the production C++ code (gpr$\backslash$\textsubscript{optim}) is heavily optimized around this bottleneck.
\subsection{Covariance Functions for Molecular Systems}
\label{sec:kernels}
The kernel encodes the assumption about which configurations should have similar energies.
If \(k(\mathbf{x}, \mathbf{x}')\) is large, the GP expects the energies at \(\mathbf{x}\) and \(\mathbf{x}'\) to be correlated, and it will interpolate smoothly between them; if \(k\) is small, the GP treats them as independent.
For molecular systems, the kernel must respect the physical symmetries of the PES, namely rotational and translational invariance, and ideally also permutation invariance for identical atoms.
A kernel operating directly on Cartesian coordinates \(\mathbf{x} \in \mathbb{R}^{3N}\) fails the first requirement immediately, because rotating all atoms changes \(\mathbf{x}\) but not \(V(\mathbf{x})\), so two identical configurations related by a rigid rotation would appear dissimilar to the GP.

Global MLIP frameworks solve this with high-dimensional structural descriptors that project the atomic environment onto a rotationally invariant representation.
SOAP (Smooth Overlap of Atomic Positions) \cite{bartokRepresentingChemicalEnvironments2013} constructs a local neighbor density around each atom, expands it in a radial-spherical basis, and forms the power spectrum, a descriptor that is automatically invariant to rotations and permutations of like atoms.
The body-order interpretation is illuminating.
A linear SOAP kernel is a three-body model (each descriptor entry involves a central atom and a pair of neighbors), and raising the kernel to the power \(\zeta\) yields a \((2\zeta+1)\)-body model \cite{deringerGaussianProcessRegression2021}.
The ACE (Atomic Cluster Expansion) \cite{drautzAtomicClusterExpansion2019} framework generalizes this construction to arbitrary body orders in a systematic manner.
These descriptors are engineered to resolve fine structural differences across all of configuration space, with convergence parameters (radial and angular truncation orders, cutoff radius) that control the trade-off between accuracy and cost.

For the \emph{local} surrogates discussed in this work \cite{fdez.galvanRestrictedVarianceConstrainedReaction2021,larsenMachinelearningenabledOptimizationAtomic2023,vishartAcceleratingCatalysisSimulations,koistinenAlgorithmsFindingSaddle2019,goswamiEfficientExplorationChemical2025}, that level of sophistication is unnecessary and carries a cost that defeats the purpose.
The GP only needs to distinguish configurations in a small neighborhood of the reaction path, where the molecular connectivity does not change and pairwise distance information captures the relevant variation.
Computing SOAP descriptors and their analytic derivatives for each of the \(M \sim 30\) training points would add overhead comparable to the GP algebra itself, erasing the wall-time savings.
More fundamentally, the derivative blocks (Section \ref{sec:gpr-derivatives}) require second-order kernel derivatives with respect to Cartesian coordinates, and the composition of a high-dimensional descriptor with the kernel introduces an additional layer of chain-rule complexity that must be handled analytically to avoid numerical noise (Section \ref{sec:idist-kernel}).
The inverse-distance feature map \cite{koistinenNudgedElasticBand2019} \(\phi_{ij} = 1/r_{ij}\) is the simplest descriptor that provides rotational and translational invariance while admitting tractable analytical derivatives.
The idea of using inverse interatomic distances as molecular features has roots in the Coulomb matrix representation \cite{ruppFastAccurateModeling2012}.

A pairwise-distance representation is preferred for local surrogates.
The stationarity of the SE kernel (the assumption that the covariance depends only on the \emph{difference} between inputs) means the GP assumes uniform fluctuations across its domain.
In Cartesian coordinates, this assumption is catastrophically wrong for a PES because the energy varies slowly near a minimum but changes by electron-volts over sub-Angstrom displacements near a repulsive wall.
The GP would need an impossibly short length scale to capture the repulsive region, which would destroy its interpolation ability in the flat valley.
By transforming to inverse interatomic distances, the energy landscape is effectively \emph{preconditioned}.
The \(1/r\) map compresses the repulsive region (where \(r\) is small and \(1/r\) changes slowly in relative terms) and stretches the long-range region (where small changes in \(r\) produce large changes in \(1/r\)).
The result is a feature space where the PES has more uniform curvature, and the stationary kernel becomes a reasonable approximation.
This is the core reason the inverse-distance kernel outperforms Cartesian kernels for molecular systems, even when both are given the same training data.

The inverse-distance feature map remains well-defined for any geometry without coincident atoms, including planar molecules and linear arrangements.
The Jacobian
\begin{equation}
\frac{\partial \phi_{ij}}{\partial z^{(k)}}
  \;=\; -\,\frac{z^{(i)} - z^{(j)}}{r_{ij}^{\,3}}
        \,(\delta_{ki} - \delta_{kj})
\label{eq:invdist_jacobian_z}
\end{equation}
vanishes identically for every \((i,j,k)\) when every atom satisfies \(z^{(k)}=0\), and by the chain rule the GP out-of-plane force \(F_z^{(k)} = -\partial V_{\text{GP}}/\partial z^{(k)}\) is zero at the planar geometry independently of the GP coefficients.
This is the physically correct answer rather than a defect: a perpendicular displacement of magnitude \(\delta z\) gives \(r_{ij}(\delta z) = r_{ij}(0) + \mathcal{O}(\delta z^{2})\), so the energy is genuinely stationary in the symmetry direction, and any function that depends on \(\mathbf{x}\) only through \(\{r_{ij}\}\) has identically zero gradient along the planar orbit.
The instant any atom is perturbed off the plane \(z^{(i)} - z^{(j)}\) is generically nonzero, the out-of-plane Jacobian block recovers full rank, and the GP regains sensitivity to all three Cartesian directions.
\subsubsection{The Inverse-Distance Squared Exponential Kernel}
\label{sec:idist-kernel}
The solution is to work with internal features that are inherently invariant.
The inverse interatomic distance provides a physically motivated feature:

\begin{equation}
\phi_{ij}(\mathbf{x}) = \frac{1}{r_{ij}(\mathbf{x})} = \frac{1}{\sqrt{\sum_{d=1}^{3}(x_{i,d} - x_{j,d})^2}}
\label{eq:invdist_feature}
\end{equation}

\begin{figure}[htbp]
\centering
\includegraphics[width=\textwidth]{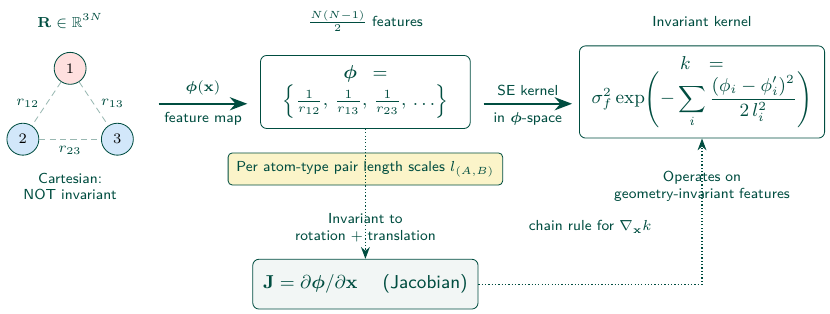}
\caption{\label{fig:invdist_feature_map}The inverse-distance feature map. Cartesian coordinates (\(\mathbb{R}^{3N}\), not invariant) are mapped to pairwise inverse distances (\(N(N{-}1)/2\) features, invariant to rotation and translation). The SE kernel operates in this feature space. The Jacobian \(\mathbf{J} = \partial\boldsymbol{\phi}/\partial\mathbf{x}\) propagates through the kernel via the chain rule to produce the derivative blocks needed for force predictions.}
\end{figure}

The inverse-distance squared exponential (SE) kernel is then:

\begin{equation}
k(\mathbf{x}, \mathbf{x}') = \sigma_c^2 + \sigma_f^2 \exp\left(-\frac{1}{2} \sum_{i} \sum_{j > i} \left(\frac{\phi_{ij}(\mathbf{x}) - \phi_{ij}(\mathbf{x}')}{l_{\phi(i,j)}}\right)^2 \right)
\label{eq:idist_kernel}
\end{equation}

where:

\begin{itemize}
\item \(\sigma_f^2\) is the signal variance, controlling the amplitude of the GP prior.
\item \(\sigma_c^2\) is a constant offset, accounting for the mean energy level.
\item \(l_{\phi(i,j)}\) are length-scale parameters, one per atom-pair type \(\phi(i,j)\), controlling how rapidly the covariance decays as the inverse distances change.
\end{itemize}

Figure \ref{fig:invdist_feature_map} shows how the Cartesian coordinates enter this kernel through the inverse-distance feature map and its Jacobian.

The \(1/r_{ij}\) feature has three properties that matter for the GP.
First, it is invariant under rigid-body motions, so the covariance between two configurations is unaffected by how they are oriented in the lab frame.
Second, and more subtle, the divergence as \(r_{ij} \to 0\) creates a natural barrier in feature space: two configurations where any atom pair has a markedly different close-contact distance are mapped to widely separated points in the inverse-distance representation.
The GP, which interpolates smoothly in feature space, cannot interpolate \emph{through} this barrier.
This means the surrogate will never predict a smooth, low-energy path through a repulsive wall, even when it has no training data in that region.
The divergence does the work that an explicit repulsive prior would otherwise have to do.
Third, the number of features \(N_{\text{pairs}} = N(N-1)/2\) is fixed for a given molecular formula regardless of the spatial arrangement, so the kernel is always well-defined.
This is a practical advantage over radial-cutoff descriptors, where the number of neighbors within a fixed radius can vary between configurations, creating a dimension mismatch that requires padding or variable-length handling.

The length-scale parameters \(l_{\phi(i,j)}\) control the GP's sensitivity to changes in each interatomic distance.
A short length scale for a particular atom pair means the GP treats small changes in that pair's inverse distance as significant (i.e., the pair is "stiff" in the model's view); a long length scale means the GP is insensitive to that pair.
In practice, the hyperparameter optimization (Section \ref{sec:hyperparameters}) learns these from the data, and bonds that are actively breaking or forming during the reaction acquire short length scales, while spectator bonds that barely change acquire long ones.
This automatic relevance determination is what allows the GP to focus its limited training data on the degrees of freedom that matter for the particular transition being studied.

The constant offset \(\sigma_c^2\) is fixed rather than optimized alongside the other hyperparameters, since with the small training sets typical of on-the-fly searches (\(M \sim 10\textrm{--}50\)), the marginal likelihood cannot reliably distinguish \(\sigma_c^2\) from \(\sigma_f^2\).
A default of \(\sigma_c^2 = 1.0\) works well for molecular systems with eV-scale energies; for 2D model surfaces (LEPS, Muller-Brown) where energies are already centered near zero, \(\sigma_c^2 = 0.0\) avoids introducing a superfluous degree of freedom.
Without it, the GP prior mean is zero, and the posterior mean would revert to zero far from the training data.
The constant kernel adds a baseline covariance that is independent of configuration, which allows the GP to represent a nonzero mean energy level.
In practice this absorbs the large absolute energy common in electronic structure calculations, so the GP only needs to model the \emph{relative} energy variations.
The constant kernel carries zero derivative with respect to any coordinate, so \(\sigma_c^2\) drops out of every derivative block of the covariance matrix and out of the cross-covariance vector \(\mathbf{k}_*\) used for force prediction; the analytical force expression contains no explicit \(\sigma_c^2\) term.
The decoupling stops there.
The GP weights come from a single joint solve over the full covariance matrix, so changing \(\sigma_c^2\) shifts the conditioning of that solve and rebalances the energy and force residuals it minimizes; an inappropriate value can still perturb the predicted forces indirectly.
The defaults above keep this indirect effect small in practice.

The kernel derivative blocks needed for Eq. \ref{eq:full_covariance} are obtained by applying the chain rule through the feature map:

\begin{equation}
\frac{\partial k}{\partial x_a} = \sum_{(i,j)} \frac{\partial k}{\partial \phi_{ij}} \frac{\partial \phi_{ij}}{\partial x_a}
\label{eq:kernel_chain_rule}
\end{equation}

where \(\partial \phi_{ij} / \partial x_a\) is the Jacobian of the inverse-distance features with respect to the Cartesian coordinates.
For the SE kernel, the partial derivative with respect to a feature is:

\begin{equation}
\frac{\partial k}{\partial \phi_{ij}} = -\sigma_f^2 \frac{\phi_{ij}(\mathbf{x}) - \phi_{ij}(\mathbf{x}')}{l_{\phi(i,j)}^2} \exp(\cdots)
\label{eq:kernel_feature_deriv}
\end{equation}

The second-order derivatives \(\partial^2 k / \partial x_a \partial x'_b\) follow analogously through the Hessian of the feature map.
In the implementation, the Jacobian of the inverse-distance features has the explicit form:

\begin{equation}
\frac{\partial \phi_{ij}}{\partial x_{i,a}} = -\frac{x_{i,a} - x_{j,a}}{r_{ij}^3}, \qquad
            \frac{\partial \phi_{ij}}{\partial x_{j,a}} = \frac{x_{i,a} - x_{j,a}}{r_{ij}^3}
\label{eq:invdist_jacobian}
\end{equation}

and the force-force block of the covariance matrix is assembled via the chain rule as \(\mathbf{K}_{FF} = \mathbf{J}_1^T \mathbf{H}_{\text{feat}} \mathbf{J}_2\), where \(\mathbf{H}_{\text{feat}}\) is the Hessian of the kernel in feature space and \(\mathbf{J}_1, \mathbf{J}_2\) are the Jacobians at the two configurations.

We stress that these derivatives \emph{must} be computed analytically.
Using nested automatic differentiation (e.g., dual-number propagation through the inverse-distance computation and the kernel exponential) introduces numerical noise of order \(\sim 10^{-8}\) in the force-force block.
This is the same magnitude as the Tikhonov regularizer \(\sigma_F^2\), so the assembled covariance matrix loses positive definiteness after approximately 10 training points.
The problem is intrinsic to the composition of the inverse (\(1/r\)), the Euclidean norm (\(\sqrt{\cdot}\)), and the exponential in the SE kernel, where each layer of dual-number arithmetic accumulates truncation error that the subsequent layer amplifies.
The MATLAB, Rust, and C++ implementations all use fully analytical derivatives for this reason.
In production codes \cite{goswamiAdaptivePruningIncreased2025}, the derivative computation is further optimized by pre-computing and caching the inverse-distance Jacobians, vectorizing the block assembly with Eigen array operations, and zeroing covariance entries below machine epsilon to prevent noise accumulation.
This level of optimization is necessary because the derivative block computation dominates the wall time of the GP update step.
In general the ill-conditioning due to the addition of derivative observations has been noted across disciplines \cite{chengSlicedGradientenhancedKriging,ulaganathanPerformanceStudyGradientenhanced2015}.
\subsection{Hyperparameter Optimization}
\label{sec:hyperparameters}
The GP model has a set of free parameters \(\boldsymbol{\theta} = \{\sigma_f^2, \sigma_c^2, \{l_{\phi(i,j)}\}, \sigma_E^2, \sigma_F^2\}\) that must be determined from the data, and have no connection to the bond lengths \cite{goswamiAdaptivePruningIncreased2025}.
We optimize these by maximizing the log marginal likelihood:

\begin{equation}
\log p(\mathbf{y} \mid \mathbf{X}, \boldsymbol{\theta}) = -\frac{1}{2}\mathbf{y}^T \mathbf{K}_{\boldsymbol{\theta}}^{-1} \mathbf{y} - \frac{1}{2}\log|\mathbf{K}_{\boldsymbol{\theta}}| - \frac{n}{2}\log(2\pi)
\label{eq:marginal_likelihood}
\end{equation}

where \(\mathbf{K}_{\boldsymbol{\theta}}\) is the full covariance matrix (including noise) and \(n\) is the total number of scalar observations.
The first term penalizes poor data fit, the second penalizes model complexity (large determinant), and the third is a normalization constant.
The gradient with respect to each hyperparameter is available in closed form:

\begin{equation}
\frac{\partial \log p}{\partial \theta_j} = \frac{1}{2} \text{tr}\left[\left(\boldsymbol{\alpha}\boldsymbol{\alpha}^T - \mathbf{K}_{\boldsymbol{\theta}}^{-1}\right) \frac{\partial \mathbf{K}_{\boldsymbol{\theta}}}{\partial \theta_j}\right]
\label{eq:mll_gradient}
\end{equation}

where \(\boldsymbol{\alpha} = \mathbf{K}_{\boldsymbol{\theta}}^{-1}\mathbf{y}\).
Maximizing the MLL is equivalent to computing the maximum a posteriori (MAP) estimate of the hyperparameters under a flat (improper) prior.
With few training points the MLL landscape is flat or multimodal, and the MAP estimate is poorly determined.
Two failure modes result: the signal variance \(\sigma_f^2\) can grow without bound (the data-fit term in Eq. \ref{eq:marginal_likelihood} dominates the complexity penalty), and the full hyperparameter vector can oscillate between competing MLL modes as each new data point shifts the landscape.
Both pathologies produce surrogates that are unrelated to the true PES and must be regularized; Section \ref{sec:map-regularization} addresses this.

Both the Rust code for this tutorial and the production C++ code use the scaled conjugate gradient (SCG) optimizer \cite{mollerScaledConjugateGradient1993}, which exploits the analytical gradient of the marginal likelihood (Eq. \ref{eq:mll_gradient}).
The hyperparameters are re-optimized at every outer iteration, which means the marginal likelihood landscape changes as data accumulates.
This re-optimization is both the source of the GP's adaptivity and, as discussed in Section \ref{sec:map-regularization}, a potential source of instability.

The necessity of per-step hyperparameter optimization is a distinguishing feature of the local surrogate regime that sets it apart from global GP potentials.
In MLIP frameworks, especially universal models \cite{mazitovPETMADLightweightUniversal2025} the kernel operates in a descriptor space (SOAP, ACE) whose structure already encodes the relevant chemical environment.
The descriptors are designed so that the kernel length scale has a physical interpretation (the Gaussian mollification width \(\sigma_a\), typically 0.3 \AA{} for systems containing hydrogen and 0.5 \AA{} for heavier elements) that is nearly universal across chemical systems.
The practitioner fixes these hyperparameters a priori from physical reasoning and then builds the training set to achieve a target accuracy, rather than optimizing hyperparameters to match a fixed dataset.
The regularization parameter is set to the estimated noise level of the training data, and data is added until the model reaches this noise floor.
The problem is, as it were, "turned upside down": instead of fitting hyperparameters to data, one chooses hyperparameters that encode prior physical knowledge and fits the \emph{data} to the model.

This inversion is possible because the high-dimensional descriptor space absorbs most of the complexity that the hyperparameters would otherwise need to capture.
The SOAP descriptor or higher order kernels \cite{bigiWignerKernelsBodyordered2024}, for instance, encodes three-body and higher correlations through its expansion in radial and angular basis functions, so a simple stationary kernel with fixed hyperparameters suffices to interpolate smoothly in descriptor space.
By contrast, the inverse-distance kernel used here operates in a lower-dimensional pairwise feature space, and the length-scale parameters compensate for the missing higher-body information by adapting to the local PES region.
As the search moves from a minimum through a transition region to a saddle point, the effective stiffness of the PES changes, and the length scales track this change.
This is why re-optimization occurs at every step and why the hyperparameter instabilities of Section \ref{sec:map-regularization} arise: the optimization is chasing a moving target.

The effect of hyperparameter choice on the surrogate is illustrated in Figure \ref{fig:mb_hyperparams}, which shows 1D slices of the GP prediction on the Muller-Brown surface for a grid of length scale and signal variance values.
A remark on the physical interpretation of the optimized hyperparameters is warranted.
In some formulations, the length scales \(l_{\phi(i,j)}\) are expected to converge to quantities related to equilibrium bond lengths \cite{garijodelrioMachineLearningBond2020,garijodelrioLocalBayesianOptimizer2019,vishartAcceleratingCatalysisSimulations} or covalent radii, but this expectation lacks a first-principles justification \cite{goswamiAdaptivePruningIncreased2025}.
One length scale per \emph{atom-pair type} (e.g., one for all C--H pairs, one for all C--C pairs) is defined, and optimizing over all instances of that type in the training set yields a global, averaged stiffness for each interaction type that reflects the local PES region explored at the current stage of the search, not a fixed molecular property.
The signal variance \(\sigma_f^2\) similarly does not correspond to a physical energy scale but controls the flexibility of the surrogate model.
The disconnect is fundamental.
The marginal likelihood (Eq.  \ref{eq:marginal_likelihood}) is maximized, a statistical quantity that measures the model's consistency with the observed data under a Gaussian assumption, and the resulting surrogate reproduces the true PES well enough that its stationary points approximate the true ones.
The hyperparameters encode the GP's many-body effects implicitly through a few parameters per pair type, and their numerical values are best understood as model-fitting artifacts rather than physical constants.

The signal variance \(\sigma_f^2\) can be handled in multiple ways.
In this framework it is a free hyperparameter optimized by MLL, with a logarithmic barrier (Section \ref{sec:map-regularization}) to prevent divergence.
An alternative is to marginalize \(\sigma_f^2\) out under a conjugate inverse-gamma prior.
The result is a Student's t-process \cite{shahStudenttProcessesAlternatives} whose predictive mean is identical to the GP's but whose heavier-tailed variance produces more conservative uncertainty estimates; this has been applied to surrogate-accelerated NEB for surface catalysis \cite{vishartAcceleratingCatalysisSimulations}.
Other kernels (e.g., Matern \cite{denzelGaussianProcessRegression2018}) can each be made to work with appropriate tuning.
The requirements are that the surrogate is locally faithful, that the posterior variance is a usable sampling-density signal (even though it does not directly measure accuracy), and that the hyperparameters do not destabilize the model; the specific mechanism for achieving these is secondary.
\subsubsection{Data-Dependent Initialization}
\label{sec:init}
\begin{figure}[htbp]
\centering
\includegraphics[width=\textwidth]{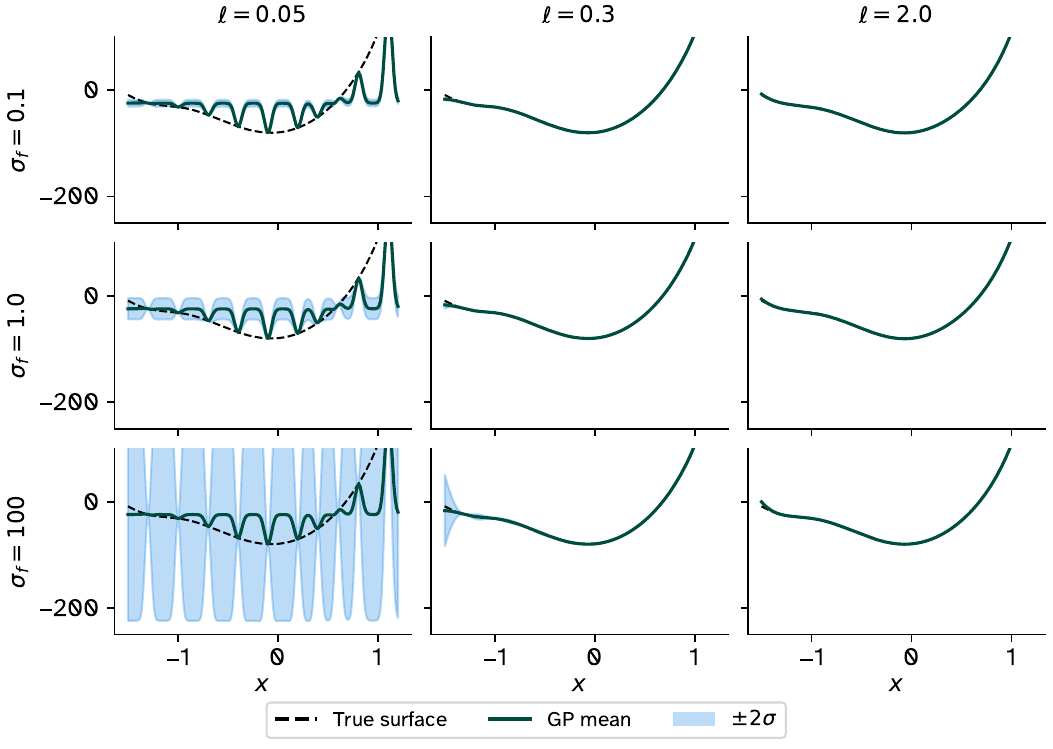}
\caption{\label{fig:mb_hyperparams}Hyperparameter sensitivity on the Muller-Brown surface. Each panel shows a 1D slice at \(y = 0.5\) with the true PES (black dashed), the GP posterior mean (teal), and the \(\pm 2\sigma\) confidence band (light blue), for nine combinations of length scale \(\ell \in \{0.05, 0.3, 2.0\}\) (columns) and signal variance \(\sigma_f \in \{0.1, 1.0, 100.0\}\) (rows). Small \(\ell\) produces noisy interpolation; large \(\ell\) over-smooths and misses barrier structure. The center cell (\(\ell = 0.3\), \(\sigma_f = 1.0\)) shows well-calibrated behavior where the confidence band tightly encloses the true surface near training data and widens appropriately in data-sparse regions.}
\end{figure}

Good initialization of the hyperparameters is critical for avoiding poor local optima.
Following Gramacy \cite{gramacySurrogatesGaussianProcess2020}, the signal variance and length scales are initialized from the data range:

\begin{align}
\sigma_f^2 &= \left(\Phi^{-1}(0.75) \cdot \frac{\text{range}(\mathbf{y})}{3}\right)^2 \label{eq:init_sigma} \\
l &= \Phi^{-1}(0.75) \cdot \frac{\text{range}(\mathbf{X})}{3} \label{eq:init_length}
\end{align}

where \(\Phi^{-1}(0.75) \approx 0.6745\) is the 75th percentile of the standard normal, and \(\text{range}(\cdot)\) denotes the data range.
This initialization places the GP in a reasonable regime where it can capture the variation in the data without overfitting.
The sensitivity to these choices is demonstrated in Figure \ref{fig:mb_hyperparams}, and the corresponding NLL landscape (Figure S3) shows how the MAP optimum balances data fit against model complexity.

\section{The Bayesian Surrogate Loop: Anatomy of the Unified Framework}
\label{sec:bo-framework}
\paragraph{Core versus extensions.}
The \emph{core} components used by every method in this review are the SE
kernel with inverse-distance features (Section~\ref{sec:gpr}), gradient
observations in the covariance matrix (Eq.~\ref{eq:full_covariance}),
farthest-point sampling for training subsets, MAP regularization of
hyperparameters, the adaptive trust radius, and the LCB inner-loop
convergence criterion (Eq.~\ref{eq:lcb_convergence}). The \emph{OT-GP
extensions} that further improve accuracy and stability for harder
problems are per-bead FPS with Earth Mover's Distance, the variance
barrier and oscillation detection, and the OTGPD adaptive inner
tolerance. A separate optional extension is the random Fourier feature
approximation (Section~\ref{sec:rff}). Section~\ref{sec:otgp} develops
each subsection by leading with the core formulation and flagging the
OT-GP refinement; readers who want the unified loop without OT-GP can
follow only the core text in each subsection.

The preceding sections develop the GP as a regression tool: given training data, it produces a posterior mean (the surrogate surface) and a posterior variance (the uncertainty).
What converts this into an optimization tool is the realization that both quantities feed naturally into an iterative decision loop.
The posterior mean provides a cheap surface on which to run standard optimizers (L-BFGS, CG, NEB relaxation), and the posterior variance provides a criterion for when and where to request the next expensive electronic structure evaluation.
This is a Bayesian optimization (BO) loop \cite{jonesEfficientGlobalOptimization1998,shahriariTakingHumanOut2016,gramacySurrogatesGaussianProcess2020}, adapted from the scalar setting (optimize an unknown function) to structured PES problems (find saddle points, minimum energy paths, and local minima) with gradient observations.

\begin{figure}[htbp]
\centering
\includegraphics[width=0.85\textwidth]{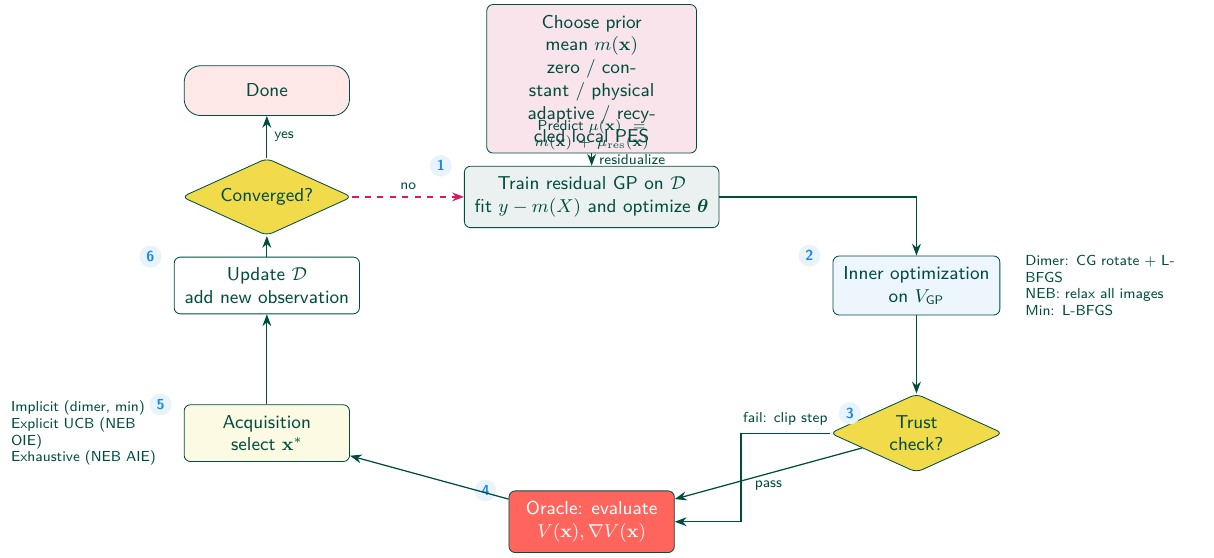}
\caption{\label{fig:bo_loop}Visual overview of the Bayesian surrogate loop (Algorithm \ref{alg:bo_loop}). Numbered steps proceed clockwise: (1) train the GP, (2) optimize on the surrogate, (3) check trust constraints, (4) evaluate the oracle, (5) select the next query point, (6) update the training set. The oracle (coral) is the only expensive step; all others operate on the cheap surrogate. Method-specific annotations indicate how each algorithm instantiates the inner optimization and acquisition steps. The prior-mean box marks one broader design-space extension that can be studied within the same outer loop without becoming the focus of the present tutorial.}
\end{figure}

The abstraction that makes this unification possible is simple: the GP operates on configurations in \(\mathbb{R}^{3N}\) with associated energy and force observations.
The GP does not know whether a state came from a dimer midpoint, a NEB image, or a minimization step.
The same kernel, the same covariance algebra, the same hyperparameter training applies regardless.
Methods differ only in which optimizer produced the current geometry and which acquisition criterion selects the next one.
Figure \ref{fig:bo_loop} gives the corresponding visual flow for the generic Bayesian surrogate loop.

\begin{algorithm}[H]
\caption{Generic Bayesian surrogate loop for PES optimization}\label{alg:bo_loop}
\begin{algorithmic}[1]
\Require Initial configuration(s) $\mathbf{X}_0$, oracle $V(\cdot)$, convergence threshold $\epsilon$
\State Evaluate oracle at $\mathbf{X}_0$; initialize $\mathcal{D} = \{(\mathbf{x}, V(\mathbf{x}), \nabla V(\mathbf{x}))\}$
\Repeat
  \State \textsc{Select} training subset $\mathcal{S} \subseteq \mathcal{D}$ \Comment{FPS, Section~\ref{sec:fps}}
  \State \textsc{Train} hyperparameters $\boldsymbol{\theta}$ on $\mathcal{S}$ \Comment{SCG on MAP-NLL, Section~\ref{sec:map-regularization}}
  \State \textsc{Build} prediction model from $\mathcal{D}$ with $\boldsymbol{\theta}$ \Comment{exact GP or RFF, Section~\ref{sec:rff}}
  \State \textsc{Optimize} on surrogate $V_{\text{GP}}$: method-specific inner loop \Comment{step 5}
  \State \textsc{Clip} proposed step via trust region \Comment{Section~\ref{sec:adaptive-trust}}
  \State \textsc{Acquire}: select next oracle point $\mathbf{x}^*$ via criterion $\alpha(\mathbf{x})$ \Comment{step 7}
  \State Evaluate oracle at $\mathbf{x}^*$; $\mathcal{D} \gets \mathcal{D} \cup \{(\mathbf{x}^*, V, \nabla V)\}$
\Until{$|\mathbf{F}(\mathbf{x}^*)| < \epsilon$}
\end{algorithmic}
\end{algorithm}

Table \ref{tab:instantiation} summarizes how each method instantiates Algorithm \ref{alg:bo_loop}.

\begin{table}[htbp]
\caption{Instantiation of Algorithm \ref{alg:bo_loop} across methods}
\label{tab:instantiation}
\centering
\small
\setlength{\tabcolsep}{3pt}
\begin{tabularx}{\textwidth}{@{}l>{\raggedright\arraybackslash}X>{\raggedright\arraybackslash}X>{\raggedright\arraybackslash}X@{}}
\hline
Step & Minimization & GP-dimer & GP-NEB\\
\hline
1. Select subset & Global FPS & Global FPS & Per-bead FPS\\
2. Train $\boldsymbol{\theta}$ & SCG on MAP & SCG on MAP & SCG on MAP\\
3. Build model & Exact/RFF & Exact/RFF & Exact/RFF\\
4. Inner optimization & L-BFGS & CG rotate + L-BFGS & NEB relaxation\\
5. Trust clip & EMD/Euclidean & EMD & EMD\\
6. Acquisition & Implicit & Implicit & MaxVariance/UCB (OIE) or exhaustive (AIE)\\
\hline
\end{tabularx}
\end{table}

The inner optimization proposes configurations by running a standard optimizer on the surrogate surface.
The surrogate is cheap, so the inner loop can run to convergence (or until the trust boundary is reached).
Two quantities govern when the inner loop should terminate and when the oracle should be consulted.
Both are derived from the GP posterior variance projected onto the subspace relevant to each method.

For saddle point methods (dimer, OT-GP dimer [OTGPD]) and NEB, the relevant uncertainty is the gradient variance perpendicular to a preferred direction \(\boldsymbol{\tau}\) (the dimer orient or the NEB path tangent):

\begin{equation}
\sigma_\perp(\mathbf{x}, \boldsymbol{\tau}) = \sqrt{\sum_{d=1}^{D} \mathrm{var}\!\left[\frac{\partial V_{\text{GP}}}{\partial x_d}\right]\!(1 - \tau_d^2)}
\label{eq:perp_variance}
\end{equation}

For minimization, no preferred direction exists and the total gradient uncertainty \(\sigma_g = \sqrt{\sum_d \mathrm{var}[\partial V_{\text{GP}} / \partial x_d]}\) replaces \(\sigma_\perp\).

The LCB \cite{hennigProbabilisticNumericsa,garridotorresLowScalingAlgorithmNudged2019} convergence criterion augments the inner loop stopping rule to prevent premature convergence in uncertain regions:

\begin{equation}
\|\nabla V_{\text{GP}}\|_{\text{eff}} = \|\nabla V_{\text{GP}}\| + \kappa \cdot \sigma(\mathbf{x}, \boldsymbol{\tau})
\label{eq:lcb_convergence}
\end{equation}

where \(\sigma\) is \(\sigma_\perp\) for saddle-point and NEB methods, or \(\sigma_g\) for minimization.
The inner loop continues until \(\|\nabla V_{\text{GP}}\|_{\text{eff}}\) drops below the GP tolerance.
When \(\kappa = 0\) this reduces to the standard gradient norm test.
In the OTGPD variant the GP tolerance itself is adapted across outer iterations.
When the true force is far from the convergence threshold the inner loop uses a loose tolerance (divisor of 2), accepting imprecise solutions on the surrogate and avoiding wasted inner steps on an inaccurate GP surface.
As the true force approaches the threshold, the divisor ramps linearly to a configured maximum, tightening the inner convergence to match the accuracy the surrogate has attained.
This schedule prevents the optimizer from overshooting on early, data-poor surrogates while still extracting full precision from well-trained models near convergence.

On the acquisition side, the NEB-OIE variant provides the clearest example of an explicit acquisition function.
The simpler one-image selector is pure image choice by maximum GP energy variance,

\begin{equation}
i^* = \arg\max_{i \in \mathcal{U}} \mathrm{var}[V_{\text{GP}}(\mathbf{R}_i)]
\label{eq:oie_selection}
\end{equation}

which prioritizes the unevaluated image where the surrogate has seen the least data in the kernel geometry.
The more aggressive alternative is a UCB criterion \cite{srinivasGaussianProcessOptimization2010} that selects the unevaluated image with the highest combined score:

\begin{equation}
i^* = \arg\max_{i \in \mathcal{U}} \left[|\mathbf{F}_i^{\text{NEB}}| + \kappa \cdot \sigma_\perp(\mathbf{R}_i, \boldsymbol{\tau}_i)\right]
\label{eq:ucb_acquisition}
\end{equation}

where \(\mathcal{U}\) is the set of unevaluated images.
This balances exploitation (images with large NEB forces) against exploration (images with high uncertainty).
When \(\kappa = 0\) the UCB selection reduces to force-only (pure exploitation); the pure-variance selector of Eq. \ref{eq:oie_selection} is a separate acquisition rule rather than the \(\kappa = 0\) limit.
In \texttt{chemgp-core} the default OIE choice is UCB, while MaxVariance remains available as the minimal pure-exploration variant.

The dual of LCB convergence operates at the oracle level.
A variance gate suppresses unnecessary oracle evaluations when the surrogate is already confident:

\begin{equation}
\text{skip oracle if } \sigma_\perp(\mathbf{x}, \boldsymbol{\tau}) < \sigma_{\text{gate}}
\label{eq:variance_gate}
\end{equation}

Three acquisition modes cover the methods in this review.
\emph{Implicit} acquisition (GP-minimize, GP-dimer, OTGPD) has no separate selection step: the inner loop proposes a configuration by optimizing on the surrogate, the trust region clips the step, and the oracle evaluates wherever the clipped step lands.
Trust violation is itself a secondary acquisition signal: it forces evaluation at the trust boundary when the proposal overshoots.
\emph{Explicit} acquisition (NEB OIE) applies either Eq. \ref{eq:oie_selection} or Eq. \ref{eq:ucb_acquisition} after inner relaxation to select the single most informative image from the unevaluated set.
This is closest to classical BO.
The chemgp-core implementation also provides MaxVariance, EI, and Thompson sampling strategies as alternatives.
\emph{Exhaustive} acquisition (NEB AIE) evaluates all \(P\) images at each outer iteration, bypassing image selection entirely.
Table \ref{tab:acquisition_criteria} summarizes the acquisition strategies used in each method.
All three share the same Bayesian surrogate loop structure (Algorithm \ref{alg:bo_loop}), differing only in how the acquisition criterion selects the next oracle point.

\begin{table}[htbp]
\caption{\label{tab:acquisition_criteria}Acquisition criteria across GP methods. GP-minimization and GP-dimer use implicit acquisition (trust-clipped step from inner loop), GP-NEB OIE uses explicit one-image selection from unevaluated images (MaxVariance or UCB; UCB is the default in \texttt{chemgp-core}), and GP-NEB AIE uses exhaustive evaluation of all images.}
\centering
\begin{tabular}{lllr}
Method & Mode & Selection Criterion & Calls/iter\\
\hline
GP-minimization & Implicit & Trust-clipped step & 1\\
GP-dimer & Implicit & Trust-clipped step & 1\\
GP-NEB OIE & Explicit & MaxVariance or UCB & 1\\
GP-NEB AIE & Exhaustive & All images & \(P\)\\
\end{tabular}
\end{table}

The three application sections that follow each instantiate Algorithm \ref{alg:bo_loop} for their specific inner optimization and acquisition criterion.
\section{GPR-Accelerated Minimum Mode Following, the GP-dimer}
\label{sec:gprdimer}
\subsection{Overview}
\label{sec:gprdimer-overview}
The standard dimer method is expensive because it is \emph{iterative at two levels}: every translation step requires multiple rotation steps, each requiring a fresh electronic structure evaluation.
A GP surrogate trained on the accumulated data replaces these inner evaluations with cheap predictions, and only the outer loop returns to the true PES to validate and extend the training set.
Whereas GP-NEB requires known initial and final states (Section \ref{sec:gprneb}), the GP-dimer requires only a starting configuration and an initial guess for the dimer orientation.

The idea of using a machine-learned surrogate to accelerate saddle point searches goes from neural networks \cite{petersonAccelerationSaddlepointSearches2016} to direct applications of Gaussian Processes \cite{denzelGaussianProcessRegression2018a,koistinenMinimumModeSaddle2020,fdez.galvanRestrictedVarianceConstrainedReaction2021} to specialized inverse-distance kernels \cite{koistinenMinimumModeSaddle2020,vishartAcceleratingCatalysisSimulations}, to improved runtime and reliability from optimal transport extensions \cite{goswamiAdaptivePruningIncreased2025}.

The specific algorithmic choices presented here (per-pair-type length scales optimized by maximum likelihood, the SE kernel in inverse-distance space, L-BFGS for translation) represent one point in a large design space.
In the Bayesian surrogate loop (Algorithm \ref{alg:bo_loop}), the GP-dimer instantiates the inner optimization as CG rotation followed by L-BFGS translation on \(V_{\text{GP}}\), and uses implicit acquisition (Section \ref{sec:bo-framework}): the oracle evaluates at the trust-clipped midpoint.
When \(\kappa > 0\), the LCB convergence criterion (Eq. \ref{eq:lcb_convergence}) prevents the inner loop from terminating in uncertain regions.
The remaining loop steps (FPS subset, SCG training, RFF prediction, EMD trust) follow the generic framework and are developed in Section \ref{sec:otgp}.
The inner components (kernel form, hyperparameter strategy, optimizer, trust region) can be varied independently, and in Section \ref{sec:gprneb}, at least three independent implementations with different kernel, descriptor, and hyperparameter choices achieve comparable performance.
The \texttt{chemgp-core} and \texttt{gpr\_optim} codes share the same algorithmic choices, though both differ from the original publications in specific convergence criteria and inner-loop heuristics.

The algorithm requires an initial exploration phase before the surrogate can take over.
Skipping it and fitting a GP to just one or two evaluations produces a surrogate that is effectively a constant surface with large uncertainty everywhere; the dimer has no meaningful curvature to follow and wanders randomly.
The two phases are:
\subsubsection{Phase 1: Initial Rotations (Finding the Minimum Mode)}
\label{sec:org7704602}
The first few electronic structure evaluations establish the minimum mode direction.
The dimer midpoint \(\mathbf{R}_0\) and one endpoint \(\mathbf{R}_1\) are evaluated on the true PES.
The dimer is then repeatedly rotated, evaluating the true PES at each new \(\mathbf{R}_1\) position, until the orientation converges.
Typically, \(\sim 6\) evaluations suffice.
The convergence criterion for this phase is either a small preliminary rotation angle (\(\omega^* < 5^{\circ}\)) or a small angle between successive converged orientations.
These initial evaluations constitute the training set for the first GP model.
\subsubsection{Phase 2: GPR Iterations (Rotation + Translation on the Surrogate)}
\label{sec:orga691ee9}
Once the initial minimum mode is established, the GP takes over.
Algorithm \ref{alg:gp_dimer} and Figure S1 (right) summarize the Phase 2 iteration.

\begin{algorithm}[H]
\caption{GP-dimer Phase 2 (surrogate iterations)}\label{alg:gp_dimer}
\begin{algorithmic}[1]
\Require Training set $\mathcal{D}$ from Phase 1, force threshold $\epsilon_{\text{force}}$
\Repeat
  \State Fit GP hyperparameters by maximizing $\mathcal{L}(\boldsymbol{\theta})$ (Eq.~\ref{eq:marginal_likelihood}) on $\mathcal{D}$
  \State Run full dimer (rotation via CG + translation via L-BFGS) on $V_{\text{GP}}(\mathbf{x})$
  \Statex \hspace{\algorithmicindent} until surrogate convergence or trust radius violated
  \State Evaluate true PES at midpoint $\mathbf{R}_0$
  \If{$|\mathbf{F}(\mathbf{R}_0)| < \epsilon_{\text{force}}$}
    \State \Return saddle point at $\mathbf{R}_0$
  \EndIf
  \State Add $(\mathbf{x}, V, \nabla V)$ to $\mathcal{D}$
\Until{converged}
\end{algorithmic}
\end{algorithm}

The surrogate prediction at a new point is:

\begin{equation}
V_{\text{GP}}(\mathbf{x}_{\text{new}} \mid \mathcal{D}, \boldsymbol{\theta}_{\text{opt}}) \approx V(\mathbf{x}_{\text{new}})
\label{eq:gp_prediction}
\end{equation}

where \(\mathcal{D} = \{(\mathbf{x}_i, V_i, \nabla V_i)\}_{i=1}^{M}\) is the training set of previously computed configurations.
The GP predictions of the force at the dimer endpoint \(\mathbf{R}_1\) replace the finite-difference extrapolations normally used.
The GP posterior mean for the gradient uses all accumulated force data rather than just the most recent pair, and therefore produces a more accurate estimate.

The source of the savings is easy to trace.
In the "improved" dimer formulation in eOn \cite{chillEONSoftwareLong2014} \footnote{\url{https://eondocs.org}} \cite{kastnerSuperlinearlyConvergingDimer2008,olsenComparisonMethodsFinding2004}, each translation step requires 5 to 15 rotation evaluations (each a full electronic structure call) to converge the orientation.
The GP replaces almost all of these inner rotations with surrogate queries that cost microseconds rather than minutes, though the first few initial rotations take true forces.
The outer loop still needs one true evaluation per translation step to validate the surrogate and extend the training set, but the inner loop is essentially free.
Benchmarks \cite{goswamiEfficientImplementationGaussian2025,goswamiAdaptivePruningIncreased2025} show the median evaluation count dropping by a factor of ten which is precisely the factor one expects from eliminating the inner rotation cost.
Despite working in Cartesian coordinates, the GP-dimer achieves performance comparable to internal-coordinate methods (Sella \cite{hermesSellaOpenSourceAutomationFriendly2022}), because the inverse-distance kernel (Eq. \ref{eq:idist_kernel}) learns per-pair length scales that adapt to the stiffness landscape of each reaction.

\begin{figure}[htbp]
\centering
\includegraphics[width=0.8\textwidth]{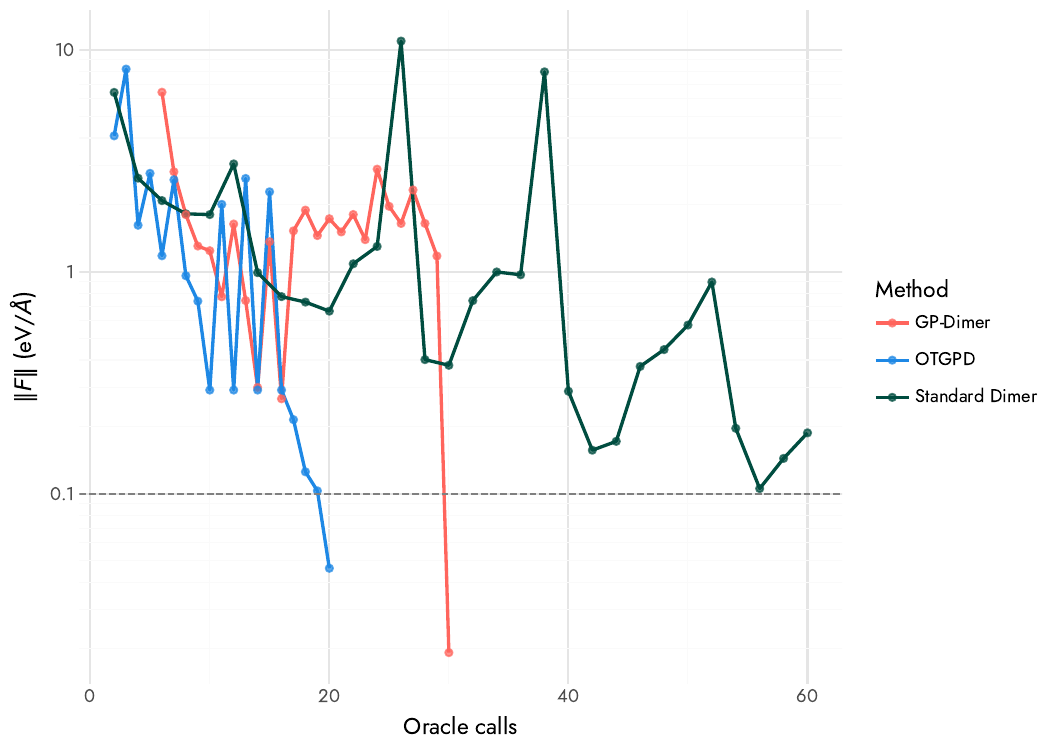}
\caption{\label{fig:rpc_dimer_convergence}Convergence comparison of the standard dimer, GP-dimer, and OT-GP dimer (OTGPD) on a molecular system (C\textsubscript{3H}\textsubscript{5} allyl radical, 8 atoms, 24 DOF) via eOn serve mode. The maximum per-atom force (eV/\AA{}) is plotted against oracle evaluations on a logarithmic scale. The OTGPD variant reaches the convergence threshold (gray dashed, 0.1 eV/\AA{}) with the fewest oracle calls; the GP-dimer shows oscillations from surrogate retraining instabilities that the OT-GP extensions suppress.}
\end{figure}
\subsection{Trust Regions and Early Stopping}
\label{sec:trust-regions}
\begin{figure}[htbp]
\centering
\includegraphics[width=0.85\textwidth]{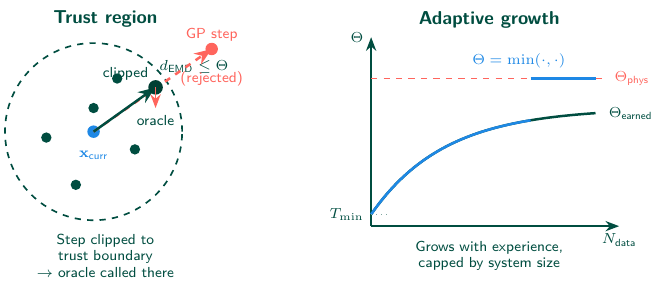}
\caption{\label{fig:trust_region_concept}Trust region mechanism. (\emph{Left}) A GP-proposed step (coral) that exceeds the trust boundary \(d_{\text{EMD}} \leq \Theta\) is clipped to the boundary; the oracle evaluates at the clipped location. (\emph{Right}) The trust radius grows with accumulated data via an exponential saturation curve (\(\Theta_{\text{earned}}\)), capped by a system-size-dependent physical ceiling (\(\Theta_{\text{phys}}\)).}
\end{figure}

Left unconstrained, the GP-guided optimizer will eventually propose a geometry that lies outside the region where the surrogate is accurate \cite{nocedalNumericalOptimization2006}.
Two distinct things can go wrong, and each requires its own safeguard.
Figure \ref{fig:trust_region_concept} illustrates the trust-boundary clipping and adaptive-radius growth used to keep proposed steps within the sampled neighborhood.

\subsubsection{Extrapolation to Unseen Geometries}
\label{sec:org5bd5311}
The first problem is that the optimizer proposes a configuration that is structurally unlike anything in the training set \cite{denzelGaussianProcessRegression2018}, and the models are best as interpolators, not extrapolators.
The GP posterior mean in such a region is pulled toward the prior mean (near zero after subtracting the constant offset), which typically produces a spurious minimum that traps the dimer.
The posterior variance is large, but the optimizer ignores variance and follows the mean.
To detect when a proposed geometry has left the neighborhood of the training data, we measure its dissimilarity to the nearest training point using the 1D-max-log distance \cite{goswamiAdaptivePruningIncreased2025}:

\begin{equation}
D_{\text{1Dmaxlog}}(\mathbf{x}_1, \mathbf{x}_2) = \max_{i,j} \left| \log \frac{r_{ij}(\mathbf{x}_2)}{r_{ij}(\mathbf{x}_1)} \right|
\label{eq:max1dlog}
\end{equation}

This metric operates on interatomic distance \emph{ratios} rather than absolute distances, so it is scale-invariant: a 10\% change in the closest atom pair registers the same whether the pair is 1 Angstrom or 3 Angstrom apart.
A proposed configuration is accepted only if its 1D-max-log distance to the nearest training point is below a threshold; otherwise, the algorithm falls back to evaluating the true PES at the trust boundary.
\subsubsection{Atoms Approaching Too Closely}
\label{sec:org1529021}
The second problem is specific to molecular systems: the optimizer can push two atoms into near-overlap.
Even if the geometry is technically within the trust region (because the \emph{log-ratio} distance to a training point is small), the electronic structure code may fail on a geometry with sub-Angstrom contacts.
A step-size limit prevents this:

\begin{equation}
L_{\max} = \frac{1}{2}(1 - r_{\text{limit}}) \cdot d_{\min}
\label{eq:mindistatm}
\end{equation}

where \(d_{\min}\) is the minimum interatomic distance in the current configuration and \(r_{\text{limit}} \in [0, 1]\) controls the conservativeness of the constraint.
Values near 1 enforce small, cautious steps; values near 0 allow larger steps.
This constraint is separate from the trust radius because the failure mode is different: the trust radius catches extrapolation in feature space, while the minimum-distance constraint catches a physically unacceptable geometry that the feature-space metric might miss.

\begin{figure}[htbp]
\centering
\includegraphics[width=0.7\textwidth]{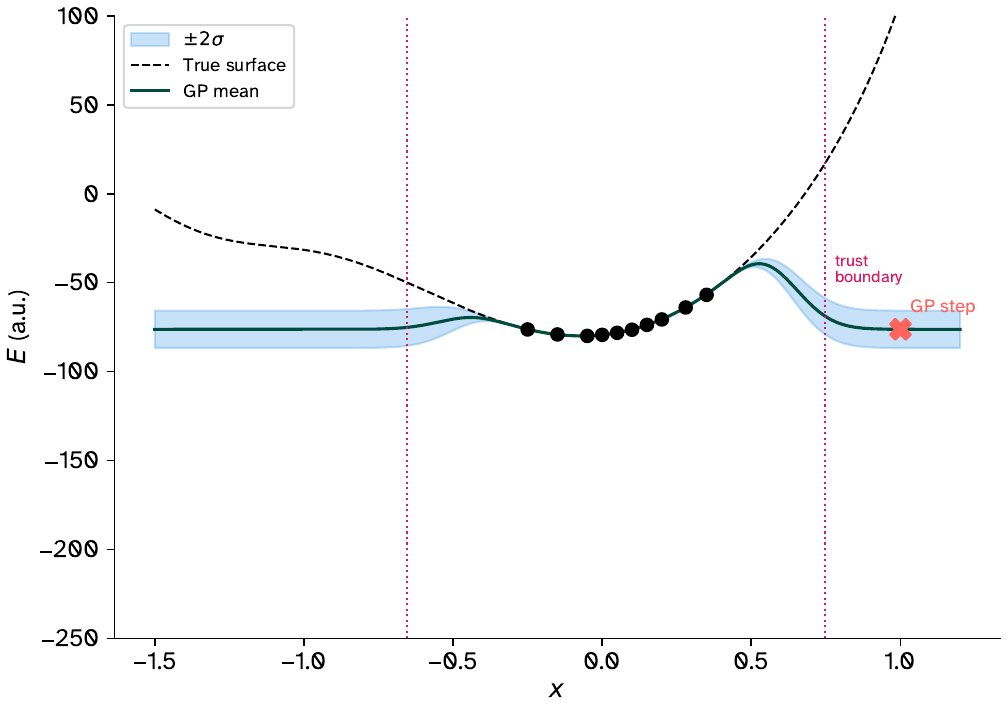}
\caption{\label{fig:mb_trust_region}Trust region mechanism on a 1D slice (\(y = 0.5\)) of the Muller-Brown surface. The GP posterior mean (teal) and \(\pm 2\sigma\) confidence band (light blue) are accurate near the training data (black dots) but diverge from the true surface (black dashed) outside the trust boundaries (magenta dotted verticals). A hypothetical GP-proposed step at \(x = 1.0\) (coral cross, labeled "GP step") falls outside the trust region, where the surrogate is unreliable; the algorithm instead evaluates the true PES at the trust boundary (teal star, "Oracle fallback").}
\end{figure}

The optimal transport extensions \cite{goswamiAdaptivePruningIncreased2025} cover walltime considerations due to the cubic scaling of the Cholesky factorization, the oscillation of surrogate surfaces on retraining, and add a trust region based on molecular similarity considerations, with bounds on hyperparameters.
Here we will demonstrate how trivially extensible these concepts are to the NEB within our framework.
This forms section \ref{sec:gprneb} where the surrogate accelerates the relaxation of multiple images along the minimum energy path.
\section{GPR-Accelerated Nudged Elastic Band}
\label{sec:gprneb}
GP-NEB instantiates Algorithm \ref{alg:bo_loop} with NEB force relaxation as the inner optimization and explicit one-image selection from unevaluated images as the acquisition step.
The AIE variant uses exhaustive acquisition (all \(P\) images per iteration); the OIE variant selects a single image via either the pure-variance criterion (Eq. \ref{eq:oie_selection}) or the UCB score (Eq. \ref{eq:ucb_acquisition}).
Per-bead FPS and EMD trust adapt the shared components to the string discretization.
The additional design choice specific to NEB is \emph{how many} images to evaluate at each outer iteration, which determines the trade-off between surrogate accuracy and the cost per cycle \cite{koistinenNudgedElasticBand2017,koistinenNudgedElasticBand2019,denzelGaussianProcessRegression2019,vishartAcceleratingCatalysisSimulations,garridotorresLowScalingAlgorithmNudged2019,raggiRestrictedVarianceMolecularGeometry2020,fdez.galvanRestrictedVarianceConstrainedReaction2021}.

Two variants bracket the design space.
In the \emph{all-images-evaluated} (AIE) variant, all \(P\) images are evaluated on the true PES at each outer iteration.
This provides a dense training set before each surrogate relaxation and, on the illustrative LEPS example, reduces the total evaluations from 156 to 100.
In the more aggressive \emph{one-image-evaluated} (OIE) variant, only the image selected by the one-image acquisition rule is evaluated.
For the pure-variance selector of Eq. \ref{eq:oie_selection}, this means the image with the largest GP posterior energy variance.
That active learning criterion selects the image where the surrogate has seen the least data in the kernel geometry (the largest posterior variance, which again is a sampling-density signal rather than a predicted error), and on the illustrative LEPS example reduces the total evaluations from 100 (AIE) to 42.
Equation \ref{eq:oie_selection} is the pure-variance criterion.
The UCB alternative of Eq. \ref{eq:ucb_acquisition} balances force magnitude against uncertainty and is the default in \texttt{chemgp-core}.
The trust region safeguards from Section \ref{sec:trust-regions} apply to both variants.
When an image drifts beyond the reliable region of the surrogate, the constraint violation triggers an evaluation at that image, which is an implicit acquisition strategy.
Figure S2 (right) illustrates both variants.

Two independent open implementations of surrogate-accelerated NEB are particularly relevant for comparison.
The CatLearn MLNEB \cite{vishartAcceleratingCatalysisSimulations}, built on ASE \cite{larsenAtomicSimulationEnvironment2017}, uses a Student's t-process (Section \ref{sec:hyperparameters}) with a single isotropic length scale for surface catalysis; the inverse-distance SE kernel with per-pair-type length scales used here is a different modeling choice \footnote{and is adopted in the upstream codebase}.
Published studies in these neighboring implementations report reductions ranging from factors of several to roughly an order of magnitude on their respective benchmark sets, despite differing kernel, descriptor, and hyperparameter choices.
The fact that these different combinations converge on comparable performance suggests that the active learning loop, not the specific surrogate model, is the primary source of the savings.

\begin{figure}[htbp]
\centering
\includegraphics[width=\textwidth]{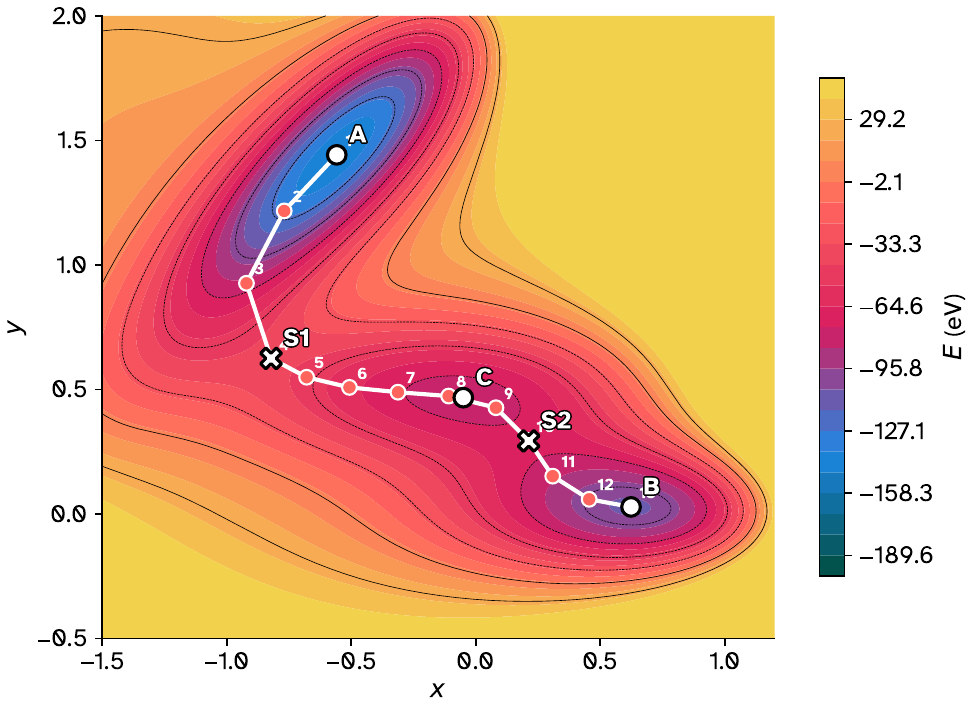}
\caption{\label{fig:mb_neb}Muller-Brown potential energy surface with NEB path overlay. Filled contours show the energy landscape with three local minima (A, B, C) and two saddle points (S1, S2). Eleven NEB images (coral circles, numbered) trace the minimum energy path from A to B through S2. The climbing image (highest-energy interior image) approximates the saddle point. Energy values are reported in the conventional Muller-Brown reduced units and clipped to the range \([-200, 50]\) for visualization.}
\end{figure}

\begin{figure}[htbp]
\centering
\includegraphics[width=\textwidth]{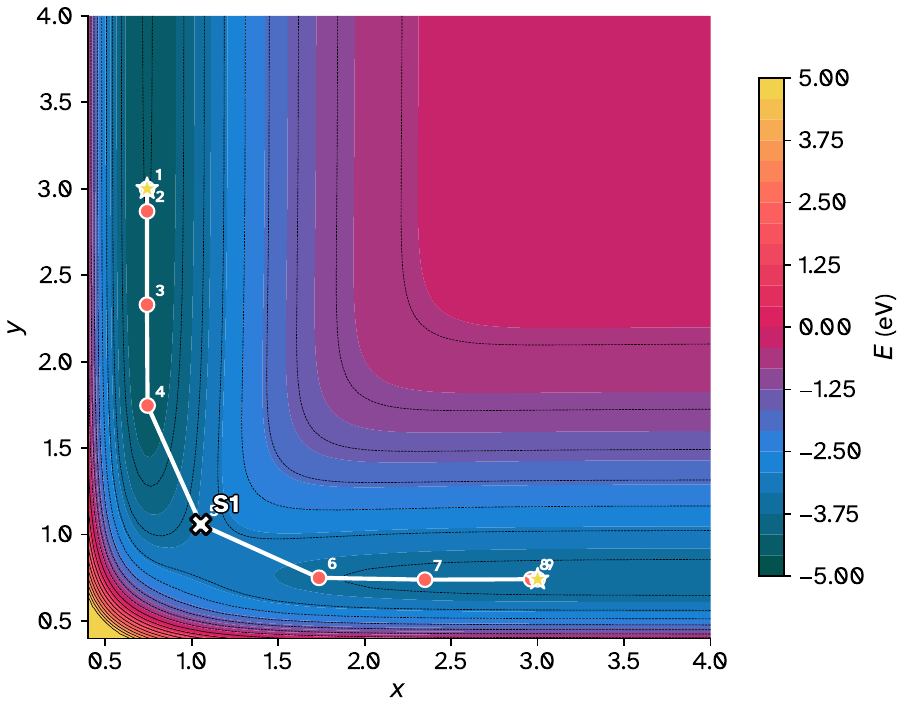}
\caption{\label{fig:leps_neb}LEPS potential energy surface with NEB path overlay. The collinear atom transfer reaction \(A + BC \to AB + C\) is plotted as a function of bond distances \(r_{AB}\) and \(r_{BC}\). Seven interior NEB images and two fixed endpoints (nine path points, coral circles numbered; endpoints also marked by yellow stars) are optimized in the full 9-dimensional coordinate space and projected onto the \((r_{AB}, r_{BC})\) plane. The climbing image converges to the saddle region near \(r_{AB} \approx r_{BC} \approx 1.0\) \AA{}. Contour spacing is 0.5 eV; energies clipped to \([-5, 5]\) eV.}
\end{figure}

\begin{figure}[htbp]
\centering
\includegraphics[width=0.8\textwidth]{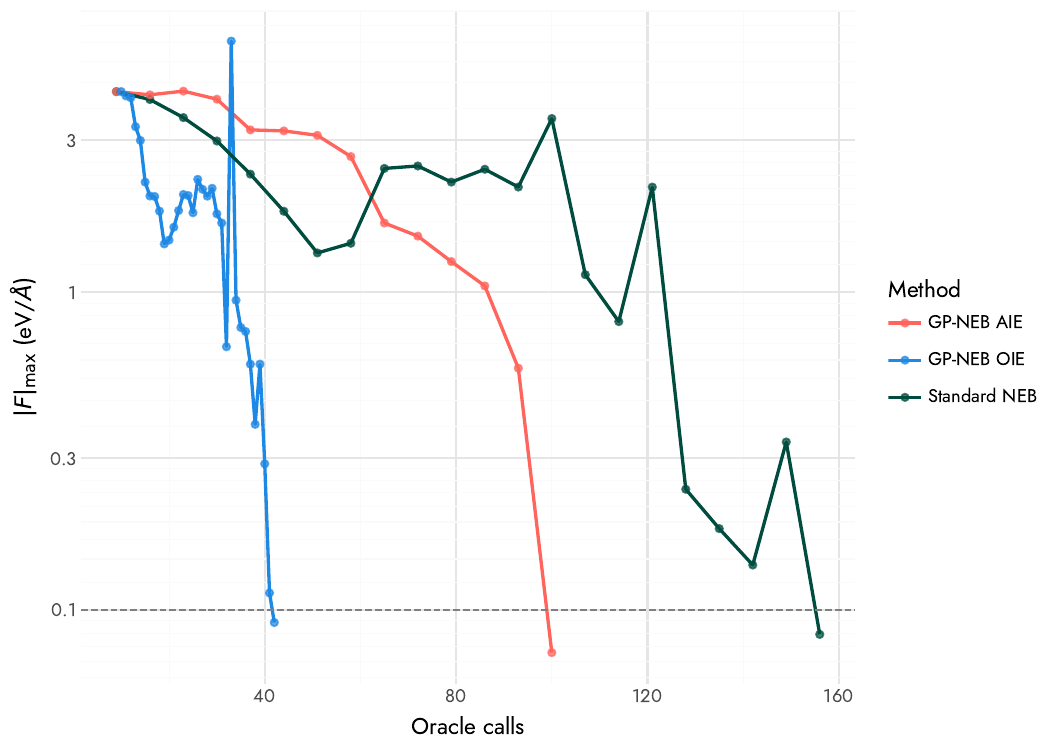}
\caption{\label{fig:leps_aie_oie}Convergence of NEB variants on the LEPS surface. Maximum per-atom force versus oracle evaluations on a logarithmic scale. Standard NEB (156 calls), AIE (100 calls), and OIE (42 calls) all reach the convergence threshold (dashed, 0.1 eV/\AA{}). The OIE variant evaluates only the highest-variance image per cycle (Eq. \ref{eq:oie_selection}) and converges fastest.}
\end{figure}

Figures \ref{fig:mb_neb}--\ref{fig:leps_aie_oie} illustrate the GP-NEB on two test surfaces.
On the Muller-Brown surface (Figure \ref{fig:mb_neb}), an 11-image climbing-image NEB resolves the path from minimum A through saddle S2 to minimum B. The LEPS surface (Figure \ref{fig:leps_neb}) provides a higher-dimensional test case modeling a collinear atom transfer \(A + BC \to AB + C\), where the 9-dimensional NEB path projects onto the \((r_{AB}, r_{BC})\) plane.
The convergence comparison in Figure \ref{fig:leps_aie_oie} quantifies the oracle savings of the AIE and OIE acquisition strategies on this surface.

In the chain-of-states picture (Section \ref{sec:chain-of-states}), the inner loop evolves the discretized path under the force field of \(V_{\text{GP}}\) rather than \(V\).
The surrogate surface is a flexible interpolator trained on finitely many data points, and it generically has \emph{more} stationary points than the true PES.
Spurious minima and saddle points arise in the regions between training configurations where the GP reverts toward its prior.
These additional critical points are an unavoidable consequence of building a smooth model from sparse data, and the outer loop exists precisely to filter them.
At each outer iteration, the true PES is evaluated at the configuration proposed by the surrogate optimization, and if the true forces are not small, the new data point eliminates the spurious feature that trapped the optimizer.
The trust region (Section \ref{sec:trust-regions}) provides a second filter, preventing the optimizer from reaching distant spurious features by confining each inner-loop step to the neighborhood where the GP has earned credibility from training data.

A property that makes this scheme well-posed is that the GP operates on Cartesian coordinates.
Every configuration visited during the inner-loop optimization, including configurations at spurious stationary points of \(V_{\text{GP}}\), is a valid atomic geometry in \(\mathbb{R}^{3N}\) that can be handed directly to the electronic structure code for evaluation.
The inverse-distance kernel uses a feature map \(\boldsymbol{\phi}(\mathbf{x}) = \{1/r_{ij}(\mathbf{x})\}\) internally, but the optimization variable remains \(\mathbf{x}\), so there is no inverse problem.
This would not hold for a method that optimized the path in descriptor space (e.g., SOAP), where the optimized images would be points in \(\mathbb{R}^{d_{\text{SOAP}}}\), and recovering Cartesian pre-images would require solving a separate inverse problem that may have no solution (not every point in descriptor space corresponds to a valid geometry) or multiple solutions (the descriptor map is not injective).
By keeping the optimization in Cartesian space and confining the descriptor to the kernel interior, the GP-NEB avoids this entirely.
Every proposed path is physically realizable, and the only question is whether it lies on the true MEP.

Path initialization matters for the GP-NEB.
The sequential image-dependent pair potential (S-IDPP) method \cite{schmerwitzImprovedInitializationOptimal2024}, which builds on the IDPP \cite{smidstrupImprovedInitialGuess2014} which interpolates interatomic distances rather than Cartesian coordinates, produces chemically reasonable initial configurations whose training data samples a more physical region of configuration space than linear interpolation would provide.
This method may be augmented by the iterative rotations and assignments algorithm \cite{gundeIRAShapeMatching2021,goswamiEnhancedClimbingImage2026}.
Linear interpolation in Cartesian coordinates often creates initial paths where atoms pass through each other or where interatomic distances become unphysically short, producing training data from a region of the PES that is irrelevant to the reaction pathway and poorly conditioned for GP learning.
These calculations may be visualized together on 2D plots \cite{goswamiTwodimensionalRMSDProjections2026}.
\section{GPR-Accelerated Minimization}
\label{sec:gprmin}
Minimization is the simplest instantiation of Algorithm \ref{alg:bo_loop}: the inner optimization is L-BFGS on \(V_{\text{GP}}\), trust clipping is Euclidean or EMD, and acquisition is implicit (the oracle evaluates at the trust-clipped L-BFGS result).
The LCB convergence criterion (Eq. \ref{eq:lcb_convergence}, with total gradient \(\sigma_g\) instead of \(\sigma_\perp\)) optionally augments the inner stopping rule.
Denzel and Kastner \cite{denzelGaussianProcessRegression2018} developed GP-accelerated minimization systematically, benchmarking a GP-based geometry optimizer against L-BFGS on 26 molecular systems and finding that the Matern kernel in Cartesian coordinates outperforms the squared exponential.
They subsequently extended the approach to internal coordinates \cite{meyerGeometryOptimizationUsing2020} and to MEP optimization \cite{denzelGaussianProcessRegression2019}.
An important distinction is that these earlier GP optimizers operate with kernels defined directly on Cartesian or internal coordinates, without the inverse-distance feature map (Section \ref{sec:idist-kernel}) that provides rotational and translational invariance.
The inverse-distance kernel, introduced for NEB by Koistinen, Asgeirsson, and Jonsson \cite{koistinenNudgedElasticBand2019} and adopted throughout the present framework, avoids the need for explicit coordinate alignment between training configurations and enables the GP to generalize across rigid-body motions of the molecule.
Algorithm \ref{alg:gp_min} summarizes the iteration.

\begin{algorithm}[H]
\caption{GP-accelerated local minimization}\label{alg:gp_min}
\begin{algorithmic}[1]
\Require Initial configuration $\mathbf{x}$, force threshold $\epsilon_{\text{force}}$
\State Evaluate $V(\mathbf{x}), \mathbf{F}(\mathbf{x})$ on true PES
\Repeat
  \State Add $(\mathbf{x}, V, \nabla V)$ to training set $\mathcal{D}$
  \State Re-optimize GP hyperparameters (Eq.~\ref{eq:marginal_likelihood})
  \State Run L-BFGS on $V_{\text{GP}}(\mathbf{x})$ until GP forces vanish or trust radius violated
  \State Evaluate true PES at new $\mathbf{x}$
  \If{$|\mathbf{F}(\mathbf{x})| < \epsilon_{\text{force}}$}
    \State \Return minimum at $\mathbf{x}$
  \EndIf
\Until{converged}
\end{algorithmic}
\end{algorithm}

\begin{figure}[htbp]
\centering
\includegraphics[width=0.8\textwidth]{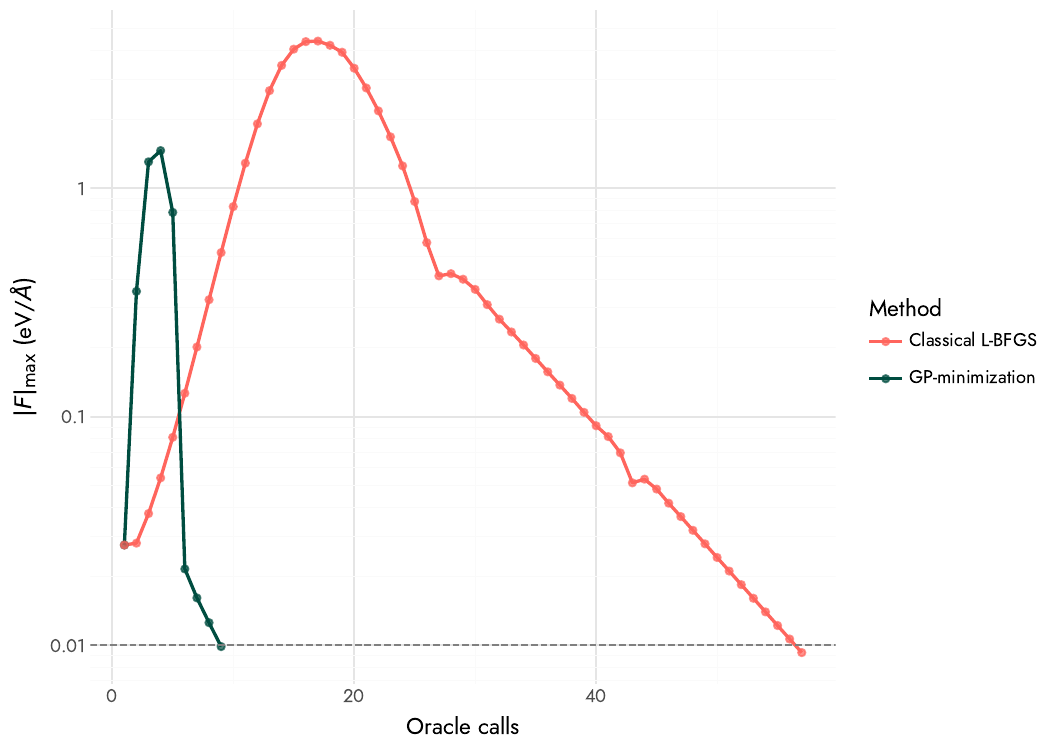}
\caption{\label{fig:leps_minimize_convergence}Convergence comparison of the GP-minimizer and classical L-BFGS on the LEPS surface. With a force convergence threshold of \(10^{-2}\) eV/\AA{}, the GP surrogate reaches the threshold in 9 oracle calls, compared with 57 for direct L-BFGS on the same starting configuration. Force values plotted on a logarithmic scale.}
\end{figure}

The trust region plays the same role as in the GP-dimer, preventing the optimizer from venturing too far from the reliable region of the surrogate.
The distance-based and interatomic-distance constraints from Section \ref{sec:trust-regions} apply directly.
Figure \ref{fig:leps_minimize_convergence} compares the GP-minimizer against classical L-BFGS on the LEPS surface.

GP-minimization does still use the GP posterior variance, but at the inner-loop stopping level via the LCB convergence criterion (Eq. \ref{eq:lcb_convergence}) rather than as an oracle gate.
Three considerations argue for the geometric trust radius over a variance gate in the minimization setting, and their interplay clarifies why NEB uses both mechanisms while minimization uses only one.

First, \emph{calibration}.
Each search in this framework builds its GP from scratch and accumulates at most a few tens of observations by the time it converges.
As noted when the predictive variance was introduced (Section\textasciitilde{}\ref{sec:gpr}), \(\sigma^2(\mathbf{x}_*)\) depends only on the kernel, the training locations, and the hyperparameters, and collapses to the noise floor at every training point whether or not the mean is accurate there.
For a per-search GP the length scales have been fit from the same handful of observations, so a low variance in their vicinity only says that the surrogate is confident it can interpolate its own data; it is not a statement about the accuracy of the posterior mean against the true PES.
Using this number as an oracle gate on each proposal would therefore over-trust the surrogate exactly when the dataset is sparse and the bias is largest.
The LCB convergence criterion is less aggressive in this respect because it enters only after the inner optimizer has converged on the current surrogate; the oracle is then queried at the candidate regardless of the variance, and the variance merely tightens the inner stopping rule.

Second, \emph{cost}.
Each variance evaluation in the full augmented GP requires a triangular solve against the Cholesky factor of \(\mathbf{K}\) whose cost is \(\mathcal{O}(M^2)\) per query and grows quadratically with the training set (Section \ref{sec:linalg}).
A variance gate that fires once per proposal adds one such solve to every inner-loop step, which in practice doubles or triples the GP overhead per outer iteration.
For minimization the inner loop can run hundreds of surrogate steps, so the gate would be evaluated hundreds of times for a single oracle call that in the trust-region design costs nothing beyond a comparison.

Third, \emph{chemical relevance}.
The geometric trust radius in EMD (Section \ref{sec:trust-regions}) measures per-atom displacement in Angstroms and is bounded below by a physical length scale (\(\sim 0.3\) \AA{} in \texttt{chemgp-core}) that reflects what a bond can reasonably do in one step regardless of what the surrogate reports.
The variance gate has no such floor: as the surrogate fills in, \(\sigma_\perp\) keeps shrinking even in regions where a single-step move would break physical bonds.
Tying the step to displacement rather than to kernel-space distance therefore encodes the same prior that a chemist would apply by hand and costs a single inner product.

NEB sits in a different trade-off because the acquisition decision is \emph{which} of \(P\) path images to evaluate next, not \emph{whether} to evaluate the proposed image at all.
The per-image variance must be computed for the selection criterion anyway (Eq. \ref{eq:oie_selection}), so the cost objection disappears and the calibration concern is mitigated by the fact that the NEB ensemble of images provides richer training data than a single-point optimizer accumulates.
NEB therefore benefits from both the geometric trust region (per image, to keep the chain well-behaved) and the variance-based image selection (per outer iteration, for acquisition), while minimization benefits only from the former.

The gains from GP acceleration are smaller for minimization than for saddle point searches, because the PES near minima is smooth and well-approximated by a quadratic, so standard L-BFGS already converges in few steps.
The GP surrogate provides the largest benefit when the starting configuration is far from the minimum or when the electronic structure cost per evaluation is high (large systems, high-level methods).
For a real molecular system, Figure \ref{fig:petmad_minimize_convergence} shows convergence of the GP-minimizer on the PET-MAD potential.

GPR-accelerated minimization is particularly useful as a subroutine in adaptive kinetic Monte Carlo (AKMC), where many local minimizations are needed to characterize the final states of transitions discovered by saddle point searches.
In AKMC, each saddle point found by the GP-dimer implies a transition to a new minimum, which must be located to continue the simulation.
Reusing the GP training data from the saddle point search, which already samples the PES near the transition path, can warm-start the minimization and reduce the number of additional evaluations needed.

Section \ref{sec:otgp} develops the Optimal Transport GP (OT-GP) extensions that address the failure modes of the basic framework and make the Bayesian surrogate loop reliable for production use.
\section{Practical Components for the Bayesian Surrogate Loop}
\label{sec:otgp}
The generic Bayesian surrogate loop of Section \ref{sec:bo-framework} (Algorithm \ref{alg:bo_loop}) needs four supporting components to be useful in production: a way to select training subsets so that hyperparameter optimization stays bounded, a way to keep MAP-NLL hyperparameter optimization from drifting into pathological regions, an adaptive trust radius that reflects how much the surrogate has actually learned, and a numerically stable solver for the kernel linear system.
Three of these (training-subset selection, MAP regularization, adaptive trust radius) admit a \emph{core} formulation that every method in this review uses, plus an OT-GP \emph{extension} that further improves accuracy or stability for harder problems.
The fourth component, the linear-algebra solver, is purely an implementation concern and lives in Section \ref{sec:linalg} alongside its OT-GP-flavored adaptive jitter strategy.
Random Fourier features (Section \ref{sec:rff}) are a separate optional extension that scales prediction to large training sets.

Each of the following subsections leads with the core formulation and then flags the OT-GP refinement explicitly.
The section answers three motivating questions, in order:
\begin{enumerate}
\item /How do we keep the surrogate honest?
\end{enumerate}
/ The signal variance can run away and the hyperparameters can oscillate, both of which produce surrogates that are unrelated to the true PES (Section \ref{sec:map-regularization}).
\begin{enumerate}
\item /How far should we trust the surrogate?
\end{enumerate}
/ A fixed trust radius is either too conservative (wasting oracle calls) or too aggressive (producing unphysical steps).
The threshold should reflect how much the GP has actually learned (Section \ref{sec:adaptive-trust}).
\begin{enumerate}
\item /Which training points matter?
\end{enumerate}
/ As the dataset grows, the cubic cost of hyperparameter optimization becomes the bottleneck.
Selecting a geometrically diverse subset keeps the cost bounded without sacrificing surrogate quality (Section \ref{sec:fps}).

Table \ref{tab:bo_components} summarizes how these components specialize for each method.

\begin{table*}[htbp]
\caption{Shared Bayesian optimization components across GP methods. Each column shows how the six components specialize for a given method. Acquisition modes and the oracle gate are formalized in Section \ref{sec:bo-framework}; the LCB convergence criterion (Eq.~\ref{eq:lcb_convergence}) governs the inner-loop stopping rule and is distinct from the acquisition step listed here.}
\label{tab:bo_components}
\centering
\small
\adjustbox{max width=\textwidth}{%
\begin{tabular}{l l l l l l}
\hline
Component & Minimize & GP-dimer & OTGPD & NEB AIE & NEB OIE \\
\hline
FPS subset & global & global & HOD & per-bead & per-bead \\
Trust metric & Euclid./EMD & EMD & EMD & EMD & EMD \\
RFF predict & optional & optional & optional & optional & optional \\
Acquisition & implicit (trust-clipped step) & implicit (trust-clipped step) & implicit (trust-clipped step, adaptive $T_{\text{GP}}$) & exhaustive & UCB (Eq.~\ref{eq:ucb_acquisition}) \\
Oracle gate & --- & $\sigma_\perp < \sigma_{\text{gate}}$ & $\sigma_\perp < \sigma_{\text{gate}}$ & --- & $\sigma_\perp$ phase \\
Inner optim & L-BFGS & rot.+trans. & rot.+trans. & L-BFGS & L-BFGS \\
\hline
\end{tabular}}
\end{table*}

\begin{figure}[H]
\centering
\adjustbox{max width=0.9\textwidth, max totalheight=0.85\textheight}{%
\begin{tikzpicture}[
  >=Stealth,
  node distance=8mm,
  box/.style={draw=RuhiTeal, fill=white, rounded corners=3pt,
              text=RuhiTeal, minimum height=2.4em, align=center,
              font=\small\sffamily, inner sep=5pt},
  coral/.style={box, fill=RuhiCoral, text=white},
  bluebox/.style={box, fill=RuhiBlue, text=white},
  decision/.style={box, diamond, fill=RuhiYellow, aspect=2, inner sep=3pt},
  start/.style={box, rounded corners=8pt},
  arr/.style={->, RuhiTeal, semithick},
  back/.style={->, RuhiPink, semithick, dashed},
  grp/.style={draw=RuhiTeal, rounded corners=5pt, inner sep=8pt, fill=black!5},
  lbl/.style={font=\scriptsize\sffamily, text=RuhiTeal},
  note/.style={font=\scriptsize\sffamily, text=RuhiPink, align=center},
  fmlink/.style={RuhiPink, dotted, semithick, -},
]

\node[coral] (oracle_eval)
  {Evaluate true PES at $\mathbf{R}_0$};

\node[decision, below=of oracle_eval] (true_conv)
  {$\|\mathbf{F}_{\text{true}}\| < \epsilon$\\and $C < 0$?};

\node[start, right=25mm of true_conv] (done)
  {Saddle found};

\node[box, below=of true_conv] (add_data)
  {Add $(\mathbf{x}, V, \nabla V)$ to $\mathcal{D}$};

\node[box, fill=RuhiPink!12, text width=4.0cm, below=9mm of add_data] (prior_mean)
  {Optional prior mean $m(\mathbf{x})$:\\zero / constant / physical\\adaptive or recycled local PES};

\node[lbl, text width=4.2cm, align=center, below=2mm of prior_mean]
  {Train on residuals $y-m(X)$, then add $m(\mathbf{x})$ back in prediction};

\node[bluebox, below=24mm of add_data] (fps)
  {FPS: select $M_{\text{sub}}$ diverse\\points from $\mathcal{D}$ (EMD metric)};

\node[bluebox, below=of fps] (train_sub)
  {Optimize hyperparameters\\on subset $\mathcal{S}$};

\node[bluebox, below=of train_sub] (barrier)
  {Augmented MLL $\mathcal{L}_{\text{eff}}$:\\$-\mu\log(\lambda_{\max} - \log\sigma_f^2)$};

\node[decision, below=of barrier] (hod_check)
  {MAP\\stable?};

\node[box, right=20mm of hod_check] (grow_sub)
  {Increase $M_{\text{sub}}$\\retry (up to $3\times$)};

\node[bluebox, below=of hod_check] (predict)
  {Predict with full $\mathcal{D}$\\using optimized $\boldsymbol{\theta}$};

\node[bluebox, below=16mm of predict] (rotate_gp)
  {Rotate $+$ translate dimer\\on GP surface};

\node[decision, below=of rotate_gp] (inner_conv)
  {GP forces\\$< T_{\text{GP}}$?};

\node[box, below=16mm of inner_conv] (emd_dist)
  {Compute $d_{\text{EMD}}(\mathbf{x}_{\text{cand}}, \mathbf{x}_{\text{nn}})$};

\node[decision, below=of emd_dist] (trust_ok)
  {$d_{\text{EMD}} \leq \Theta$?};

\begin{scope}[on background layer]
  \node[grp, fit=(fps)(train_sub)(barrier)(hod_check)(grow_sub)(predict),
        label={[lbl, anchor=south west, yshift=-2pt]above right:OT-GP Training Pipeline}] (cl_train) {};
  \node[grp, fit=(rotate_gp)(inner_conv),
        label={[lbl, anchor=south west, yshift=-2pt]above right:Inner Loop: Dimer on $V_{\text{GP}}$}] (cl_inner) {};
  \node[grp, fit=(emd_dist)(trust_ok),
        label={[lbl, anchor=south west, yshift=-2pt]above right:Adaptive Trust Region (EMD-based)}] (cl_trust) {};
\end{scope}

\node[note, left=20mm of fps] (fm1)
  {FM1$+$FM3:\\Scaling $+$ Permutation};

\node[note, left=20mm of barrier] (fm4)
  {FM4:\\Variance explosion};

\node[note, left=20mm of hod_check] (fm2)
  {FM2:\\Hyperparameter instability};

\node[note, left=20mm of trust_ok] (fm_all)
  {All failure\\modes};

\draw[arr] (oracle_eval) -- (true_conv);
\draw[arr] (true_conv) -- node[above, lbl] {Yes} (done);
\draw[arr] (true_conv) -- node[right, lbl] {No} (add_data);
\draw[arr] (add_data) -- (prior_mean);
\draw[arr] (prior_mean) -- (fps);
\draw[arr] (fps) -- (train_sub);
\draw[arr] (train_sub) -- (barrier);
\draw[arr] (barrier) -- (hod_check);
\draw[arr] (hod_check) -- node[right, lbl] {Yes} (predict);
\draw[arr] (predict) -- (rotate_gp);
\draw[arr] (rotate_gp) -- (inner_conv);
\draw[arr] (inner_conv) -- node[right, lbl] {Yes} (emd_dist);
\draw[arr] (emd_dist) -- (trust_ok);

\draw[arr] (trust_ok.east) -- ++(30mm,0)
  node[above, lbl, pos=0.3] {Step accepted}
  |- (oracle_eval.east);

\draw[back] (hod_check) -- node[above, note] {No} (grow_sub);
\draw[back] (grow_sub) |- (fps);

\draw[back] (inner_conv.west) -- +(-14mm,0)
  node[above, note, pos=0.4] {No}
  |- (rotate_gp.west);

\draw[back] (trust_ok.west) -- +(-25mm,0)
  node[below, note, pos=0.5] {Rejected}
  node[below=3mm, note, pos=0.5] {(targeted acquisition)}
  |- (oracle_eval.west);

\draw[fmlink] (fm1) -- (fps);
\draw[fmlink] (fm4) -- (barrier);
\draw[fmlink] (fm2) -- (hod_check);
\draw[fmlink] (fm_all) -- (trust_ok);

\end{tikzpicture}%
}%
\caption{Decision flow of the OT-GP framework. The training pipeline (FPS, MAP regularization) and adaptive trust region (EMD-based) address the failure modes of the basic GP-dimer. The optional prior-mean branch shown here is included to place later extensions in the same design space, not to redefine the main algorithmic thread discussed in the text.}
\label{fig:otgp_pipeline}
\end{figure}
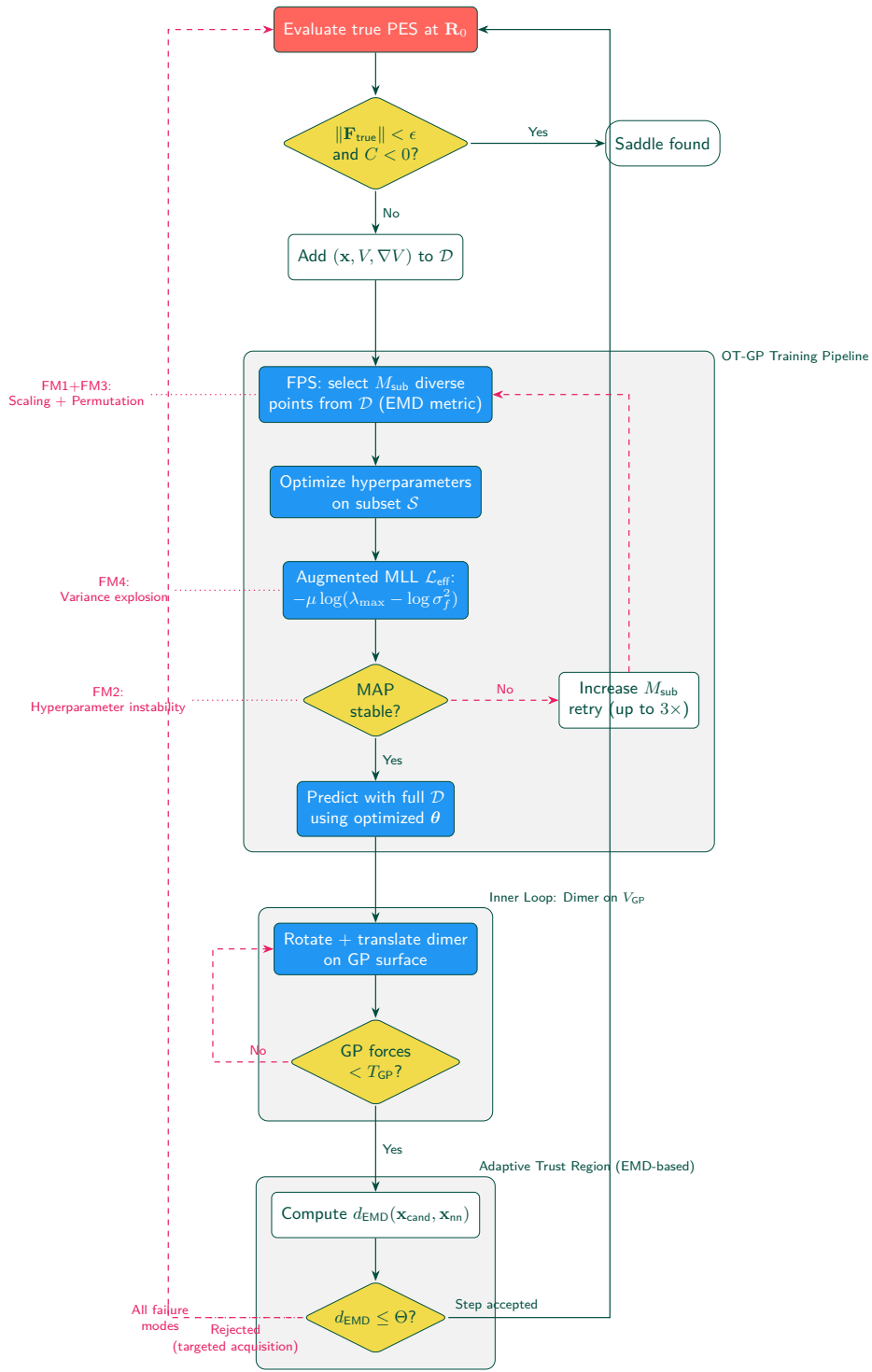

Each of these components operates at the level of GP construction and hyperparameter management, not at the level of the optimizer itself.
A dimer, a NEB, and a local minimizer all build a GP from accumulated data, re-optimize hyperparameters at each step, and propose new configurations on the surrogate.
They all inherit the same failure modes, and they all benefit from the same stabilization mechanisms.
We develop the per-bead FPS extension to NEB explicitly in Section \ref{sec:fps}; MAP regularization and the adaptive trust radius apply to any method without modification.
Together, these changes reduce the failure rate from approximately 12\% to approximately 2\% across 500+ benchmark reactions \cite{goswamiAdaptivePruningIncreased2025}.
Figure \ref{fig:otgp_pipeline} shows the decision flow of the full pipeline.

\subsection{Farthest Point Sampling with Earth Mover's Distance}
\label{sec:fps}
As the search progresses and more electronic structure calculations are performed, the training set grows and the covariance matrix inversion becomes the dominant cost.
For a system with \(N_{\text{atoms}}\) atoms and \(M_{\text{data}}\) collected configurations, the hyperparameter optimization involves repeated inversions at cost \(\mathcal{O}((M_{\text{data}} \cdot N_{\text{atoms}})^3)\).

The fix is to decouple hyperparameter optimization from prediction.
We optimize hyperparameters on a small, geometrically spread-out subset \(\mathcal{S} \subset \mathcal{X}\), chosen by farthest point sampling (FPS), while still using all collected data \(\mathcal{X}\) for prediction \cite{deComparingMoleculesSolids2016,imbalzanoAutomaticSelectionAtomic2018}:

\begin{equation}
\mathbf{x}_{\text{next}} = \arg\max_{\mathbf{x}_i \in \mathcal{X} \setminus \mathcal{S}} \left[\min_{\mathbf{x}_j \in \mathcal{S}} D(\mathbf{x}_i, \mathbf{x}_j)\right]
\label{eq:fps}
\end{equation}

\begin{figure}[htbp]
\centering
\includegraphics[width=\textwidth]{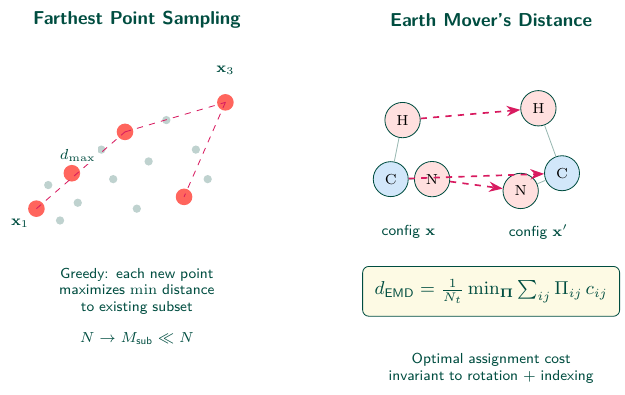}
\caption{\label{fig:fps_emd_concept}(\emph{Left}) Farthest point sampling selects a geometrically spread-out subset (coral) from the full training set (gray) by greedily maximizing the minimum distance to the existing selection. (\emph{Right}) The Earth mover's distance measures structural dissimilarity between two molecular configurations as the optimal transport cost of matching their atom-pair distance distributions; it is invariant to rotation and atom indexing.}
\end{figure}

Figure \ref{fig:fps_emd_concept} shows the FPS selection rule and the EMD-based structural comparison used for molecular configurations.

At each step, FPS picks the training point that is farthest from everything already selected, and repeats until \(M_{\text{sub}}\) points are chosen.
Hyperparameters are optimized on this subset \(\mathcal{S}\), but the full dataset \(\mathcal{X}\) is used for GP prediction.
This bounds the optimization cost at \(\mathcal{O}((M_{\text{sub}} \cdot N_{\text{atoms}})^3)\) with \(M_{\text{sub}} \ll M_{\text{data}}\).
Two details matter in practice.
First, the two most recent configurations are always forced into \(\mathcal{S}\), regardless of their FPS rank, so that the hyperparameter estimates remain relevant to the current surrogate neighborhood.
Second, 10 points is a good starting size for \(M_{\text{sub}}\), but if the MAP estimate is unstable (detected by the oscillation monitor in Section \ref{sec:map-regularization}), the subset grows adaptively up to \(M_{\text{sub}} = 30\).
The growth is triggered by a global signal (hyperparameter oscillation) rather than by local predictive variance, because kernel hyperparameters are global PES properties that require geometrically diverse data, whereas high local variance is better resolved by evaluating the true PES at that point (Section \ref{sec:adaptive-trust}).

Extending FPS from the dimer to the NEB requires accounting for the string discretization.
The NEB approximates the continuous MEP (Eq. \ref{eq:mep}) as a chain of \(P+1\) images, each of which samples a different local region of the PES.
Configurations near the reactant minimum occupy a different part of configuration space than those near the saddle point, and the kernel length scales appropriate for one region need not suit the other.
A single global FPS subset across all images mixes configurations from these different PES regions, producing hyperparameter estimates that compromise between them.
The natural solution is to maintain one FPS subset \(\mathcal{S}_i\) per image \(i\), so that each local surrogate draws its hyperparameters from configurations in the relevant neighborhood.
This per-bead structure mirrors the NEB force decomposition itself, and just as the NEB force (Eq. \ref{eq:neb_force}) acts independently on each image's perpendicular subspace, the FPS selects data independently for each image's local GP.
In practice, each image maintains its own \(\mathcal{S}_i\) with the same greedy selection rule (Eq. \ref{eq:fps}) applied to the subset of training data within a cutoff distance of that image in the EMD metric.

This technique differs from sparse GPs \cite{rasmussenGaussianProcessesMachine2006,gramacyLaGPLargeScaleSpatial2016} which introduce \(M \ll N\) pseudo-inputs optimized jointly with hyperparameters, approximating the full posterior at \(\mathcal{O}(M^2 N)\) cost.
That machinery suits the regime of large, static training sets.
In the on-the-fly setting here, the training set starts nearly empty, grows by a few points per iteration, and rarely exceeds a few dozen configurations; re-optimizing inducing point locations at every step would add cost without benefit.
FPS selects only \emph{actually observed} configurations for hyperparameter optimization and retains the full dataset for prediction; no information is discarded.
For problems that grow beyond roughly 100 evaluations, the RFF approach in Section \ref{sec:rff} addresses the crossover to the large-data regime.

Beyond the computational savings, FPS improves the conditioning of the kernel matrix.
Well-separated configurations produce small off-diagonal kernel entries (the SE kernel decays exponentially with distance), which by the Gershgorin circle theorem keeps the eigenvalues of \(\mathbf{K}\) away from zero.
This numerical stability benefit is independent of the cost savings and would justify FPS even if the hyperparameter optimization were cheap.

\begin{figure}[htbp]
\centering
\includegraphics[width=0.8\textwidth]{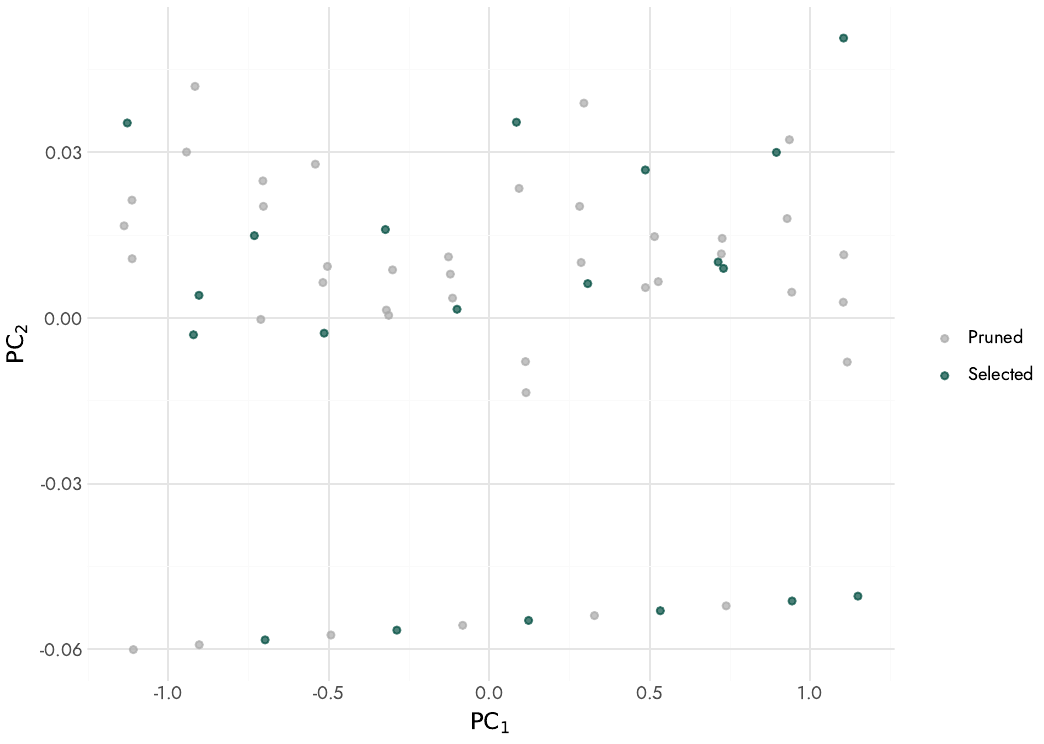}
\caption{\label{fig:leps_fps}Farthest point sampling (FPS) on the LEPS surface. Approximately 50 candidate configurations from a GP-NEB run are projected onto their first two principal components in inverse-distance feature space. FPS selects 20 points (teal diamonds) that maximally cover the feature space; pruned points (gray circles) lie near already-selected configurations and would add redundancy to the training set without improving kernel matrix conditioning.}
\end{figure}

Figure \ref{fig:leps_fps} illustrates the FPS selection on the LEPS surface.
After a GP-NEB run collects approximately 50 candidate configurations, FPS selects 20 maximally diverse points in inverse-distance feature space, as visualized through PCA projection.
The selected points (teal diamonds) span the feature space uniformly, while the pruned points (gray circles) cluster in already-represented regions.

The distance metric \(D\) in Eq. \ref{eq:fps} determines which configurations the algorithm considers "similar" and which it considers "different."
A bad choice here can sabotage the entire FPS selection.
The 1D-max-log distance (Eq. \ref{eq:max1dlog}) compares atoms by fixed index, and this breaks down whenever chemically equivalent atoms swap positions.
The classic example is a methyl group rotation: three hydrogens rotate by 120 degrees, but because each hydrogen keeps its original index, the fixed-index metric registers a large geometric displacement even though the molecule has barely changed.
The trust region then rejects the configuration as "too far" from the training data, or FPS treats two nearly identical structures as maximally different, wasting a slot in the subset.
What we need is a distance that matches atoms of the same element before measuring displacement.

The Earth Mover's Distance (EMD) \cite{rubnerEarthMoversDistance2000} does exactly this.
For each atom type \(t\), we solve a linear assignment problem that optimally pairs atoms between two configurations to obtain the per-type average displacement:

\begin{equation}
\bar{d}_t = \frac{1}{N_t} \min_{\pi \in \Pi_{N_t}} \sum_{k=1}^{N_t} \|\mathbf{r}_{k,t}^{(1)} - \mathbf{r}_{\pi(k),t}^{(2)}\|
\label{eq:emd_pertype}
\end{equation}

where \(N_t\) is the number of atoms of type \(t\) and \(\Pi_{N_t}\) is the set of all permutations.
The overall distance is the maximum per-type displacement:

\begin{equation}
D(\mathbf{x}_i, \mathbf{x}_j) = \max_t \bar{d}_t(\mathbf{x}_i, \mathbf{x}_j)
\label{eq:emd_overall}
\end{equation}

Two design choices in this definition deserve explanation.
First, the per-element averaging by \(1/N_t\) makes the metric scale-independent: a 5-atom molecule and a 50-atom molecule with the same reactive core produce comparable distance values.
Without this normalization, adding inert atoms to the system (a larger solvent shell, a surface slab with more layers) would shrink the per-atom contribution and make the metric blind to the actual chemical change.
This property is what lets us define a single trust radius threshold that works across systems of different sizes (Section \ref{sec:adaptive-trust}).
Second, the \(\max\) over atom types ensures that a large displacement of \emph{any} chemical species is detected, even if most other atoms are stationary.
Together with the permutation optimization in Eq. \ref{eq:emd_pertype}, which resolves failure mode (3) described earlier, these choices produce a distance that tracks genuine structural change rather than labeling artifacts.

To see the permutation invariance in action, consider a proton (indexed \(k\)) transferring between two chemically equivalent sites (atoms \(m\) and \(n\)).
The initial and final states are energetically degenerate, and the true structural difference is small.
But a fixed-index metric like the 1D-max-log distance (Eq. \ref{eq:max1dlog}) sees atom \(k\) far from its original position and reports a large displacement, because it cannot recognize that relabeling \(m\) and \(n\) would reconcile the structures.
The EMD solves the assignment problem and correctly identifies the proton transfer as a small rearrangement.
In chemgp-core (\texttt{src/emd.rs}), the optimal assignment is computed by the Hungarian algorithm \cite{kuhnHungarianMethodAssignment1955} at \(\mathcal{O}(N_t^3)\) cost, matching the C++ gpr$\backslash$\textsubscript{optim} implementation.
For the small atom groups typical of saddle point searches (\(N_t \leq 20\)), this cost is negligible.

The per-type linear assignment problem (Eq. \ref{eq:emd_pertype}) is a discrete optimal transport problem, which gives the framework its name.
The "earth" being moved is a unit point mass at each atomic position, and the "work" is the total displacement needed to transform one configuration into the other \cite{goswamiAdaptivePruningIncreased2025}.
\subsection{Regularizing the MAP Estimate}
\label{sec:map-regularization}
With few data points the MLL landscape is poorly determined, and the MAP estimate of the hyperparameters exhibits two pathologies.
First, the signal variance \(\sigma_f^2\) grows without bound because the data-fit term in Eq. \ref{eq:marginal_likelihood} dominates the complexity penalty.
The resulting surrogate interpolates the training data exactly and produces steep, unphysical gradients between data points.
Soft penalties (L1 or L2 on \(\sigma_f^2\)) are easily overwhelmed by the MLL gradient, so a hard ceiling is needed.
Second, the full hyperparameter vector oscillates between competing MLL modes at successive iterations: the marginal likelihood landscape shifts every time a new data point arrives or the FPS subset changes, and with a small training set the optimizer has no strong reason to prefer one local minimum over another.
The surrogate surface changes shape erratically, and the search makes no net progress.

We address the first pathology with a logarithmic barrier drawn from interior point methods \cite{boydConvexOptimization2004,potraInteriorpointMethods2000}, augmenting the MLL:

\begin{equation}
\mathcal{L}_{\text{eff}}(\boldsymbol{\theta}) = \log p(\mathbf{y} \mid \mathcal{S}, \boldsymbol{\theta}) - \mu \log(\lambda_{\max} - \log \sigma_f^2)
\label{eq:barrier}
\end{equation}

where \(\lambda_{\max}\) is an absolute upper bound for \(\log \sigma_f^2\) and \(\mu \geq 0\) is the barrier strength, which grows with the training set size so that the GP has room to adapt when data is scarce and progressively less room to inflate its variance as evidence accumulates.
For the second pathology, a sign-reversal diagnostic over a sliding window detects oscillation: when a large fraction of the hyperparameters reverse direction at every step, the algorithm grows the FPS subset size \(M_{\text{sub}}\) (Section \ref{sec:fps}) and re-runs the optimization, up to three retries.
Adding more geometrically diverse data sharpens the MLL landscape and constrains the optimizer.

Both fixes are standard regularization of the MAP estimate: the barrier imposes a hard constraint on one parameter, and the subset growth sharpens the MLL landscape for all parameters.
\subsection{Adaptive Trust Radius}
\label{sec:adaptive-trust}
A fixed trust radius is either too conservative (wasting oracle calls) or too aggressive (producing unphysical geometries).
The natural solution is to let the radius grow with the GP's experience, measured via the intensive EMD (Eq. \ref{eq:emd_overall}) for size-independent thresholds.
A candidate configuration \(\mathbf{x}_{\text{cand}}\) is accepted only if:

\begin{equation}
d_{\text{EMD}}(\mathbf{x}_{\text{cand}}, \mathbf{x}_{\text{nn}}) \leq \Theta(N_{\text{data}}, N_{\text{atoms}})
\label{eq:emd_trust}
\end{equation}

where \(\mathbf{x}_{\text{nn}}\) is the nearest neighbor in the training set and \(\Theta\) follows an exponential saturation curve:

\begin{equation}
\Theta_{\text{earned}}(N_{\text{data}}) = T_{\min} + \Delta T_{\text{explore}} \cdot \left(1 - e^{-k N_{\text{data}}}\right), \quad k = \frac{\ln 2}{N_{\text{half}}}
\label{eq:trust_saturation}
\end{equation}

with a physical ceiling:

\begin{equation}
\Theta_{\text{phys}}(N_{\text{atoms}}) = \max\left(a_{\text{floor}},\, \frac{a_A}{\sqrt{N_{\text{atoms}}}}\right)
\label{eq:trust_ceiling}
\end{equation}
The final threshold is \(\Theta = \min(\Theta_{\text{earned}}, \Theta_{\text{phys}})\).
The earned component (Eq. \ref{eq:trust_saturation}) grows rapidly with the first few data points and saturates, so that late-stage evaluations do not keep inflating the step size.
The physical ceiling (Eq. \ref{eq:trust_ceiling}) scales as \(1/\sqrt{N_{\text{atoms}}}\) because per-atom displacements are smaller in larger systems.
Here \(a_A\) is a characteristic atomic length scale (approximately 1.0 \AA{}, the typical bond length), and \(a_{\text{floor}}\) is a minimum threshold to prevent the trust radius from becoming too small for small systems.
When a proposed step violates the trust radius, the algorithm evaluates the true PES at the rejected configuration and adds the result to the training set, turning the violation into targeted acquisition that concentrates the electronic structure budget near the transition path.
\subsection{Scaling via Random Fourier Features}
\label{sec:rff}
The FPS strategy (Section \ref{sec:fps}) bounds the hyperparameter optimization cost, but prediction still requires the full \(M(1+3N) \times M(1+3N)\) covariance matrix.
For long NEB paths or large systems where the training set grows beyond \(\sim 50\) configurations, even the prediction step becomes a bottleneck.
Random Fourier features (RFF) \cite{rahimiRandomFeaturesLargeScale2007,novelliFastFourierFeatures2025} provide a way to decouple hyperparameter training from prediction, using the per-bead FPS subset for the former and all available data for the latter.

The mathematical basis is Bochner's theorem \cite{rasmussenGaussianProcessesMachine2006}, which states that any stationary kernel is the Fourier transform of a non-negative spectral measure.
For the SE kernel in inverse-distance space, the spectral density is Gaussian:

\begin{equation}
k(\mathbf{x}, \mathbf{x}') = \sigma_f^2 \exp\left(-\sum_{(i,j)} \frac{(\phi_{ij}(\mathbf{x}) - \phi_{ij}(\mathbf{x}'))^2}{l_{\phi(i,j)}^2}\right) = \sigma_f^2 \int p(\boldsymbol{\omega}) e^{i\boldsymbol{\omega}^T(\boldsymbol{\phi}(\mathbf{x}) - \boldsymbol{\phi}(\mathbf{x}'))} d\boldsymbol{\omega}
\label{eq:bochner}
\end{equation}

where \(p(\boldsymbol{\omega}) = \mathcal{N}(\mathbf{0}, 2\,\text{diag}(1/l_{\phi(i,j)}^2))\).
Drawing \(D_{\text{rff}}\) frequency vectors \(\boldsymbol{\omega}_m \sim p(\boldsymbol{\omega})\) and random phases \(b_m \sim \text{Uniform}[0, 2\pi)\), the kernel is approximated by an inner product of finite-dimensional feature vectors:

\begin{equation}
k(\mathbf{x}, \mathbf{x}') \approx \mathbf{z}(\mathbf{x})^T \mathbf{z}(\mathbf{x}'), \quad z_m(\mathbf{x}) = \sigma_f \sqrt{\frac{2}{D_{\text{rff}}}} \cos(\boldsymbol{\omega}_m^T \boldsymbol{\phi}(\mathbf{x}) + b_m)
\label{eq:rff_features}
\end{equation}

This converts the GP from a kernel machine into a Bayesian linear regression problem in the \(D_{\text{rff}}\)-dimensional feature space.
Training reduces to solving a linear system.

\begin{equation}
\boldsymbol{\alpha} = (\mathbf{Z}^T \boldsymbol{\Lambda} \mathbf{Z} + \mathbf{I})^{-1} \mathbf{Z}^T \boldsymbol{\Lambda} \mathbf{y}
\label{eq:rff_blr}
\end{equation}

where \(\mathbf{Z}\) is the \(n_{\text{obs}} \times D_{\text{rff}}\) design matrix and \(\boldsymbol{\Lambda} = \text{diag}(1/\sigma_E^2, \ldots, 1/\sigma_F^2, \ldots)\) contains the observation precisions. The cost is \(\mathcal{O}(n_{\text{obs}} \cdot D_{\text{rff}} + D_{\text{rff}}^3)\), which replaces the exact GP cost of \(\mathcal{O}(n_{\text{obs}}^3)\). For \(n_{\text{obs}} = 400\) and \(D_{\text{rff}} = 200\), this is roughly a 1000\(\times\) speedup. The predictive variance retains a closed form, \(\text{var}[f(\mathbf{x}_*)] = \mathbf{z}_*^T (\mathbf{Z}^T \boldsymbol{\Lambda} \mathbf{Z} + \mathbf{I})^{-1} \mathbf{z}_*\), preserving the uncertainty-based acquisition that drives the active learning loop.

Derivative observations enter naturally through the chain rule.
The Jacobian of the RFF feature vector with respect to Cartesian coordinates is:

\begin{equation}
\frac{\partial z_m}{\partial x_a} = -\sigma_f \sqrt{\frac{2}{D_{\text{rff}}}} \sin(\boldsymbol{\omega}_m^T \boldsymbol{\phi}(\mathbf{x}) + b_m) \sum_{(i,j)} \omega_{m,(i,j)} \frac{\partial \phi_{ij}}{\partial x_a}
\label{eq:rff_jacobian}
\end{equation}

where \(\partial \phi_{ij} / \partial x_a\) is the inverse-distance Jacobian (Eq. \ref{eq:invdist_jacobian}).
Each gradient observation contributes a row of \(\mathbf{Z}\) via this Jacobian, so the design matrix has the same blocked structure (energy rows, then force rows) as the full covariance matrix.
The difference is that the matrix dimensions are \(n_{\text{obs}} \times D_{\text{rff}}\) rather than \(n_{\text{obs}} \times n_{\text{obs}}\), and \(D_{\text{rff}}\) is a user-chosen constant that does not grow with the data.

RFF fitting and prediction operate in the inverse-distance feature space \(\boldsymbol{\phi}(\mathbf{x})\), not in Cartesian coordinates.
The random frequencies \(\boldsymbol{\omega}_m\) come from the spectral density of the SE kernel in feature space (Eq. \ref{eq:bochner}), and the design matrix \(\mathbf{Z}\) evaluates the cosine features at \(\boldsymbol{\phi}(\mathbf{x})\) rather than at \(\mathbf{x}\).
The SE kernel remains stationary in \(\boldsymbol{\phi}\), but the composite kernel \(k(\boldsymbol{\phi}(\mathbf{x}), \boldsymbol{\phi}(\mathbf{x}'))\) loses Cartesian stationarity because \(\boldsymbol{\phi}\) depends nonlinearly on \(\mathbf{x}\).
Bochner's theorem still applies in the feature space where stationarity holds, and Cartesian gradients follow by composing the RFF gradient in feature space with the inverse-distance Jacobian (Eq. \ref{eq:invdist_jacobian}) through the chain rule, exactly as in Eq. \ref{eq:rff_jacobian}.

The conceptual connection to per-bead FPS is direct.
In the exact GP, FPS selects a subset for hyperparameter optimization; all data is used for prediction, but the prediction cost is cubic in the total data size.
RFF takes the separation one step further.
The hyperparameters (which determine the spectral density \(p(\boldsymbol{\omega})\)) are still optimized on the FPS subset at \(\mathcal{O}(M_{\text{sub}}^3)\) cost, and then the RFF model is built using \emph{all} training data at the lower \(\mathcal{O}(n_{\text{obs}} \cdot D_{\text{rff}})\) cost.
This two-stage strategy, hyperparameters from a subset, prediction from the full set, exploits the structural insight that kernel hyperparameters are global properties of the PES that can be estimated from a diverse subset, while prediction accuracy benefits from every available data point.

The division of labor is: FPS controls \emph{which} data enters the hyperparameter optimization (bounding its \(\mathcal{O}(M_{\text{sub}}^3)\) cost), while RFF controls \emph{how} prediction is performed on the full dataset (replacing the \(\mathcal{O}(M^3)\) exact solve with an \(\mathcal{O}(M \cdot D_{\text{rff}})\) linear regression).
The two mechanisms are orthogonal and can be enabled independently, though they are most beneficial in combination.

The required \(D_{\text{rff}}\) depends on the dimensionality of the inverse-distance feature space (i.e., the number of atom pairs \(N_{\text{pairs}} = N(N-1)/2\)), because the RFF must approximate the SE kernel in this space (Eq. \ref{eq:bochner}).
For 2D model surfaces (3 atoms, 3 inverse-distance features), \(D_{\text{rff}} \sim 50\textrm{--}100\) suffices.
For a 9-atom molecule (36 inverse-distance features), \(D_{\text{rff}} \sim 500\) is needed for the AIE variant to converge reliably; lower values (e.g., 300) introduce sufficient approximation error to stall the climbing-image convergence.
As a practical rule, \(D_{\text{rff}} \gtrsim 10 \, N_{\text{pairs}}\) provides a reasonable starting point, though the exact threshold depends on the kernel length-scale spectrum and should be verified by comparing RFF predictions against the exact GP on held-out test points.

\begin{figure}[htbp]
\centering
\includegraphics[width=0.8\textwidth]{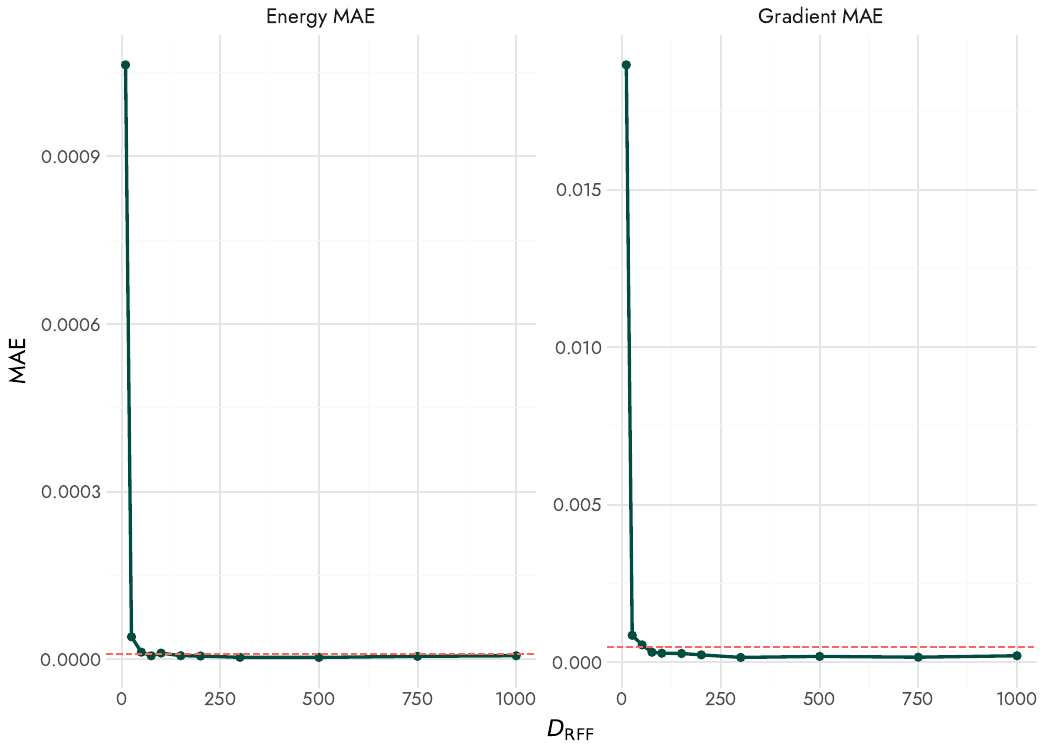}
\caption{\label{fig:leps_rff_quality}RFF approximation quality on the LEPS surface (H\(_3\), 3 inverse-distance features). Energy MAE (top) and gradient MAE (bottom) between RFF and exact GP predictions on held-out test points, plotted against \(D_{\text{rff}}\). The kernel is well approximated by \(D_{\text{rff}} \sim 100\).}
\end{figure}

Figure \ref{fig:leps_rff_quality} quantifies the RFF approximation quality on the LEPS surface (3 atoms, 3 inverse-distance features).
The energy and gradient MAE between RFF and exact GP predictions are plotted against \(D_{\text{rff}}\) for held-out test configurations.
The approximation error drops below \(10^{-4}\) eV by \(D_{\text{rff}} = 100\) and continues to decrease monotonically, confirming that low-dimensional kernels are well approximated by modest numbers of random features.

The computational savings from RFF scale favorably.
For a training set of \(M\) configurations with \(N\) atoms each, the exact GP prediction requires \(\mathcal{O}((M(1+3N))^2)\) operations per test point (matrix-vector products with the inverse covariance), while RFF prediction costs \(\mathcal{O}(D_{\text{rff}} \cdot N_{\text{pairs}})\).
For a typical system with \(M = 50\) training points, this is a reduction from \(\sim 10^7\) to \(\sim 10^4\) operations per inner-loop prediction.
In the NEB context, where each outer iteration may involve hundreds of inner-loop evaluations across \(P\) images, this per-evaluation speedup translates to a significant wall-clock reduction.
The hyperparameter optimization remains the dominant cost, but because FPS bounds that at \(\mathcal{O}(M_{\text{sub}}^3)\) with \(M_{\text{sub}} \sim 10\textrm{--}30\), the combined FPS+RFF strategy keeps the total GP overhead well below the electronic structure cost at each outer iteration.

The RFF extension presented here extends the OT-GP framework to larger systems across the Dimer, NEB and minimization.
In practice, molecular systems benefit from a reduced GP tolerance divisor (e.g. 3 rather than 10) compared to toy surfaces, because the higher-dimensional surrogate is less reliable in extrapolation regions.
Enabling RFF prediction in the OTGPD inner loop further stabilizes molecular runs by smoothing the surrogate away from training data, suppressing the oscillations that exact GP prediction can exhibit when the inverse-distance feature space is sparsely sampled.
\subsection{Implementation: Cholesky Solver and Adaptive Jitter}
\label{sec:linalg}
Every prediction and every step of MAP-NLL hyperparameter optimization
requires solving the linear system
\begin{equation}
  \mathbf{K}\,\boldsymbol{\alpha} = \mathbf{y},
  \label{eq:gp-linear-system}
\end{equation}
where \(\mathbf{K}\) is the augmented covariance matrix of dimension
\(M(1+3N)\) (energies and forces stacked as in
Eq.\textasciitilde{}\ref{eq:full_covariance}) and \(\mathbf{y}\) holds the corresponding
observations.
A naive matrix inverse \(\boldsymbol{\alpha} = \mathbf{K}^{-1}\mathbf{y}\)
is both numerically unstable and unnecessarily expensive
\cite{rasmussenGaussianProcessesMachine2006}.
For a symmetric positive-definite \(\mathbf{K}\) the standard recipe is
the Cholesky factorization
\begin{equation}
  \mathbf{K} = \mathbf{L}\mathbf{L}^{\top},
  \quad \mathbf{L} \text{ lower-triangular},
  \label{eq:cholesky}
\end{equation}
followed by two triangular solves: forward substitution
\(\mathbf{L}\,\mathbf{z} = \mathbf{y}\) followed by back substitution
\(\mathbf{L}^{\top}\boldsymbol{\alpha} = \mathbf{z}\).
The factorization itself costs \(\mathcal{O}(M^3)\) flops with a small
constant; the two triangular solves are \(\mathcal{O}(M^2)\) each.
The Cholesky factor \(\mathbf{L}\) also gives the log determinant for free,
\(\log \det \mathbf{K} = 2\sum_i \log L_{ii}\), which is the dominant term in
the negative log marginal likelihood that drives hyperparameter
optimization.
The implementation in \texttt{chemgp-core} calls the \texttt{faer::llt} routine in the
Rust \texttt{covariance.rs} module and reuses \(\mathbf{L}\) for both the linear
solve and the log-determinant computation.

Two failure modes break this recipe in practice.
First, as the dataset fills in, nearby configurations contribute nearly
collinear rows to \(\mathbf{K}\) and the smallest eigenvalue can fall below
the working precision, so that
\(\mathbf{K}\) is positive definite analytically but indefinite in finite
arithmetic.
Second, during MAP-NLL optimization the proposed hyperparameters
themselves can drive \(\mathbf{K}\) close to singular long before the data
do, especially when \(\sigma_f^2\) becomes large or a length scale collapses.
The textbook remedy is to add a small multiple of the identity to the
diagonal of \(\mathbf{K}\): write
\begin{equation}
  \widetilde{\mathbf{K}}(\eta) = \mathbf{K} + \eta\,\mathbf{I},
  \label{eq:jitter}
\end{equation}
and choose \(\eta \ge 0\) just large enough that the Cholesky factorization
of \(\widetilde{\mathbf{K}}(\eta)\) succeeds.
The added \(\eta\) acts as Tikhonov regularization on
Eq.\textasciitilde{}\ref{eq:gp-linear-system} and shifts the spectrum of \(\mathbf{K}\) by
\(\eta\) without changing its eigenvectors.

The OT-GP refinement here is to set \(\eta\) \emph{adaptively} rather than to a
fixed constant.
The implementation starts with
\(\eta_0 = 10^{-8}\,\max(\mathrm{diag}(\mathbf{K}))\), scaled to the matrix
so that the same code is meaningful across the eV (molecular) and
reduced-unit (model surface) regimes.
If the Cholesky of \(\widetilde{\mathbf{K}}(\eta_0)\) fails, the jitter is
increased geometrically,
\(\eta_{k+1} = 10\,\eta_k\), and the factorization is retried, up to a
configurable retry limit.
The first \(\eta_k\) that succeeds is the regularizer used for that step.
This gives a one-line answer to the reviewer-2 question of a \emph{lower}
bound on the regularizer: the bound is the smallest \(\eta_k\) in this
geometric ladder that keeps \(\widetilde{\mathbf{K}}(\eta_k)\) positive
definite, which is set by the data and the current hyperparameters rather
than fixed in advance.
We tried fixed lower bounds and found the adaptive ladder more robust
across the regimes the tutorial covers.

A complementary safeguard works at the hyperparameter optimization level
rather than on \(\mathbf{K}\) directly.
The MAP-NLL surface develops a runaway-variance basin in which the
optimizer pushes \(\sigma_f^2 \to \infty\) to absorb residual error;
without a check, the SCG optimizer will follow it.
The barrier of Eq.\textasciitilde{}\ref{eq:barrier} caps \(\log \sigma_f^2\) at
\(\lambda_{\max} = \ln(2)\), and the negative log-likelihood evaluation
returns \(+\infty\) whenever this upper bound is violated.
Combined with the adaptive jitter on the linear-algebra side, the GP
training loop never aborts on a numerically singular \(\mathbf{K}\): the
solver finds a viable \(\eta_k\), the NLL evaluation either returns a
finite value or a sentinel \(+\infty\), and the SCG optimizer is steered
back into the feasible region.
\section{Illustrative Examples}
\label{sec:examples}
The examples in this section are intended to teach three different aspects of the unified loop: how the surrogate behaves on a surface that can be visualized directly, how the acquisition logic changes between path-search variants, and how the same machinery carries over to a realistic molecular potential.
They are not presented as a stand-alone validation suite; larger benchmark studies, repeated-run statistics, and broader molecular validation are reported in the companion production papers \cite{goswamiEfficientImplementationGaussian2025,goswamiAdaptivePruningIncreased2025}.
The Muller-Brown and LEPS examples are therefore retained as didactic controls, not as molecular ground truths for the inverse-distance representation itself.
In the code these toy surfaces are handled through the Cartesian kernel path rather than the molecular inverse-distance kernel, so their oracle-call counts are useful for illustrating the Bayesian loop and for showing negative cases, but not for judging the molecular speedups associated with the inverse-distance covariance.

The chemgp-core crate includes three toy potentials (in \texttt{src/potentials.rs}: Muller-Brown, LEPS, and Lennard-Jones) that illustrate the algorithms on two-dimensional surfaces where the GP behavior can be visualized directly.
They are pedagogically useful precisely because they admit direct plotting, but they should not be conflated with the molecular inverse-distance-kernel setting that motivates the production benchmark claims.
\subsection{Muller-Brown Potential}
\label{sec:orgb896a91}
The Muller-Brown surface \cite{mullerLocationSaddlePoints1979} (Figure \ref{fig:mb_neb}) is a standard benchmark for saddle point searches, with three minima and two saddle points connected by curved MEPs.
Because the surface is two-dimensional, the GP behavior can be visualized directly: Figure \ref{fig:mb_gp_progression} shows how the surrogate evolves from a crude approximation with few training points to a faithful replica of the true PES as data accumulates.
The variance landscape (Figure \ref{fig:mb_variance}) guides the active learning criterion, concentrating evaluations near the reaction path.
The GP surface closely matches the true PES within the trust region (Figure \ref{fig:mb_trust_region}), but diverges outside it, which is the expected behavior of a local surrogate.
The NEB path connecting minima A and B through saddle S2 is shown in Figure \ref{fig:mb_neb}.
\subsection{LEPS Potential}
\label{sec:org2a50e3b}
The London-Eyring-Polanyi-Sato surface \cite{satoNewMethodDrawing1955} (Figure \ref{fig:leps_neb}) models a collinear atom-transfer reaction \(A + BC \to AB + C\).
The MEP has a single saddle point with pronounced curvature and is an ideal test for the GP-NEB (Figure \ref{fig:leps_neb}).
Classical NEB converges in 156 oracle evaluations and AIE in 100.
The OIE variant converges in 40 outer iterations (42 total evaluations including endpoints), making this the cleanest toy example for seeing how one-image acquisition changes the outer-loop accounting without changing the underlying NEB mechanics (Figure \ref{fig:leps_aie_oie}).
\subsection{PET-MAD Molecular System}
\label{sec:org947791b}
\begin{figure}[htbp]
\centering
\includegraphics[width=0.8\textwidth]{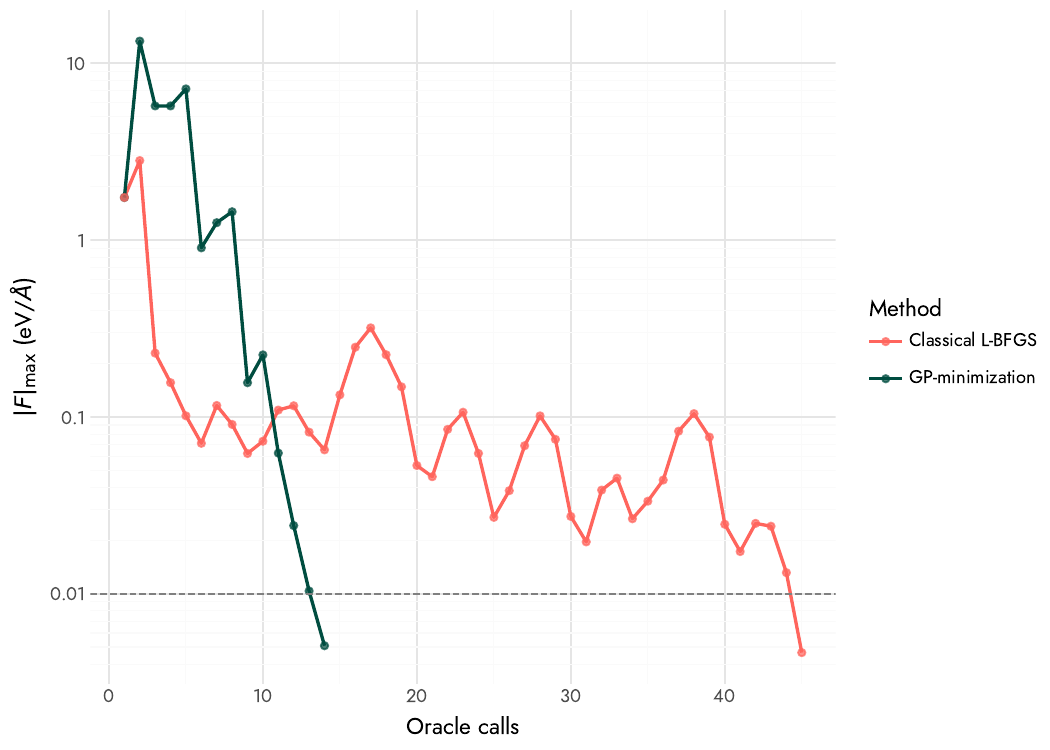}
\caption{\label{fig:petmad_minimize_convergence}Convergence of GP-accelerated minimization on a real molecular system (PET-MAD potential). The maximum per-atom force is plotted against the number of oracle evaluations on a logarithmic scale.}
\end{figure}

The PET-MAD universal potential \cite{mazitovPETMADLightweightUniversal2025}, trained on the MAD dataset \cite{mazitovMassiveAtomicDiversity2025}, provides a realistic test beyond toy surfaces where the inverse-distance kernel operates on physical interatomic distances.
For saddle point searches, Figure \ref{fig:rpc_dimer_convergence} compares the standard dimer, GP-dimer, and OTGPD on the C\textsubscript{3H}\textsubscript{5} allyl radical (8 atoms, 24 DOF): the point of the figure is not that every GP-flavored variant wins uniformly, but that the shared surrogate loop is usable on a genuine molecular PES and that the OT-GP refinements suppress the retraining oscillations visible in the simpler GP-dimer.
Figure \ref{fig:petmad_minimize_convergence} then shows the same loop applied to local minimization on a molecular system, where the maximum per-atom force drops below the convergence threshold with fewer oracle evaluations than the direct optimizer on this example.

\begin{figure}[htbp]
\centering
\includegraphics[width=0.8\textwidth]{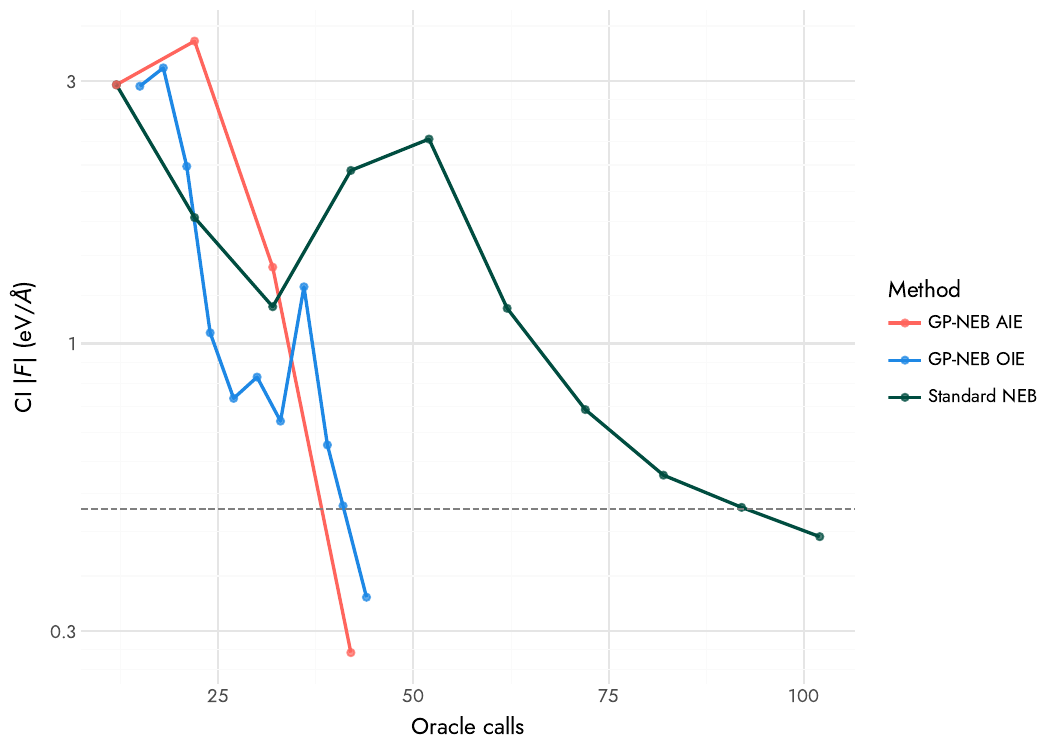}
\caption{\label{fig:system100_convergence}Illustrative convergence comparison of GP-NEB variants on a 9-atom cycloaddition (PET-MAD surface, 27 DOF). Climbing-image force is plotted against oracle evaluations on a logarithmic scale for the classical NEB, AIE, and OIE update patterns. The figure is used here to compare the qualitative behavior of the three outer-loop choices on the same molecular path; the tighter CI-targeted literature-style benchmark is summarized separately in the Supporting Information.}
\end{figure}

\begin{figure}[htbp]
\centering
\includegraphics[width=0.8\textwidth]{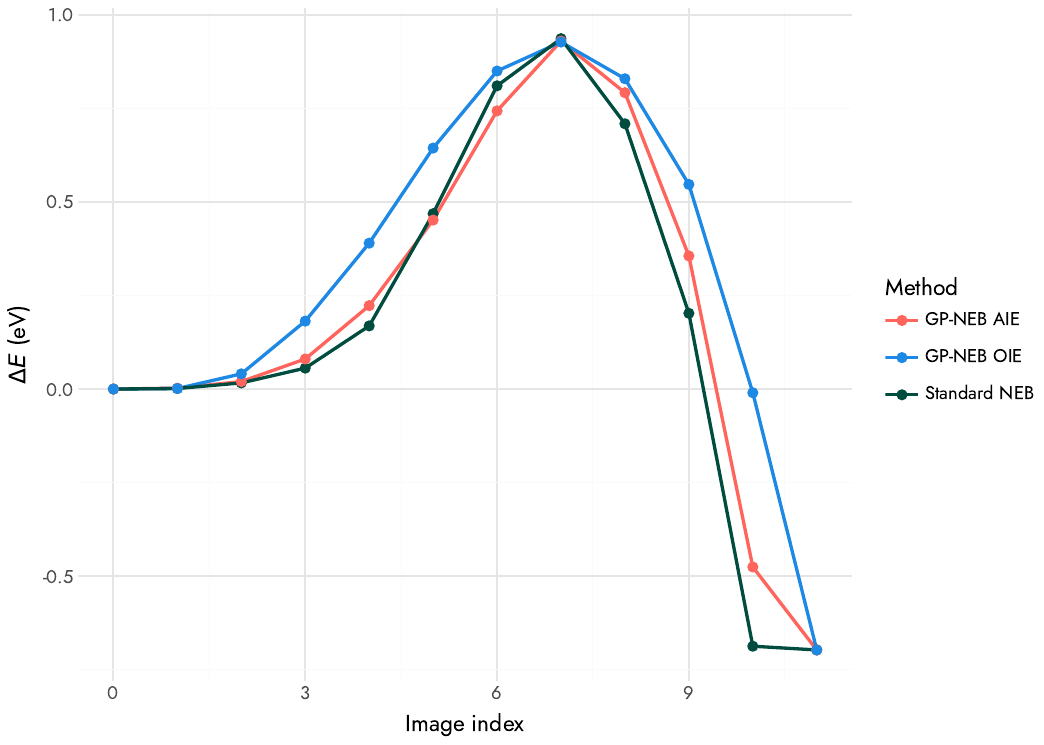}
\caption{\label{fig:system100_neb_profile}Energy profiles along the converged MEP for the cycloaddition system on the PET-MAD surface. All three NEB variants recover the same barrier and exothermic product basin. The AIE profile is slightly shifted near the saddle region but converges to the same endpoints.}
\end{figure}

\begin{figure}[htbp]
\centering
\includegraphics[width=\textwidth]{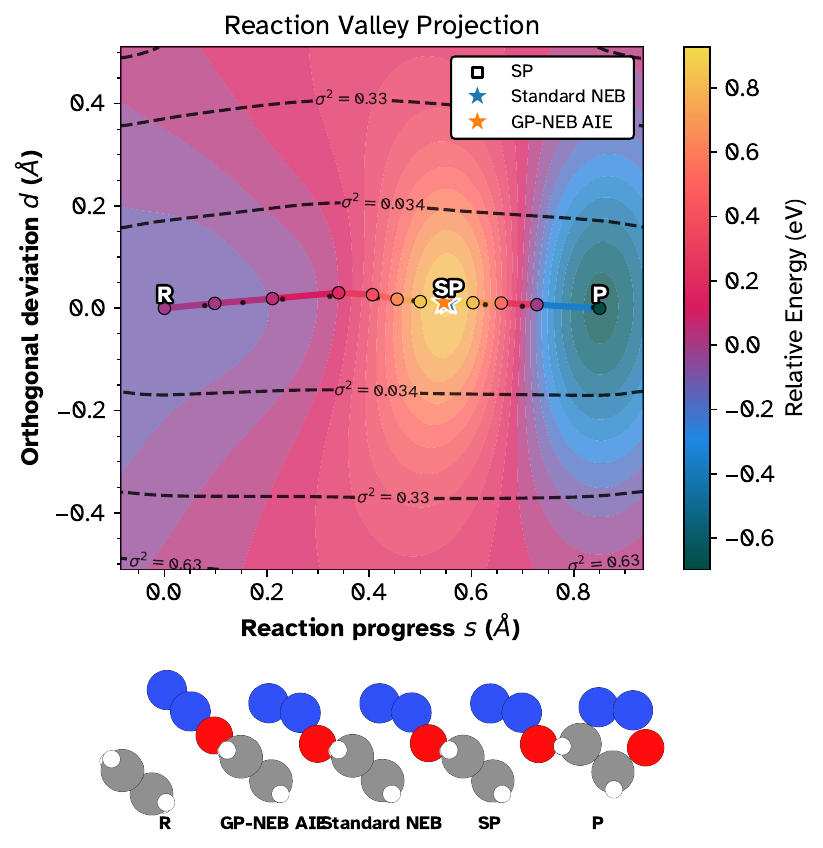}
\caption{\label{fig:system100_neb_landscape}Reaction valley projection \cite{goswamiTwodimensionalRMSDProjections2026} of the NEB paths on the PET-MAD surface. Reaction progress \(s\) and orthogonal deviation \(d\) are projected RMSD coordinates. Standard NEB (blue stars) and GP-NEB AIE (orange stars) saddle points are compared; both lie near the true saddle (black square). Dashed contours show GP variance (\(\sigma^2\)); the paths remain within the well-sampled region where the surrogate has accumulated data, which in combination with the per-step trust region is what keeps them close to the true saddle. Bottom: molecular snapshots at key points along the reaction coordinate.}
\end{figure}

For NEB, a 9-atom cycloaddition system (\(\text{C}_2\text{H}_4 + \text{N}_2\text{O}\), 27 degrees of freedom, 36 inverse-distance features) on the PET-MAD surface illustrates how the GP-NEB variants scale to molecular reactions.
Figure \ref{fig:system100_convergence} compares classical NEB, AIE, and OIE on this system in the tutorial’s illustrative mode, where the goal is to show how the three update patterns behave on the same molecular path rather than to present a stand-alone benchmark campaign.
The tighter literature-aligned CI-based comparison, using the climbing-image force criterion as the stopping metric, is summarized separately in the Supporting Information benchmark table.
Figure \ref{fig:system100_neb_profile} shows the energy profiles along the converged MEP; all three variants recover the same barrier and exothermic product basin, which is the pedagogical point of the example.
Figure \ref{fig:system100_neb_landscape} shows the reaction valley projection \cite{goswamiTwodimensionalRMSDProjections2026} comparing the standard NEB and AIE saddle points; both paths lie within the low-variance region where the surrogate is well-trained, and the molecular snapshots along the bottom illustrate the structural progression from reactant to product.

\section{Conclusions}
\label{sec:conclusions}
Gaussian process regression provides a practical framework for accelerating saddle point searches on potential energy surfaces.
The local surrogate approach builds a GP on-the-fly from electronic structure evaluations during a single search and can substantially reduce the number of expensive oracle calls compared to classical methods, with the largest gains appearing in data-poor, anisotropic saddle-search regimes rather than uniformly across all examples.

The inverse-distance kernel delivers rotational and translational invariance through the feature map \(\phi_{ij} = 1/r_{ij}\), and the learned length-scale parameters automatically identify which interatomic distances govern the reaction.
This feature map also preconditions the PES by homogenizing the effective curvature, enabling accurate interpolation with a stationary kernel despite the wide range of stiffness in molecular systems.
The analytical derivative blocks are essential for numerical stability; automatic differentiation through the inverse-distance computation introduces noise that destroys positive definiteness of the covariance matrix.

The primary goal of the methodology described here is to make that unification explicit.
Surrogate-assisted stationary-point searches admit many valid modalities and implementation choices, but they can still be understood through one Bayesian optimization outer loop.
That viewpoint lets minimization, dimer-based saddle search, CI-NEB, OTGPD refinements, and prior-mean or meta-GP variants be discussed within one tutorial without pretending that every branch should collapse into one code path.

The accompanying Rust code is a reference implementation, with documentation and plotting helpers that expose the implementation choices and bottlenecks rather than hiding them.
Each equation maps to a specific function, and the same binaries run the illustrative examples discussed in the text.

It is expected that Bayesian methods will take root within the community and become an applied aspect of the field, much as optimization techniques have.

The Supporting Information contains algorithm flowcharts for the classical and GP-accelerated dimer and NEB methods, the LEPS marginal-likelihood landscape, hyperparameter defaults, implementation mapping tables, derivation and code details for the GPR, dimer, NEB, trust-region, EMD, and RFF components, and the illustrative benchmark summary used in this tutorial.
\begin{acknowledgement}
The author thanks the anonymous reviewers for comments that improved the clarity of the manuscript. The author thanks Prof. Birgir Hrafnkelsson, Prof. Thomas Bligaard, Dr. Andreas Vishart, and Dr. Miha Gunde for helpful discussions on the methodology. R.G. also acknowledges valuable discussions with Dr. Amrita Goswami, Dr. Moritz Sallermann, Prof. Debabrata Goswami, Mrs. Sonaly Goswami, and Mrs. Ruhila Goswami. Financial support from Dr. Guillaume Fraux and Prof. Michele Ceriotti of Lab-COSMO, EPFL is gratefully acknowledged. The figure color scheme was designed by Ruhila Goswami. The author thanks his family, pets, plants, birds, and garden creatures for their patience and support throughout this work.
\end{acknowledgement}
\section{For Table of Contents Only}
\label{sec:orgfa72648}

\begin{center}
\includegraphics[width=3.25in]{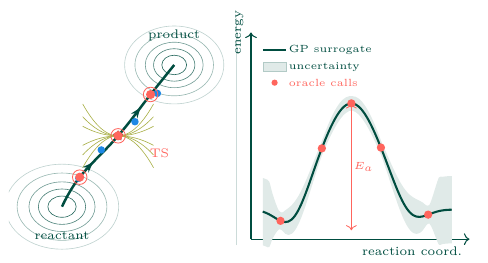}
\end{center}

GP surrogates accelerate saddle point searches on PES.
\label{sec:orgb1169f8}

\begin{appendix}
\section{Algorithm Flowcharts}
\label{sec:org05363cb}

The following flowcharts visualize the classical and GP-accelerated versions of the dimer method and NEB.
Each pair shows the unmodified algorithm (left) alongside its GP-accelerated counterpart (right).

\begin{center}
\centering
\begin{minipage}{0.48\linewidth}
\centering
\includegraphics[width=\linewidth,height=0.32\textheight,keepaspectratio]{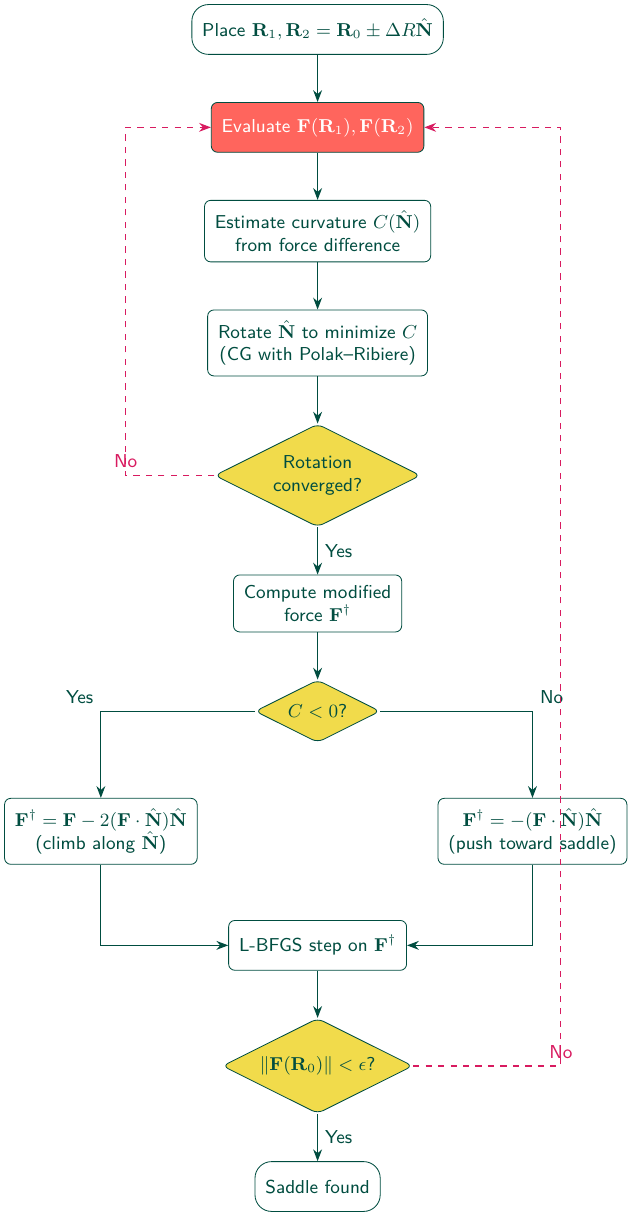}
\end{minipage}\hfill
\begin{minipage}{0.48\linewidth}
\centering
\includegraphics[width=\linewidth,height=0.32\textheight,keepaspectratio]{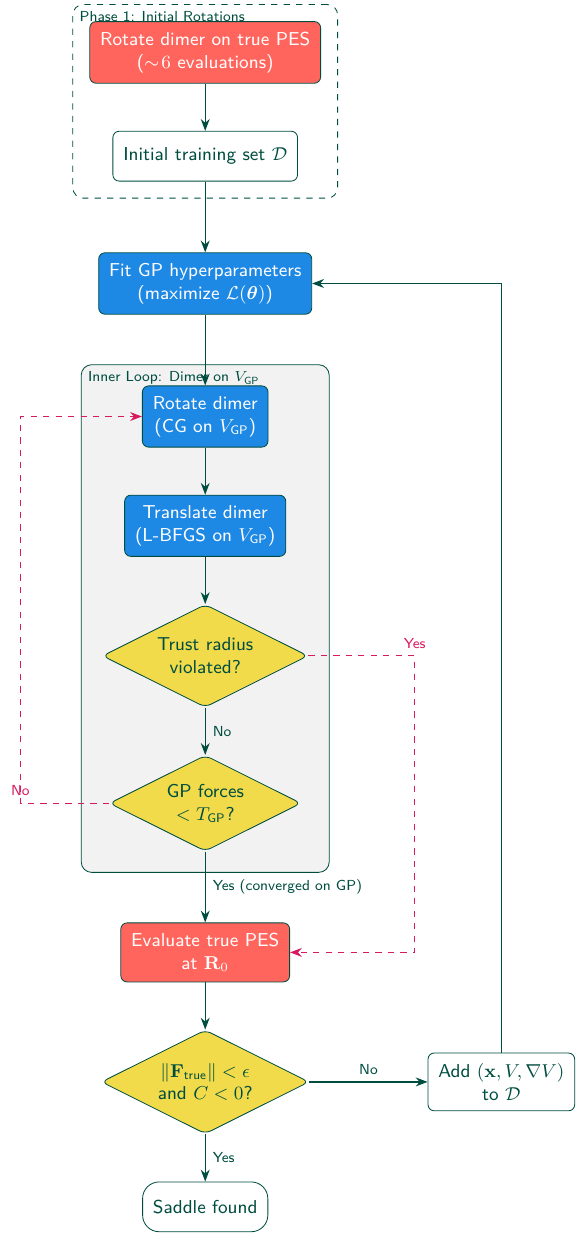}
\end{minipage}
\captionof{figure}{Classical dimer (left) and GP-dimer (right) flowcharts. The classical version evaluates the true PES at every rotation step; the GP-accelerated version performs rotations and translations on the surrogate surface, querying the oracle only when trust radius violations occur or GP-level convergence is achieved.}
\label{fig:si_dimer_flows}
\end{center}

\begin{center}
\centering
\begin{minipage}{0.48\linewidth}
\centering
\includegraphics[width=\linewidth,height=0.32\textheight,keepaspectratio]{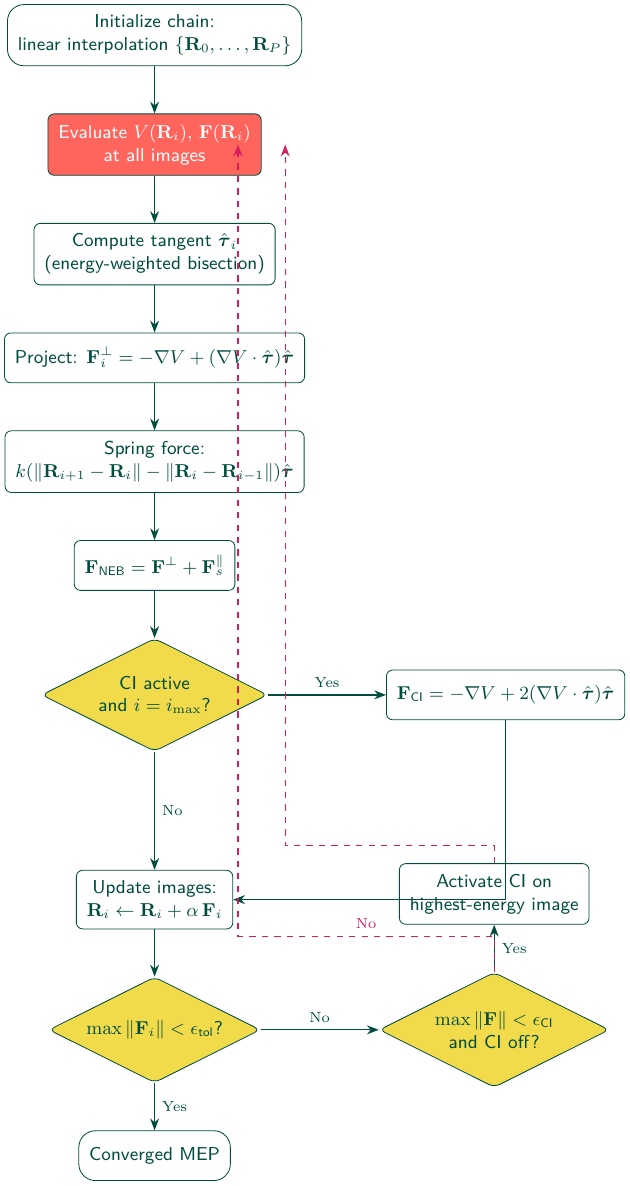}
\end{minipage}\hfill
\begin{minipage}{0.48\linewidth}
\centering
\includegraphics[width=\linewidth,height=0.32\textheight,keepaspectratio]{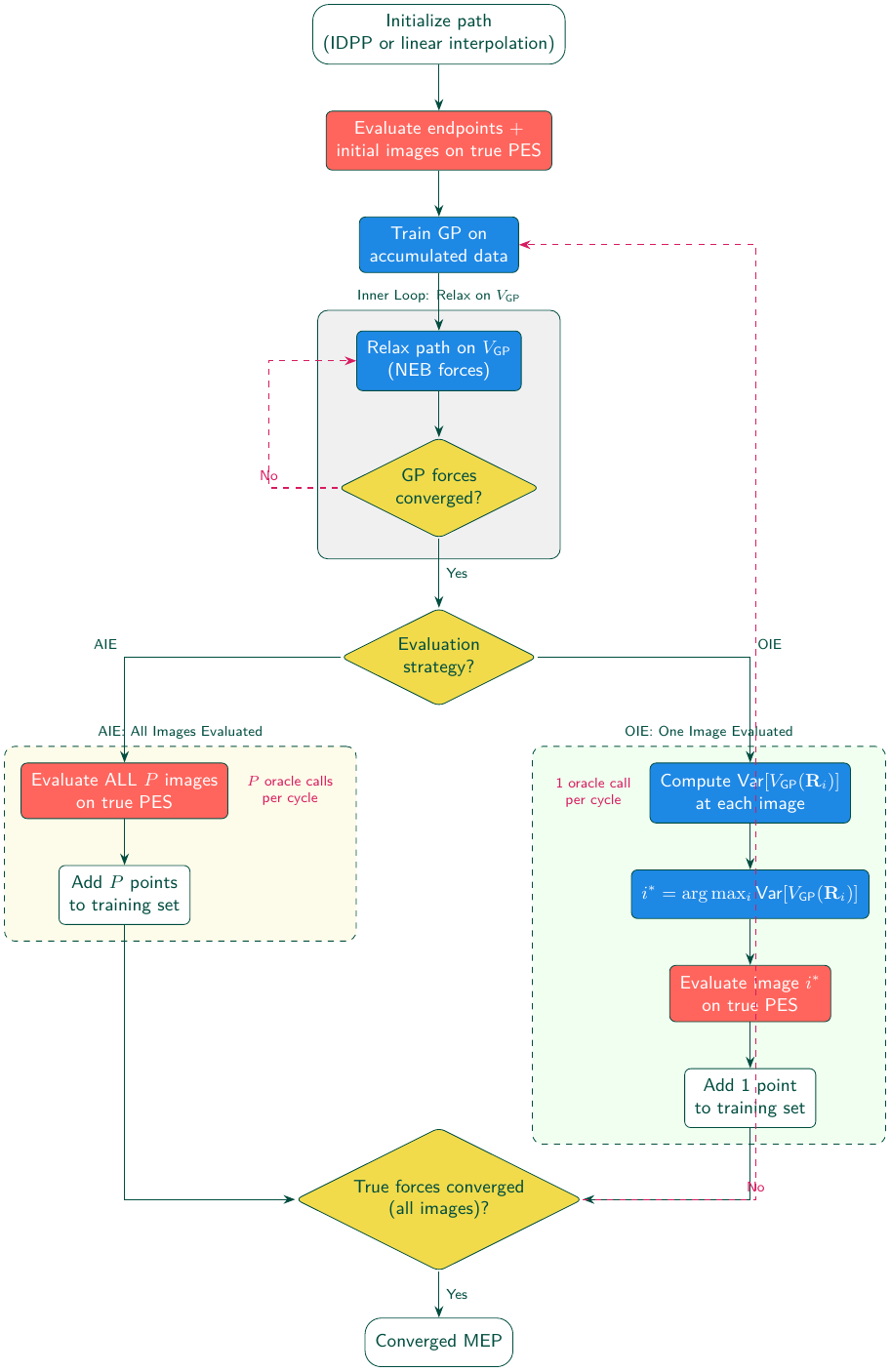}
\end{minipage}
\captionof{figure}{Classical NEB (left) and GP-NEB (right) flowcharts. The classical version optimizes all images via L-BFGS; the GP-NEB variant uses one-image evaluation with a configurable acquisition rule (pure variance or UCB in the current implementation) to reduce oracle queries.}
\label{fig:si_neb_flows}
\end{center}
\section{Marginal Likelihood Landscape}
\label{sec:org85d2865}

\begin{figure}[htbp]
\centering
\includegraphics[width=0.8\textwidth]{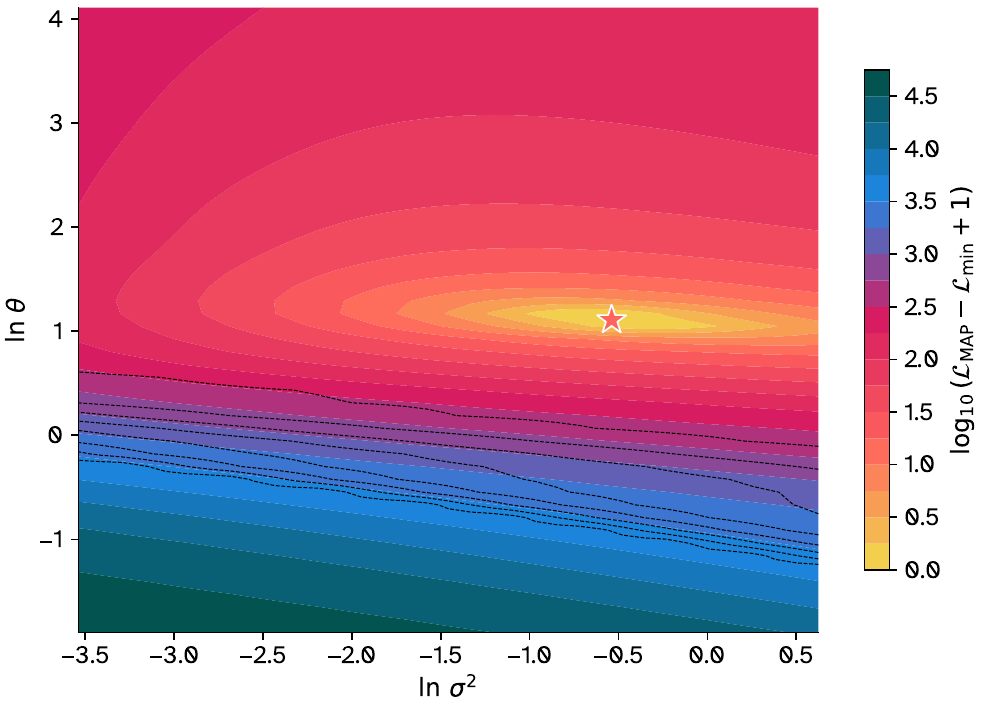}
\caption{\label{fig:si_leps_nll_landscape}MAP-regularized negative log marginal likelihood landscape on the LEPS surface. The filled contours show the NLL as a function of the log-hyperparameters \(\ln\sigma^2\) and \(\ln\theta\), computed on a 40 \(\times\) 40 grid from 5 training points near the reactant. The dashed black contours show the gradient norm. The coral star marks the MAP optimum that SCG converges to. Regions where the Cholesky factorization fails (hyperparameters that produce a non-positive-definite covariance matrix) are masked.}
\end{figure}
\section{Hyperparameter Defaults}
\label{sec:org1e233b5}

Table \ref{tab:si_hyperparameters} collects default values for all hyperparameters used in GP-dimer, GP-NEB, and GP-minimization.
These are starting values tuned for molecular systems with DFT energies (eV scale) and small to medium-sized systems (10-50 atoms).
Adjustments may be needed for model potentials, very large systems, or different energy units.

\begin{table}[!ht]
\centering
\caption{Default hyperparameters for GP-accelerated saddle point searches. Parameters marked with $^\dagger$ are documented in \cite{goswamiAdaptivePruningIncreased2025}.}
\label{tab:si_hyperparameters}
\begingroup
\footnotesize
\setlength{\tabcolsep}{3pt}
\renewcommand{\arraystretch}{0.92}
\begin{tabularx}{\textwidth}{ll>{\raggedright\arraybackslash}X}
\hline
\textbf{Parameter} & \textbf{Default} & \textbf{Meaning} \\
\hline
\multicolumn{3}{l}{\textit{Kernel}} \\
$\sigma_c^2$ & 1.0 & Constant offset (eV-scale energies); set to 0.0 for centered model surfaces \\
$\sigma_f^2$ & Free (MLL) & Signal variance (optimized via MAP-NLL) \\
$l_{\phi(i,j)}$ & Free (MLL) & Inverse-distance length scales (per atom-type pair, optimized) \\
\hline
\multicolumn{3}{l}{\textit{Noise and Regularization}} \\
$\sigma_E^2$ & $10^{-8}$ & Energy noise (Tikhonov regularizer, not physical noise) \\
$\sigma_F^2$ & $10^{-8}$ & Force noise (Tikhonov regularizer) \\
Jitter & $10^{-8} \max(\text{diag}(\mathbf{K}))$ & Initial Cholesky jitter; grows $10\times$ per retry \\
\hline
\multicolumn{3}{l}{\textit{MAP Regularization$^\dagger$}} \\
$\mu_0$ & $10^{-4}$ & Initial logarithmic barrier strength (Eq.~\ref{eq:barrier}) \\
$\alpha$ & $10^{-3}$ & Barrier growth rate with training set size \\
$\mu_{\max}$ & 0.5 & Maximum barrier strength \\
$\lambda_{\max}$ & $\ln(2)$ & Upper bound for $\log \sigma_f^2$ (barrier ceiling) \\
\hline
\multicolumn{3}{l}{\textit{Dimer Method}} \\
$\Delta R$ & 0.01 \AA & Dimer midpoint-to-endpoint separation (\texttt{dimer\_sep}, Eq.~\ref{eq:dimer_images}) \\
$T_{\text{GP}}$ & $10^{-3}$ & GP-level force convergence threshold (eV/\AA) \\
\hline
\multicolumn{3}{l}{\textit{FPS Subset Selection}} \\
$M_{\text{sub,init}}$ & 10 & Initial FPS subset size \\
$M_{\text{sub,max}}$ & 30 & Maximum subset size (after oscillation retries) \\
\hline
\multicolumn{3}{l}{\textit{Trust Radius}} \\
$T_{\min}$ & 0.1 \AA & Minimum trust radius (per-atom EMD) \\
$\Delta T_{\text{explore}}$ & 0.4 \AA & Exploration increment (saturation curve amplitude) \\
$N_{\text{half}}$ & 5 & Half-life for exponential saturation (Eq.~\ref{eq:trust_saturation}) \\
$a_{\text{floor}}$ & 0.3 \AA & Minimum physical ceiling (small systems) \\
$a_A$ & 1.0 \AA & Atomic length scale for size-dependent ceiling (Eq.~\ref{eq:trust_ceiling}) \\
\hline
\multicolumn{3}{l}{\textit{NEB Acquisition}} \\
$\kappa_{\text{UCB}}$ & 2.0 & Exploration weight for Upper Confidence Bound (OIE) \\
\hline
\multicolumn{3}{l}{\textit{Random Fourier Features}} \\
$D_{\text{RFF}}$ & 200 & Number of random features (for large training sets $M > 100$) \\
\hline
\end{tabularx}
\endgroup
\end{table}
\section{Connection to Code: chemgp-core}
\label{sec:code}
The chemgp-core crate (source at \url{https://github.com/lode-org/ChemGP}, documentation at \url{https://lode-org.github.io/ChemGP/}) is the pedagogical reference implementation of the algorithms described in this review.
Each listing below is extracted from the crate source, and the same binary runs the illustrative benchmarks reported throughout.
Production-scale studies use the C++ gpr$\backslash$\textsubscript{optim} code reported in the companion JCTC and ChemPhysChem papers \cite{goswamiEfficientImplementationGaussian2025,goswamiAdaptivePruningIncreased2025}; chemgp-core shares the same algorithmic core and was aligned with gpr$\backslash$\textsubscript{optim} through a ten-fix campaign covering rotation fitting, convergence criteria, trust radius, and rigid-body projection.
Table \ref{tab:code_map} summarizes the module correspondence.

\begin{table}[htbp]
\caption{Conceptual mapping between mathematical ideas, chemgp-core Rust modules, and gpr\_optim C++ classes. Both implementations share the same algorithmic structure; the Rust code runs the benchmarks reported in this review.}
\label{tab:code_map}
\centering
\small
\setlength{\tabcolsep}{3pt}
\begin{tabularx}{\textwidth}{@{}>{\raggedright\arraybackslash}p{0.22\textwidth}>{\raggedright\arraybackslash}X>{\raggedright\arraybackslash}p{0.25\textwidth}@{}}
\hline
Concept & chemgp-core (Rust) & gpr\_optim (C++)\\
\hline
Inverse-distance kernel & \texttt{kernel.rs} (\texttt{MolInvDistSE}) & \texttt{InvDistSE}\\
Derivative kernel blocks & \texttt{kernel.rs} (\texttt{molinvdist\_kernel\_blocks}) & \texttt{GPKernelBlocks}\\
GP training (SCG) & \texttt{train.rs} + \texttt{scg.rs} + \texttt{nll.rs} & \texttt{GPModel}\\
Dimer method & \texttt{dimer.rs} (\texttt{gp\_dimer}) & \texttt{DimerSearch}\\
NEB method & \texttt{neb.rs} + \texttt{neb\_oie.rs} & \texttt{NEBSearch}\\
OT-GP Dimer & \texttt{otgpd.rs} (\texttt{otgpd}) & \texttt{OTGPDimer}\\
FPS + EMD & \texttt{sampling.rs} + \texttt{emd.rs} & \texttt{FPSSampler, EMDDistance}\\
Trust regions & \texttt{trust.rs} & \texttt{TrustRegion}\\
Random Fourier features & \texttt{rff.rs} (\texttt{build\_rff}) & \texttt{RFFModel}\\
L-BFGS & \texttt{lbfgs.rs} + \texttt{optim\_step.rs} & \texttt{LBFGS}\\
Constant kernel & \texttt{covariance.rs} + \texttt{rff.rs} & --\\
\hline
\end{tabularx}
\end{table}

\begin{table}[H]
\caption{Equation-to-function mapping for chemgp-core. Each equation in the main text maps to a specific function in the Rust implementation. The annotations in the code listings connect variable names to the mathematical symbols in the equations.}
\begin{tabular}{lll}
\hline
Equation & chemgp-core Function & Purpose \\
\hline
Eq. \ref{eq:curvature} & \texttt{curvature()} & Dimer curvature estimate \\
Eq. \ref{eq:polak_ribiere} & \texttt{rotational\_force()} & CG rotation force \\
Eq. \ref{eq:modified_force} & \texttt{translational\_force()} & Householder reflection \\
Eq. \ref{eq:neb_force} & \texttt{neb\_force()} & NEB total force \\
Eq. \ref{eq:cov_f_df}--\ref{eq:cov_df_df} & \texttt{molinvdist\_kernel\_blocks()} & Derivative covariance blocks \\
Eq. \ref{eq:full_covariance} & \texttt{build\_full\_covariance()} & Assemble $\mathbf{K}_{\text{full}}$ \\
Eq. \ref{eq:cholesky_solve} & \texttt{robust\_cholesky()} & Guarded Cholesky factorization \\
Eq. \ref{eq:gp_mean} & \texttt{predict()} & Posterior mean prediction \\
Eq. \ref{eq:gp_var} & \texttt{predict\_with\_variance()} & Posterior mean + variance \\
Eq. \ref{eq:invdist_jacobian} & \texttt{invdist\_jacobian()} & Inverse-distance Jacobian \\
Eq. \ref{eq:emd_pertype}--\ref{eq:emd_overall} & \texttt{emd\_distance()} & EMD distance (per-type + overall) \\
Eq. \ref{eq:trust_ceiling} & \texttt{adaptive\_trust\_threshold()} & Physical trust ceiling \\
Eq. \ref{eq:bochner} & \texttt{build\_rff()} & RFF model construction \\
\hline
\end{tabular}
\label{tab:equation_to_function}
\end{table}

The code snippets below are extracted from chemgp-core, with annotations connecting variables to the equations in the preceding sections.
The same functions run the examples in this review and are deployed via the eOn saddle point search framework \cite{chillEONSoftwareLong2014} for production calculations.
\subsection{Kernel Evaluation and Analytical Derivative Blocks}
\label{sec:code-kernel}
The kernel evaluation (Eq. \ref{eq:idist_kernel}) computes the squared Mahalanobis distance in inverse-distance feature space and applies the SE exponential.
The variable \texttt{d2} accumulates the sum \(\sum_{(i,j)} \theta_{(i,j)}^2 (\phi_{ij}(\mathbf{x}) - \phi_{ij}(\mathbf{x}'))^2\), where \texttt{inv\_lengthscales} stores the \(\theta_{(i,j)} = 1/l_{\phi(i,j)}\) values and \texttt{compute\_inverse\_distances} returns the feature vector \(\boldsymbol{\phi}(\mathbf{x})\).

\emph{Parameterization convention}: The paper's kernel (Eq. \ref{eq:idist_kernel}) uses the standard SE form with a \(1/2\) factor in the exponent: \(k(\mathbf{x}, \mathbf{x}') = \sigma^2 \exp(-1/2 \sum_i (\phi_i(\mathbf{x}) - \phi_i(\mathbf{x}'))^2 / l_i^2)\). The code absorbs the \(1/2\) into the inverse lengthscale definition, so \(\theta_{\text{code}} = 1/(l_{\text{paper}} \sqrt{2})\).
The two forms are mathematically equivalent; the code's parameterization simplifies the derivative computation.

\begin{lstlisting}[language=rust,numbers=none]
impl MolInvDistSE {
    pub fn eval(&self, x: &[f64], y: &[f64]) -> f64 {
        let fx = compute_inverse_distances(x, &self.frozen_coords);
        let fy = compute_inverse_distances(y, &self.frozen_coords);
        let mut d2 = 0.0;
        if !self.feature_params_map.is_empty() {
            for i in 0..fx.len() {
                let idx = self.feature_params_map[i];
                let val = (fx[i] - fy[i]) * self.inv_lengthscales[idx];
                d2 += val * val;
            }
        } else {
            let theta = self.inv_lengthscales[0];
            for i in 0..fx.len() {
                let diff = fx[i] - fy[i];
                d2 += diff * diff;
            }
            d2 *= theta * theta;
        }
        self.signal_variance * (-d2).exp()
    }
}
\end{lstlisting}

The derivative blocks (\texttt{kernel\_blocks}, Listing S1 in the Supporting Information) compute the four covariance components between energy and force observations via chain rule through the inverse-distance Jacobian.
\subsection{Covariance Matrix Assembly and GP Training}
\label{sec:code-training}
The full covariance matrix (Eq. \ref{eq:full_covariance}) is assembled by calling \texttt{kernel\_blocks} for every pair of training configurations and placing the resulting \(2 \times 2\) block structure at the appropriate indices.
For \(N\) training points with \(D = 3N_{\text{atoms}}\) coordinates each, the matrix has dimension \(N(1+D) \times N(1+D)\): the first \(N\) rows/columns correspond to energies, and the remaining \(ND\) to forces.
The noise variances \(\sigma_E^2\) and \(\sigma_F^2\) are added to the respective diagonal blocks.

\begin{lstlisting}[language=rust,numbers=none]
pub fn build_full_covariance(
    kernel: &Kernel, x_data: &[f64], dim: usize, n: usize,
    noise_e: f64, noise_g: f64, jitter: f64, const_sigma2: f64,
) -> Mat<f64> {
    let total = n * (1 + dim);
    let mut k_mat = Mat::<f64>::zeros(total, total);
    for i in 0..n {
        let xi = &x_data[i * dim..(i + 1) * dim];
        let b = kernel.kernel_blocks(xi, xi);
        k_mat[(i, i)] = b.k_ee + const_sigma2 + noise_e + jitter;  // const_sigma2: added to all E-E entries (rank-1 matrix sigma_c^2 * 11^T)
        let s_g = n + i * dim;
        for d in 0..dim {
            k_mat[(i, s_g + d)] = b.k_ef[d];
            k_mat[(s_g + d, i)] = b.k_fe[d];
            k_mat[(s_g + d, s_g + d)] += noise_g + jitter;
        }
        for di in 0..dim {
            for dj in 0..dim {
                k_mat[(s_g + di, s_g + dj)] = b.k_ff[(di, dj)];
            }
        }
        for j in (i + 1)..n {
            let xj = &x_data[j * dim..(j + 1) * dim];
            let b = kernel.kernel_blocks(xi, xj);
            let j_s = n + j * dim;
            k_mat[(i, j)] = b.k_ee + const_sigma2;
            k_mat[(j, i)] = b.k_ee + const_sigma2;
            for d in 0..dim {
                k_mat[(i, j_s + d)] = b.k_ef[d];
                k_mat[(j_s + d, i)] = b.k_ef[d];
                k_mat[(s_g + d, j)] = b.k_fe[d];
                k_mat[(j, s_g + d)] = b.k_fe[d];
            }
            for di in 0..dim {
                for dj in 0..dim {
                    k_mat[(s_g + di, j_s + dj)] = b.k_ff[(di, dj)];
                    k_mat[(j_s + dj, s_g + di)] = b.k_ff[(di, dj)];
                }
            }
        }
    }
    // Floor sub-epsilon entries (matches MATLAB GPstuff: C(C<eps)=0)
    let eps = f64::EPSILON;
    for r in 0..total {
        for c in 0..total {
            if k_mat[(r, c)].abs() < eps { k_mat[(r, c)] = 0.0; }
        }
    }
    k_mat
}
\end{lstlisting}

The training loop (\texttt{train\_model}, Listing S2) minimizes the MAP-regularized NLL (Eq. \ref{eq:marginal_likelihood}) using the SCG optimizer \cite{mollerScaledConjugateGradient1993} with log-space reparameterization.

The GP-dimer main loop (\texttt{gp\_dimer}, Listing S3) alternates between outer iterations (oracle evaluations that grow the training set) and inner iterations (rotation and translation on the GP surface).
Convergence requires both small translational force and negative curvature.
The trust radius check (Eq. \ref{eq:max1dlog} or Eq. \ref{eq:emd_trust}) breaks the inner loop when exceeded.

The GP-NEB OIE outer loop (\texttt{gp\_neb\_oie}, Listing S4) selects images via configurable acquisition strategies: \texttt{MaxVariance}, \texttt{Ucb} (NEB force plus perpendicular uncertainty; the default), or \texttt{ExpectedImprovement}.
A critical implementation detail: acquisition uses forces computed \emph{before} inner relaxation, because re-predicting at relaxed positions with sparse training data produces unreliable NEB forces.

The RFF model (\texttt{build\_rff}, Listing S5) replaces the exact GP with Bayesian linear regression in a random feature space sampled from the kernel's spectral density (Section \ref{sec:rff} of the main text).
Prediction reduces to a dot product at \(\mathcal{O}(D_{\text{rff}})\) cost per component.
Activating RFF is a single configuration change:

\begin{lstlisting}[language=rust,numbers=none]
let config = NEBConfig {
    rff_features: 500,      // 0 = exact GP; >0 = RFF approximation
    max_gp_points: 40,      // per-bead subset for hyperparameter training
    acquisition: AcquisitionStrategy::Ucb,
    lcb_kappa: 2.0,         // UCB exploration weight
    const_sigma2: 0.0,      // constant kernel: 1.0 for molecular PES, 0.0 for models
    ..Default::default()
};
\end{lstlisting}
\subsection{Code Listings}
\label{sec:org2b4f356}

The listings here cover the inverse-distance kernel derivative blocks (Listing S1), MAP-regularized hyperparameter training (Listing S2), GP-dimer main loop (Listing S3), GP-NEB OIE acquisition loop (Listing S4), and Random Fourier feature construction (Listing S5).

*Listing S1.
\section{Derivative blocks for the inverse-distance kernel (\texttt{kernel\_blocks}).}
\label{sec:orgbb88b17}
The chain-rule structure projects the feature-space Hessian to Cartesian coordinates via the inverse-distance Jacobians.

\begin{lstlisting}[language=rust,numbers=none]
pub fn molinvdist_kernel_blocks(
    k: &MolInvDistSE, x1: &[f64], x2: &[f64],
) -> KernelBlocks {
    let (f1, j1) = invdist_jacobian(x1, &k.frozen_coords);
    let (f2, j2) = invdist_jacobian(x2, &k.frozen_coords);
    let nf = f1.len();

    // Per-feature theta^2 values
    let theta2: Vec<f64> = (0..nf).map(|i| {
        let idx = if k.feature_params_map.is_empty() { 0 }
                  else { k.feature_params_map[i] };
        k.inv_lengthscales[idx].powi(2)
    }).collect();

    // SE kernel value
    let r: Vec<f64> = (0..nf).map(|i| f1[i] - f2[i]).collect();
    let d2: f64 = (0..nf).map(|i| theta2[i] * r[i] * r[i]).sum();
    let kval = k.signal_variance * (-d2).exp();


    // Feature-space gradient: dk/df
    let mut dk_df2 = vec![0.0; nf];
    let mut dk_df1 = vec![0.0; nf];
    for i in 0..nf {
        let v = 2.0 * kval * theta2[i] * r[i];
        dk_df2[i] = v;
        dk_df1[i] = -v;
    }
    // Feature-space Hessian: H[i,j] = 2*kval*(theta2[i]*delta_ij - 2*u[i]*u[j])
    let u: Vec<f64> = (0..nf).map(|i| theta2[i] * r[i]).collect();
    let mut h_feat = Mat::<f64>::zeros(nf, nf);
    for i in 0..nf {
        h_feat[(i, i)] = 2.0 * kval * (theta2[i] - 2.0 * u[i] * u[i]);
        for j in (i + 1)..nf {
            let val = -4.0 * kval * u[i] * u[j];
            h_feat[(i, j)] = val;
            h_feat[(j, i)] = val;
        }
    }

    // Chain rule: project to Cartesian coordinates
    let k_ee = kval;
    let k_ef = mat_t_vec(&j2, &dk_df2);  // J2^T * dk/df2  (D x 1)
    let k_fe = mat_t_vec(&j1, &dk_df1);  // J1^T * dk/df1  (D x 1)
    let k_ff = jt_h_j(&j1, &h_feat, &j2); // J1^T H J2     (D x D)
    KernelBlocks { k_ee, k_ef, k_fe, k_ff }
}
\end{lstlisting}

*Listing S2.
\section{MAP-regularized hyperparameter training via SCG (\texttt{train\_model}).}
\label{sec:orgb17cb14}
All hyperparameters are optimized in log-space; Cholesky failure returns infinity to reject infeasible configurations.

\begin{lstlisting}[language=rust,numbers=none]
pub fn train_model(model: &mut GPModel, iterations: usize) {
    // Pack hyperparameters to log-space
    let mut w0 = Vec::with_capacity(1 + model.kernel.n_ls_params());
    w0.push(model.kernel.signal_variance().ln());
    for &l in model.kernel.inv_lengthscales() {
        w0.push(l.ln());
    }
    let w_prior = w0.clone();
    let prior_var = compute_prior_variances(&model.kernel);

    let mut fg = |w: &[f64]| -> (f64, Vec<f64>) {
        nll_and_grad(w, &model.x_data, model.dim, model.n_train,
                     &model.y, &model.kernel, model.noise_var,
                     model.grad_noise_var, model.jitter,
                     &w_prior, &prior_var, model.const_sigma2,
                     model.prior_dof, model.prior_s2, model.prior_mu)
    };

    let config = ScgConfig {
        max_iter: iterations,
        tol_f: 1e-4,
        lambda_init: model.scg_lambda_init,
        ..Default::default()
    };

    let result = scg_optimize(&mut fg, &w0, &config);
    if result.converged || result.f_best < f64::INFINITY {
        let sigma2 = result.w_best[0].exp();
        let inv_ls: Vec<f64> =
            result.w_best[1..].iter().map(|v| v.exp()).collect();
        model.kernel = model.kernel.with_params(sigma2, inv_ls);
    }
}
\end{lstlisting}

*Listing S3.
\section{GP-dimer main loop (\texttt{gp\_dimer}).}
\label{sec:org5b6ae68}
The outer loop evaluates the oracle and grows the training set; the inner loop performs rotation and translation on the GP surface.

\begin{lstlisting}[language=rust,numbers=none]
pub fn gp_dimer(
    oracle: &OracleFn, x_init: &[f64], orient_init: &[f64],
    kernel: &Kernel, config: &DimerConfig,
) -> DimerResult {
    let mut state = DimerState {
        r: x_init.to_vec(),
        orient: normalize_vec(orient_init),
        dimer_sep: config.dimer_sep,
    };
    let mut td = TrainingData::new(x_init.len());

    // Bootstrap: evaluate midpoint and one dimer endpoint
    let (e, g) = oracle(x_init);
    td.add_point(x_init, e, &g);
    let r1 = dimer_endpoint(&state);
    let (e1, g1) = oracle(&r1);
    td.add_point(&r1, e1, &g1);

    for _outer in 0..config.max_outer_iter {
        // FPS subset selection for hyperparameter training
        let td_sub = select_fps_subset(&td, &state.r, config);

        // Train GP on subset, predict on full data (RFF if configured)
        let mut gp = GPModel::new(kernel.clone(), &td_sub, ...);
        train_model(&mut gp, config.gp_train_iter);
        let model = build_pred_model(&gp.kernel, &td, config);

        // Inner loop: rotate + translate on GP surface
        for _inner in 0..config.max_inner_iter {
            rotate_dimer(&mut state, &model, config);
            let (g0, g1, _e0) = predict_dimer_gradients(&state, &model);
            let f_trans = translational_force(&g0, &state.orient);
            if vec_norm(&f_trans) < config.t_force_gp { break; }
            let r_new = translate_dimer_lbfgs(&state, &g0, &g1, config);
            if exceeds_trust(&r_new, &td, config) { break; }
            state.r = r_new;
        }

        // Evaluate oracle at proposed position
        let (e_true, g_true) = oracle(&state.r);
        td.add_point(&state.r, e_true, &g_true);

        // Converge when true force is small AND curvature is negative
        let f_true = translational_force(&g_true, &state.orient);
        if vec_norm(&f_true) < config.t_force_true && curvature < 0.0 {
            return DimerResult { converged: true, oracle_calls: td.npoints(), .. };
        }
    }
    DimerResult { converged: false, oracle_calls: td.npoints(), .. }
}
\end{lstlisting}

*Listing S4.
\section{GP-NEB OIE acquisition loop (\texttt{gp\_neb\_oie}).}
\label{sec:orgdf38f2e}
The \texttt{select\_image} function supports \texttt{MaxVariance}, \texttt{Ucb}, and \texttt{ExpectedImprovement} strategies.

\begin{lstlisting}[language=rust,numbers=none]
pub fn gp_neb_oie(
    oracle: &OracleFn, x_start: &[f64], x_end: &[f64],
    kernel: &Kernel, config: &NEBConfig,
) -> NEBResult {
    let mut images = init_path(x_start, x_end, config);
    let mut td = TrainingData::new(x_start.len());

    // Evaluate endpoints and midpoint
    for x in &[x_start, x_end, &images[config.images / 2]] {
        let (e, g) = oracle(x);
        td.add_point(x, e, &g);
    }

    for _outer in 0..config.max_outer_iter {
        let model = train_and_build_model(&td, kernel, config);

        // Acquire: select unevaluated image by acquisition strategy
        let i_eval = select_image(
            &config.acquisition, &images, &energies,
            &unevaluated, &model, &cached_forces, config,
        );

        // Evaluate oracle at selected image
        let (e, g) = oracle(&images[i_eval]);
        td.add_point(&images[i_eval], e, &g);

        // Relax path on GP surface (L-BFGS inner loop)
        images = oie_inner_relax(&model, &images, &td, config);

        // Convergence check on true forces
        let max_f = compute_neb_forces(&images, &model, config).max_f;
        if max_f < config.conv_tol { /* verify unevaluated images */ }
    }
}
\end{lstlisting}

*Listing S5.
\section{Random Fourier feature construction (\texttt{build\_rff}).}
\label{sec:org0491734}
Frequency vectors are sampled from the kernel's spectral density (Bochner's theorem); the constant kernel adds one extra basis function.

\begin{lstlisting}[language=rust,numbers=none]
pub fn build_rff(
    kernel: &Kernel, x_train: &[f64], y_train: &[f64],
    dim: usize, n: usize, d_rff: usize,
    noise_var: f64, grad_noise_var: f64, seed: u64,
    const_sigma2: f64,
) -> RffModel {
    let inv_ls = kernel.inv_lengthscales();
    let d_feat = kernel.n_features(dim);

    // Sample frequencies from N(0, 2*theta^2 * I) (Bochner's theorem)
    let mut rng = StdRng::seed_from_u64(seed);
    let mut w = Mat::<f64>::zeros(d_rff, d_feat);
    for f in 0..d_feat {
        let idx = kernel.pair_type_index(f);
        let scale = (2.0f64).sqrt() * inv_ls[idx];
        for i in 0..d_rff {
            w[(i, f)] = rng.sample::<f64, _>(StandardNormal) * scale;
        }
    }
    let b: Vec<f64> = (0..d_rff).map(|_| rng.random::<f64>() * 2.0 * PI).collect();
    let c = kernel.signal_variance().sqrt() * (2.0 / d_rff as f64).sqrt();

    // Design matrix Z: d_eff = d_rff + 1 (extra column for const kernel)
    let d_eff = d_rff + 1;
    let n_obs = n * (1 + dim);
    let mut z = Mat::<f64>::zeros(n_obs, d_eff);
    for i in 0..n {
        let (zi, j_z) = rff_features(&w, &b, c, &x_train[i*dim..(i+1)*dim]);
        for f in 0..d_eff { z[(i, f)] = zi[f]; }         // Eq. rff_features
        for d in 0..dim {
            for f in 0..d_eff { z[(n + i*dim + d, f)] = j_z[(f, d)]; }
        }
    }

    // Bayesian linear regression: A = Z^T diag(prec) Z + I
    let prec = build_precision(n, dim, noise_var, grad_noise_var);
    let a = zt_diag_z_plus_eye(&z, &prec, d_eff);
    let llt = a.llt(Side::Lower).expect("RFF Cholesky failed");
    let rhs = zt_diag_y(&z, &prec, y_train, d_eff);
    let alpha = llt.solve(&rhs);
    RffModel { w, b, c, alpha, a_chol: llt, dim, const_sigma2, .. }
}
\end{lstlisting}
\section{Mathematical Derivations}
\label{sec:org72f6072}

\subsection{Rigid-Body Mode Basis Construction}
\label{sec:org0be994e}

The projection of rigid-body modes requires an orthonormal basis \(\{\mathbf{u}_k\}_{k=1}^{6}\) spanning the 6 external degrees of freedom.
For a molecule with \(N\) atoms and Cartesian coordinates \(\mathbf{x} = (x_1, y_1, z_1, \ldots, x_N, y_N, z_N)^T \in \mathbb{R}^{3N}\), the six basis vectors correspond to three translations and three infinitesimal rotations.

The unnormalized translation vectors are:

\begin{align}
\mathbf{t}_x &= (1, 0, 0, 1, 0, 0, \ldots, 1, 0, 0)^T, \\
\mathbf{t}_y &= (0, 1, 0, 0, 1, 0, \ldots, 0, 1, 0)^T, \\
\mathbf{t}_z &= (0, 0, 1, 0, 0, 1, \ldots, 0, 0, 1)^T.
\end{align}

The unnormalized infinitesimal rotation vectors (about the molecular center of mass \(\mathbf{r}_0\)) are:

\begin{align}
\mathbf{r}_x &= (0, z_1 - z_0, -(y_1 - y_0), \ldots, 0, z_N - z_0, -(y_N - y_0))^T, \\
\mathbf{r}_y &= (-(z_1 - z_0), 0, x_1 - x_0, \ldots, -(z_N - z_0), 0, x_N - x_0)^T, \\
\mathbf{r}_z &= (y_1 - y_0, -(x_1 - x_0), 0, \ldots, y_N - y_0, -(x_N - x_0), 0)^T.
\end{align}

where \((x_i, y_i, z_i)\) are the coordinates of atom \(i\), and \((x_0, y_0, z_0)\) is the center of mass.

These six vectors are linearly independent but not orthonormal.
The Gram-Schmidt process applied in the order \((\mathbf{t}_x, \mathbf{t}_y, \mathbf{t}_z, \mathbf{r}_x, \mathbf{r}_y, \mathbf{r}_z)\) produces the orthonormal basis \(\{\mathbf{u}_k\}_{k=1}^{6}\):

\begin{align}
\mathbf{u}_1 &= \frac{\mathbf{t}_x}{\|\mathbf{t}_x\|}, \\
\mathbf{u}_2 &= \frac{\mathbf{t}_y - (\mathbf{t}_y \cdot \mathbf{u}_1)\mathbf{u}_1}{\|\mathbf{t}_y - (\mathbf{t}_y \cdot \mathbf{u}_1)\mathbf{u}_1\|}, \\
\mathbf{u}_3 &= \frac{\mathbf{t}_z - \sum_{j=1}^{2} (\mathbf{t}_z \cdot \mathbf{u}_j)\mathbf{u}_j}{\|\mathbf{t}_z - \sum_{j=1}^{2} (\mathbf{t}_z \cdot \mathbf{u}_j)\mathbf{u}_j\|}, \\
\mathbf{u}_{k+3} &= \frac{\mathbf{r}_k - \sum_{j=1}^{k+2} (\mathbf{r}_k \cdot \mathbf{u}_j)\mathbf{u}_j}{\|\mathbf{r}_k - \sum_{j=1}^{k+2} (\mathbf{r}_k \cdot \mathbf{u}_j)\mathbf{u}_j\|}, \quad k \in \{x, y, z\}.
\end{align}

In practice, the translation vectors are already orthogonal (and of equal norm \(\sqrt{N}\)), so the Gram-Schmidt process only needs to orthogonalize the rotation vectors against the translations and against each other.
This basis is computed once per molecular geometry and cached; the projection then costs \(\mathcal{O}(N)\) per step.
\subsection{Hyperparameter Oscillation Detection}
\label{sec:org192918d}

The oscillation diagnostic mentioned in Section \ref{sec:map-regularization} of the main text detects when the MAP estimate of hyperparameters oscillates between competing local minima as new data arrives.
For a hyperparameter vector \(\boldsymbol{\theta}(t) = (\theta_1(t), \ldots, \theta_K(t))\) at outer iteration \(t\), the per-component oscillation indicator is:

\begin{equation}
O_j(t) = \begin{cases}
1, & \text{if } (\theta_j(t) - \theta_j(t-1))(\theta_j(t-1) - \theta_j(t-2)) < 0, \\
0, & \text{otherwise}.
\end{cases}
\end{equation}

This indicator is 1 when hyperparameter \(j\) reverses direction (sign change in the gradient) between consecutive steps, and 0 otherwise.
The fraction of oscillating components over a sliding window of length \(W\) (typically \(W = 5\) in chemgp-core) is:

\begin{equation}
f_{\text{osc}}(t) = \frac{1}{KW} \sum_{j=1}^{K} \sum_{s=t-W+1}^{t} O_j(s).
\end{equation}

When \(f_{\text{osc}}(t)\) exceeds a threshold \(p_{\text{osc}}\) (default \(p_{\text{osc}} = 0.8\) in chemgp-core), the algorithm detects instability and triggers growth of the FPS subset \(M_{\text{sub}}\) (Section \ref{sec:fps}).
The subset grows incrementally (default: +2 points per retry) up to a maximum size (default: 30).
This adaptive subset sizing sharpens the MLL landscape by adding geometrically diverse training data, constraining the optimizer to a narrower region of hyperparameter space.
\subsection{Kernel Block Structure}
\label{sec:org9f7fcb5}

The full covariance matrix for derivative observations has a block structure that arises from the chain rule applied to the inverse-distance feature map.
Throughout this section and in \texttt{chemgp-core/src/kernel.rs}, these derivative blocks are written for energy gradients \(\nabla V\); atomic forces enter later through the separate convention \(\mathbf{F} = -\nabla V\).
For two configurations \(\mathbf{x}_1\) and \(\mathbf{x}_2\), the kernel blocks are:

\begin{align}
k_{ee} &= k(\phi(\mathbf{x}_1), \phi(\mathbf{x}_2)) \\\
k_{ef} &= \frac{\partial k}{\partial \mathbf{x}_2} = \left(\frac{\partial k}{\partial \boldsymbol{\phi}}\right)^T \frac{\partial \boldsymbol{\phi}}{\partial \mathbf{x}_2} = \mathbf{J}_2^T \frac{\partial k}{\partial \boldsymbol{\phi}} \\\
k_{fe} &= \frac{\partial k}{\partial \mathbf{x}_1} = \mathbf{J}_1^T \frac{\partial k}{\partial \boldsymbol{\phi}} \\\
k_{ff} &= \frac{\partial^2 k}{\partial \mathbf{x}_1 \partial \mathbf{x}_2^T} = \mathbf{J}_1^T \left( \frac{\partial^2 k}{\partial \boldsymbol{\phi} \partial \boldsymbol{\phi}^T} - 2 \frac{\partial k}{\partial \boldsymbol{\phi}} \otimes \frac{\partial \boldsymbol{\phi}}{\partial \mathbf{x}_2} \right) \mathbf{J}_2
\end{align}

where \(\mathbf{J}_i = \partial \boldsymbol{\phi} / \partial \mathbf{x}_i\) is the Jacobian of the feature map at configuration \(\mathbf{x}_i\), and the feature-space Hessian is computed analytically for the SE kernel.
\subsection{Earth Mover's Distance for Trust Regions}
\label{sec:orgcb8c5a4}

The EMD-based trust region uses the per-type distance (Eq. \ref{eq:emd_pertype}) to compute a size-independent metric.
For two configurations \(\mathbf{x}_1\) and \(\mathbf{x}_2\) with atom types \(\{t_i\}\), the per-type distance is:

\begin{equation}
d_{\text{EMD}}^{(t)} = \min_{\pi \in \Pi_t} \sum_{i \in \mathcal{I}_t^{(1)}} \sum_{j \in \mathcal{I}_t^{(2)}} c_{ij}^{(t)} \pi_{ij}
\label{eq:emd_pertype_supplement}
\end{equation}

where \(\Pi_t\) is the set of joint distributions with marginals matching the counts of type \(t\) atoms, and \(c_{ij}^{(t)} = \|\mathbf{x}_1^{(i)} - \mathbf{x}_2^{(j)}\|\) is the ground cost.
To match the implementation in \texttt{chemgp-core/src/emd.rs}, the per-type transport cost is converted to a mean displacement by dividing by the number of atoms of that type, and the overall intensive EMD is the maximum over atom types:

\begin{equation}
d_{\text{EMD}}(\mathbf{x}_1, \mathbf{x}_2) = \max_t \frac{d_{\text{EMD}}^{(t)}}{N_t}
\label{eq:emd_overall_supplement}
\end{equation}
This is the same convention as Eq. \ref{eq:emd_overall} in the main text.
\subsection{OIE Acquisition Criterion Derivation}
\label{sec:org26d6815}

The Upper Confidence Bound (UCB) acquisition criterion for NEB OIE balances exploitation (large NEB forces) against exploration (high uncertainty).
The criterion is:

\begin{equation}
\alpha(\mathbf{R}_i) = |\mathbf{F}_i^{\text{NEB}}| + \kappa \cdot \sigma_\perp(\mathbf{R}_i, \boldsymbol{\tau}_i)
\label{eq:ucb_acquisition_supplement}
\end{equation}

where \(\sigma_\perp\) is the perpendicular gradient variance (Eq. \ref{eq:perp_variance}).
This is derived from the Gaussian tail bound: with probability at least \(1 - \delta\),

\begin{equation}
|F_d^{\text{true}} - F_d^{\text{GP}}| \leq \kappa \cdot \sigma(F_d^{\text{GP}}), \quad \kappa = \sqrt{2 \log(1/\delta)}
\end{equation}

Applying this bound to the NEB force magnitude gives the UCB criterion as a conservative estimate of the true force magnitude.
Within Eq. \ref{eq:ucb_acquisition_supplement}, \(\kappa = 0\) gives force-only selection.
The separate pure-variance selector used by \texttt{AcquisitionStrategy::MaxVariance} is instead Eq. \ref{eq:oie_selection} from the main text, where image choice is based only on the GP energy variance.
\section{Benchmark Summary}
\label{sec:bench-summary}
Table \ref{tab:bench-summary} consolidates the oracle-call counts that appear across the figures of the main text and the runs shipped with \texttt{chemgp-core}.
The numbers come directly from the JSONL outputs of the corresponding example binaries (\texttt{cargo run -{}-{}release -{}-{}example <name>}) and use the convergence thresholds documented in those examples.
For CI-based path-search methods, the convergence basis is the climbing-image force rather than the whole-band maximum force.
Production-scale benchmarks across hundreds of molecular reactions are reported in the companion papers \cite{goswamiEfficientImplementationGaussian2025,goswamiAdaptivePruningIncreased2025}; the table below reflects the illustrative cases used in this review and is intended for reproduction, not as a stand-alone validation.
No repeated-run uncertainty intervals or wall-clock comparisons are claimed for this tutorial table; wall time is reported separately in the benchmark harness rather than in the main tutorial summary.
The toy-surface rows are included as pedagogical controls and negative cases, not as molecular validation of the inverse-distance kernel.
In the executable examples these rows use the Cartesian toy-surface kernel path, whereas the molecular benchmarks use the inverse-distance kernel; the toy-surface call counts are therefore not evidence for or against the molecular speedups attributed to the inverse-distance representation.

\begin{table}[htbp]
\caption{Oracle-call counts to convergence for the illustrative benchmarks used in this review. Each row corresponds to a \texttt{chemgp-core} example whose execution trace drives a figure in the main text or a literature-aligned molecular benchmark in the accompanying ChemGP harness. "Baseline" denotes the classical reference algorithm for that task. "Variant A" and "Variant B" are method-specific comparison columns so the table can mix minimization, dimer, and CI-targeted path-search examples without forcing them into one GP/OT-GP taxonomy. Smaller is better.}
\label{tab:bench-summary}
\centering
\scriptsize
\setlength{\tabcolsep}{2pt}
\begin{tabularx}{\textwidth}{@{}>{\raggedright\arraybackslash}p{0.13\textwidth}>{\raggedright\arraybackslash}p{0.19\textwidth}>{\raggedright\arraybackslash}Xrrr@{}}
\hline
Task & System & Variants & Baseline & Variant A & Variant B\\
\hline
Minimization & LEPS (9D) & gp\_minimize vs L-BFGS & 57 & 9 & -\\
Minimization & Muller-Brown & gp\_minimize vs L-BFGS & 34 & 50 & -\\
Minimization & PET-MAD molecule & gp\_minimize vs L-BFGS & 45 & 13 & -\\
Single saddle & Muller-Brown & gp\_dimer vs std. dimer & 6 & 7 & -\\
Single saddle & LEPS (9D) & gp\_dimer / OTGPD vs dimer & 10 & 8 & 8\\
Single saddle & d000 metatomic molecular run & standard dimer vs OTGPD & 50 & - & 13\\
Path (NEB) & LEPS (9D) & gp\_neb AIE / OIE vs NEB & 156 & 100 & 42\\
Path (CI-NEB) & system100 (PET-MAD) & literature baseline OIE vs CI-NEB & 112 & - & 23\\
\hline
\end{tabularx}
\end{table}

For minimization, the GP overhead does not pay off uniformly: on the cheap Muller-Brown surface, where each oracle call has trivial cost, the GP loop loses to L-BFGS (50 vs 34 oracle calls).
That row should be read as a negative control for the surrogate loop on a cheap Cartesian toy surface, not as a statement about the molecular inverse-distance kernel.
The picture flips as the oracle becomes expensive: the PET-MAD minimization case shows a roughly threefold reduction, and the dimer and CI-targeted path-search rows show the largest oracle savings because the surrogate guides the search through anisotropic regions where direct methods make many short steps.
The final system100 row should be read with its method definition in mind: the OIE benchmark is a CI-targeted comparison using the climbing-image force as the stopping basis (\(CI|F| < 0.1\) eV/\AA{}), not a whole-band max-force comparison.
This pattern is consistent with the conditions under which surrogate acceleration is expected to help (Section 1, main text) and with the broader benchmarks in \cite{goswamiEfficientImplementationGaussian2025,goswamiAdaptivePruningIncreased2025}.

\end{appendix}
\section{References}
\label{sec:org4217771}
\bibliography{otgp_tutorial,manual}
\end{document}